\documentclass[11pt, a4paper, table, xdraw]{article}
\usepackage[preprint]{acl}
\usepackage{times,latexsym}
\usepackage{url}
\usepackage[T1]{fontenc}
\usepackage[utf8]{inputenc}
\usepackage{microtype}
\usepackage{inconsolata}
\usepackage{graphicx}

\usepackage{adjustbox}
\usepackage{booktabs}
\usepackage{multirow}

\usepackage{placeins}

\newcommand{\huggingface}[1]{\href{https://huggingface.co/#1}{{#1}}}
\newcommand{\github}[1]{\href{https://github.com/#1}{{github.com/#1}}}
\newcommand{\apache}{\href{https://www.apache.org/licenses/LICENSE-2.0}{Apache 2.0}}
\newcommand{\mitlicense}{\href{https://opensource.org/license/mit}{MIT}}
\newcommand{\ccbyncfour}{\href{https://creativecommons.org/licenses/by-nc/4.0}{CC BY-NC 4.0}}
\newcommand{\ccbyncsafour}{\href{https://creativecommons.org/licenses/by-nc-sa/4.0}{CC BY-NC-SA 4.0}}
\newcommand{\ccbyncsathree}{\href{https://creativecommons.org/licenses/by-nc-sa/3.0}{CC BY-NC-SA 3.0}}
\newcommand{\ccbysafour}{\href{https://creativecommons.org/licenses/by-sa/4.0}{CC BY-SA 4.0}}

\definecolor{GermanRed}{HTML}{B02D5F}
\definecolor{DialectBlue}{HTML}{62A8EF}

\title{Standard-to-Dialect Transfer Trends Differ across Text and Speech:\\A Case Study on Intent and Topic Classification in German Dialects}

\usepackage{fontawesome5}
\newcommand{\lmu}{\faMountain}
\newcommand{\mcml}{\faRobot}
\newcommand{\textMod}{\faPen}
\newcommand{\speechMod}{\faComment[regular]}

\author{%
Verena Blaschke\kern1pt\textsuperscript{\lmu\mcml}
\quad
Miriam Winkler\kern1pt\textsuperscript{\lmu}
\quad
Barbara Plank\kern1pt\textsuperscript{\lmu\mcml}
 \\
\textsuperscript{\lmu} MaiNLP lab, CIS, LMU Munich, Germany\\
\textsuperscript{\mcml} Munich Center for Machine Learning, Germany\\
\texttt{\{verena.blaschke, b.plank\} @lmu.de}}

\newcommand{\textspeechmassive}{\mbox{(Speech-)}\allowbreak{}MASSIVE}
\newcommand{\xlmr}{\mbox{XLM-R}}
\newcommand{\xlsr}{\mbox{XLS-R}}

\colorlet{blueMid}{blue!55!white}
\colorlet{redDark}{red!50!gray}
\newcommand{\wrong}[1]{\textcolor{redDark}{#1}}
\newcommand{\barok}[1]{\textcolor{blueMid}{#1}}
\newcommand{\gloss}[1]{\textit{\textcolor{gray}{#1}}}

\begin{document}
\maketitle

\begin{abstract}
Research on cross-dialectal transfer from a standard to a non-standard dialect variety has typically focused on text data.
However, dialects are primarily spoken, and non-standard spellings 
cause issues in text processing.
We compare standard-to-dialect transfer in three settings: text models, speech models, and cascaded systems where speech first gets automatically transcribed and then further processed by a text model.
We focus on German dialects in the context of written and spoken intent classification -- releasing the first dialectal audio intent classification dataset -- with supporting experiments on topic classification.
The speech-only setup provides the best results on the dialect data while the text-only setup works best on the standard data. 
While the cascaded systems lag behind the text-only models for German, they perform relatively well on the dialectal data if the transcription system generates normalized, standard-like output.
\end{abstract}

\section{Introduction}

While most natural language processing (NLP) research focuses on highly standardized languages with large amounts of training and evaluation data available, a subfield focusing on dialect NLP has recently been emerging, studying the (lack of) robustness NLP tools exhibit towards dialectal variation \cite{zampieri-etal-2020-natural, joshi-etal-2025-natural}.

In dialect NLP, oftentimes no training but only evaluation data are available for non-standard dialect varieties.
In such cases it is common to train NLP models on a higher-resource, closely related standard language instead.
Part of what makes such standard-to-dialect transfer challenging in contrast to standard transfer learning is the large degree of spelling variation exhibited in dialectal writing, as it frequently leads to infelicitous subword tokenizations \cite{aepli-etal-2023-findings, blaschke-etal-2023-manipulating, srivastava-chiang-2025-calling, kanjirangat-etal-2025-tokenization}.
Such challenges have led to research on processing dialectal data with token-free models, such as for example pixel models \cite{munoz-ortiz-etal-2025-evaluating}.
These input representation challenges also bring about the question of how robust speech models are towards dialectal variation, given the continuous nature of their inputs.
This has to some extent been investigated for automatic speech recognition (ASR; \S\ref{sec:related-work}), but not yet for classification tasks.
Focusing on speech is especially relevant for processing non-standard dialects, as these varieties are predominantly spoken.

\begin{figure}
    \centering
    \includegraphics[width=\linewidth]{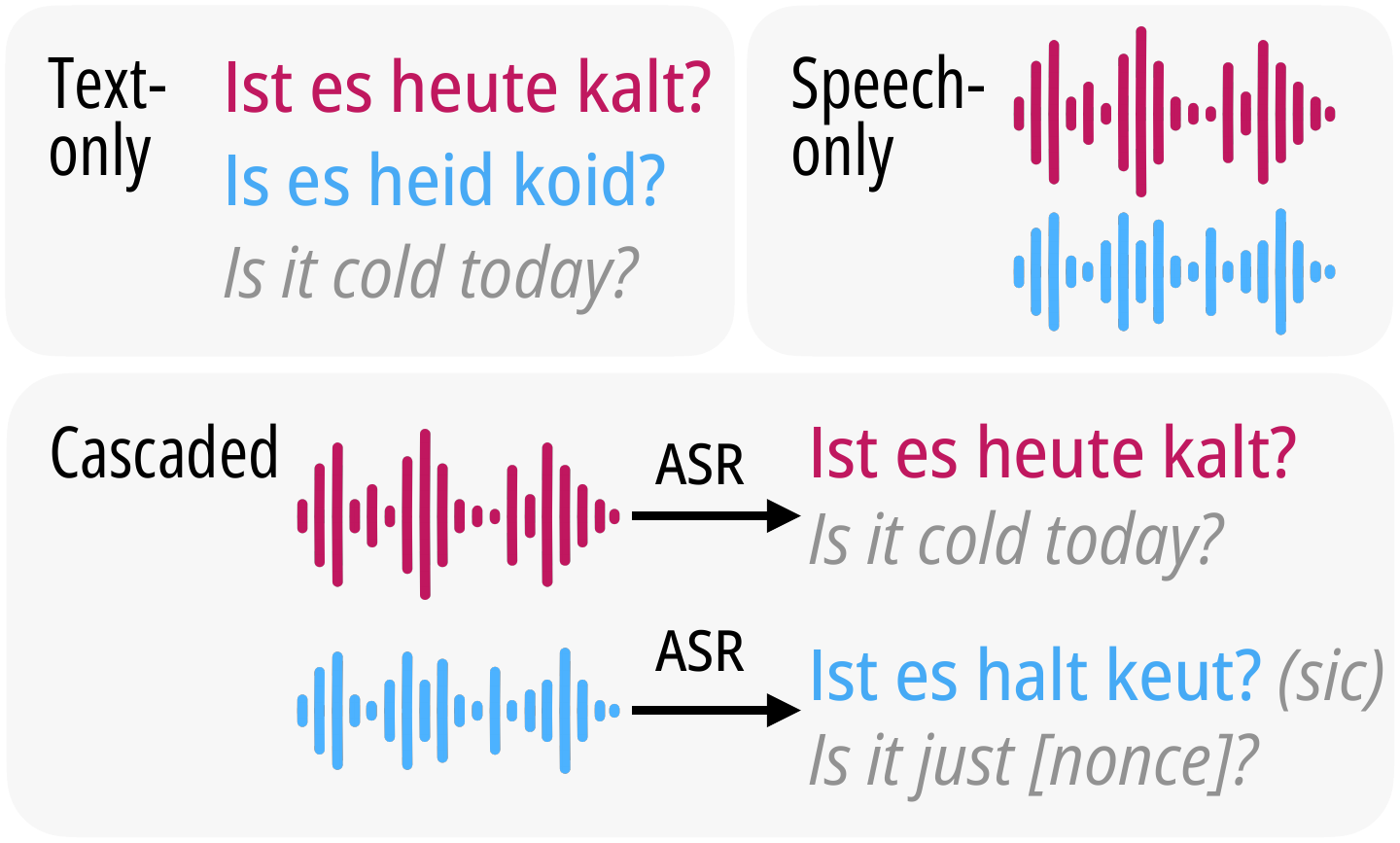}
    \caption{We compare three evaluation setups for \textcolor{GermanRed}{German} and \textcolor{DialectBlue}{dialectal} text and speech data.
    }
    \label{fig:setups}
\end{figure}

Our work primarily focuses on intent classification in the context of queries for virtual 
assistants for two reasons:
Firstly, dialect-robust 
virtual assistants
are seen as especially desirable in a survey on potential NLP technologies for German dialects \cite{blaschke-etal-2024-dialect}.
Secondly, this task naturally lends itself to the spoken domain, but most available datasets only include written data.
This can partially be explained by the conventional cascaded paradigm of first automatically transcribing audio data (cf.\ \citealp{faruqui-hakkani-tur-2022-revisiting}), and then classifying the transcribed text as a second step.
However, this approach ignores the shortcomings of current ASR systems with respect to low-resource and non-standardized languages.

In this study, we extend a text-only intent classification dataset to audio with recordings in German and a related non-standard dialect, Bavarian.
We compare three setups across a variety of text and speech encoders and ASR models (Figure~\ref{fig:setups}): models trained and evaluated on the written version of the dataset (``text-only''), on the spoken version (``speech-only'') or trained on text and evaluated on ASR transcripts of the spoken input (``cascaded'').
We replicate our findings in a supporting experiment on a conceptually similar task (topic classification) with German and Swiss German (dialect).

We present the following findings, which show that results for standard languages do not necessarily generalize to even closely related non-standard dialects:
\begin{enumerate}
    \item While the cascaded models lag behind the text-only models for the German data, systems whose ASR step in practice normalizes the data outperform the text-only models for dialectal input~(\S\ref{sec:analysis-cascaded-text}).
    \item Although the speech-only systems perform on par with the cascaded systems on the German data, they greatly outperform the cascaded systems on the dialectal data~(\S\ref{sec:analysis-speech-cascaded}).
    \item Although the text-only models perform much better than the speech-only models on the German data, the situation is largely reversed for the dialectal data~(\S\ref{sec:analysis-speech-text}).
\end{enumerate}

\noindent
Furthermore, we contribute a spoken intent dataset for German and Bavarian~(\S\ref{sec:data-intents}). %
We make the dataset\footnote{\url{https://doi.org/10.5281/zenodo.19554427}} and our code and model predictions\footnote{\href{https://github.com/mainlp/dialects-text-vs-speech}{\texttt{github.com/mainlp/dialects-text-vs-speech}}} publicly available.

\section{Related Work}
\label{sec:related-work}

\paragraph{Intent classification for written dialects}
Several works have focused on standard-to-dialect transfer for written dialects, starting with the xSID dataset \cite{van-der-goot-etal-2021-masked},
and including multiple shared tasks \cite{aepli-etal-2023-findings, malaysha-etal-2024-arafinnlp, scherrer-etal-2025-findings}.
A common theme in these works is the gap between the performance on the standard variety and the non-standard dialects.
Intent classification systems have been shown to be somewhat affected by syntactic variation \cite{artemova-etal-2024-exploring}.
Other works have focused on improving intent classification by mitigating the effect of infelicitous subword tokenizations for dialectal text by manipulating the subword tokenization of the standard language as well \cite{srivastava-chiang-2023-fine, blaschke-etal-2025-add} or by using token-free models \cite{munoz-ortiz-etal-2025-evaluating}.

\paragraph{Spoken dialect processing}
Most work on spoken dialect processing has focused on ASR, covering a variety of languages like 
Arabic \cite{djanibekov-etal-2025-dialectal}, 
Dutch \cite{bosch-2000-asr},
English \cite{torgbi-etal-2025-adapting},
German \cite{beringer-etal-1998-german}, 
Greek \cite{vakirtzian-etal-2024-speech}, 
Frisian \cite{amooie-etal-2025-enhancing},
Irish \cite{lonergan-etal-2023-towards}, 
Mandarin \cite{tang-etal-2021-kespeech}, 
Telugu \cite{yadavalli-etal-2022-multi-task},
Wenzhounese \cite{gao-etal-2025-wenzhou},
and 
Yorùbá \cite{ahia-etal-2024-voices}.
Common themes in these works are performance gaps between standard and non-standard varieties, the effect of dialectal pronunciation differences, as well as lexical and syntactic differences between the dialect and the standard.

The most relevant works for our study analyze the performance of current ASR models on German dialects.
In a study on automatically transcribing German dialects, \citet{blaschke-etal-2025-multi-dialectal} find that ASR models differ in how much they normalize their output towards standard German. In a similar vein, \citet{dolev-etal-2024-whisper} and \citet{gorisch-schmidt-2024-evaluating} find that Whisper ASR models \cite{radford-etal-2023-robust} tend to replace colloquial or regional German syntactic constructions with constructions that are more common in written German, although this is not always successful \cite{sicard-etal-2023-spaiche}.

Focusing on a different task -- question answering -- \citet{faisal-etal-2021-sd-qa} evaluate cascaded models on different nation-level varieties of five languages.
They find performance gaps between varieties of the same language and a positive correlation between transcription quality and question answering performance.

\paragraph{Spoken intent and topic classification}
A range of recent works has focused on spoken intent and topic classification.
\citet{bastianelli-etal-2020-slurp} present an English-language intent classification dataset that was later extended into the multilingual Speech-MASSIVE by \citet{lee-etal-2024-speech-massive}.
Similarly, \citet{schmidt-etal-2025-fleurs-slu} and \citet{keller-glavas-2025-speechtaxi} present large-scale multilingual speech classification datasets.
\citet{rajaa2022skits2i} present an Indian English intent classification dataset.
Two datasets include multiple varieties of the same language: MInDS-14 \cite{gerz-etal-2021-multilingual} includes standard English accents from three countries, with no consistent performance differences between them.
ITALIC contains recordings from different Italian regions \cite{koudounas-etal-2023-italic}, but no analysis of performance differences across regions.
On the engineering side, \citet{wang-etal-2023-whislu} and \citet{he-garner-2023-interpreter} explore the benefits of multi-task learning for spoken intent detection.
In the context of cascaded systems, \citet{faruqui-hakkani-tur-2022-revisiting} argue for a better integration of ASR systems and text-based classification models.

\section{Experiments}
\label{sec:experiments}

\subsection{Setups}
\label{sec:experimental-setups}

We compare three setups.
In the \textbf{text-only} setup, the models are trained on the German text data and evaluated on the German and dialectal text data. 
In the \textbf{cascaded} setup, the same models are evaluated on ASR transcriptions of the German and dialectal speech data.%
\footnote{We additionally train some cascaded models on ASR transcriptions, but find that they show the same transfer trends as the cascaded models trained on gold-standard text data~(\S\ref{sec:appendix-cascaded-additional}).}
Lastly, in the \textbf{speech-only} setup, the models are trained on the German audio data and evaluated on the German and dialectal audio data.

\begin{table}[t]
\setlength{\tabcolsep}{2pt}
\adjustbox{max width=\linewidth}{%
\begin{tabular}{@{}lll@{}lllll@{}}
\toprule
eng & \multicolumn{7}{l}{Remind me to call Stephanie on Tuesday} \\ %
deu & Erinnere & mich &  & Stephanie & am & Dienstag & anzurufen \\
bar & Erinner & mi & d & Stephanie & am & Dienstag & ozumruafa \\
\gloss{} & \gloss{remind} & \gloss{me} & \gloss{the} & \gloss{Stephanie} & \gloss{on} %
& \gloss{Tuesday} & \gloss{to.call} \\
\midrule
\end{tabular}}

\adjustbox{max width=\linewidth}{%
\begin{tabular}{@{}lllllll@{}}
eng & \multicolumn{6}{l}{Because many services that we use today...} \\
deu & Denn & viele & Dienste, & die wir & heute & nutzen, \\
gsw-be & Denn & vili & Dienschtä, & womer & hüt & nutzä, \\
gsw-lu & Well & vell & Dienst, & womer & höt & bruched, \\
 & \gloss{because} & \gloss{many} & \gloss{services} & \gloss{that we} & \gloss{today} & \gloss{use} \\
\end{tabular}}

\adjustbox{max width=\linewidth}{%
\begin{tabular}{@{}llllllll@{}}
eng & \multicolumn{6}{l}{...only work with the help of satellites.}\\ %
deu & funktionieren & nur & mit & Hilfe & von & Satelliten. \\
gsw-be & funktionierä & numä & mit & Hiuf & vo & Satellitä. \\
gsw-lu & fonktionierted & nor & met & Helf & vo & Satellite. \\
& \gloss{work} & \gloss{only} & \gloss{with} & \gloss{help} & \gloss{of} & \gloss{satellites} & \phantom{....} \\
 \bottomrule
\end{tabular}}
\caption{\textbf{Standard German and dialectal sentences from the datasets we use} (top: German and Bavarian sentences from xSID; bottom: German and two of the Swiss German sentences -- Bern, Lucerne -- from SwissDial).
While there are some grammatical and lexical differences between the varieties, the most salient differences are on the level of pronunciation/spelling.
Note that the xSID sentences were independently translated from English -- while the sentence pair in this table is parallel on a word level, this does not apply to all sentences. For more examples, see Table~\ref{tab:asr-examples} in Appendix~\ref{sec:appendix-asr-examples}.
}
\label{tab:dialect-examples}
\end{table}

\begin{table}[t]
\adjustbox{max width=\linewidth}{%
\begin{tabular}{@{}l@{}l@{\,}l@{~}l@{\,}r@{~}r@{~}r@{~}l@{}}
\toprule
\textbf{Dataset} & \multicolumn{2}{l}{\textbf{Mod}} & \textbf{Lang} & \textbf{Train} & \textbf{Dev} & \textbf{Test} &  \\
\midrule
\multicolumn{7}{@{}l}{\textit{Intent classification}}  \\
Speech-MASSIVE & &\speechMod & deu & 2.5k & 459 & 361 & \multirow{3}{*}{\hspace{-1em}$\left.\begin{array}{l}
                \\
                \\\end{array}\right\rbrace$p\rlap{.}} \\
MASSIVE (deu) & \textMod && deu & 2.5k & 459 & 361 & \\
MASSIVE (bar) & \textMod && bar & --~ & --~ & 361 &  \\
xSID & \textMod && deu, bar &  --~ &  --~ & 2$\times$412 & \multirow{2}{*}{\hspace{-1em}$\left.\begin{array}{l}
                \\\end{array}\right\rbrace$p\rlap{.}} \\
xSID-audio \textit{(new)} & &\speechMod & deu, bar & --~ & --~ & 2$\times$412 &  \\
\midrule
\multicolumn{3}{@{}l}{\textit{Topic classification}} &  &  &  &  \\
SwissDial (deu) & \textMod & & deu & 1.5k & 194 & 396 & \multirow{2}{*}{\hspace{-1em}$\left.\begin{array}{l}
                \\\end{array}\right\rbrace$p\rlap{.}} \\
SwissDial (gsw) & \textMod{} & \speechMod & gsw \rlap{(8 dials)} & -- & --~ & 8$\times$396 & \\
\bottomrule
\end{tabular}}
\caption{\textbf{Datasets used in this study.} 
The numbers of instances reflect the subsets we used in our experiments (for instance, in order to have comparable intent label sets across datasets), e.g., the full version of xSID-audio contains 300 development and 500 test instances per variety.
Several datasets are parallel (\textit{p.})\ to each other.
Modality: \textMod{}\,=\,text, \speechMod\,=\,speech. 
}
\label{tab:datasets}
\end{table}

\subsection{Data}
\label{sec:data}

We use data in Standard German (deu) and two related dialects: Bavarian (bar) and Swiss German (gsw).
Neither Bavarian nor Swiss German are standardized.
Both are primarily spoken, but also occasionally written, especially in informal situations.
The most salient differences between the standard and non-standard varieties are on the pronunciation level (which is reflected when they are written), although there are also some lexical and grammatical differences (Table~\ref{tab:dialect-examples}).

To the best of our knowledge, no classification datasets with parallel spoken and written data in a standard and non-standard variety are available for any other language/dialect.

\paragraph{Main experiment: intent classification}
\label{sec:data-intents}

Our main set of experiments concerns intent classification.
We use (and introduce) data in German and a closely related dialect, Bavarian.
The data consist of queries and commands for virtual assistants annotated with 
user intents.

We use the German splits of the text-based MASSIVE dataset \cite{fitzgerald-etal-2023-massive} and its audio counterpart Speech-MASSIVE \cite{lee-etal-2024-speech-massive} for training models.
For evaluation, we use the subset of the MASSIVE test set that has been translated into Bavarian \cite{winkler-etal-2024-slot} as well as the corresponding German \textspeechmassive{} instances.
We furthermore use the German and Bavarian test splits of xSID \cite{van-der-goot-etal-2021-masked, winkler-etal-2024-slot}.%
\footnote{xSID is an evaluation dataset, and does not provide (high-quality) training splits for German or Bavarian.
For the Bavarian data, we use the split from Upper Bavaria (``de-ba''). 
} %
Because xSID is only released as a text-based dataset, we additionally create audio recordings (read speech).
The audios are recorded in-house by a native speaker of both German and Bavarian who also translated the written Bavarian split.
The German and Bavarian recordings are identical in content, speaker, and recording conditions, and only differ in their language variety.
Appendix~\ref{sec:appendix-datasheet} contains the data statement for xSID-audio.

We map the instances in \textspeechmassive{} to the label set used by xSID (Appendix~\ref{sec:appendix-intent-labels}), and exclude all instances that do not match this mapping.
As shown in Table~\ref{tab:datasets}, this leaves 2.5k/459/361 train/dev/test instances for \textspeechmassive{} and 412 test instances for xSID, and 10 labels.%
\footnote{The full xSID-audio~0.1 release contains recordings for the entire Bavarian and German development and test sets (300+500 sentences, 16 intents).} %
There is no overlap in speakers between the training and evaluation splits.
Because all text is lowercased in MASSIVE, we lowercase the xSID data and the ASR transcriptions as well.

\paragraph{Supporting experiment: topic classification}
\label{sec:data-topics}

To replicate our findings to the extent possible with 
available data,
we further include experiments on 
topic classification, a task 
conceptually similar to intent classification.
Except for our 
extension of xSID, the dataset below is the only 
classification dataset that contains parallel (or otherwise comparable) dialectal and standard-variety data where at least the dialectal split is available as both spoken and written data.

We use SwissDial \cite{swissdial}, which contains German sentences that are also parallel in up to eight Swiss German dialects.
The German sentences are only available as text data, but the dialectal sentences are both written and spoken, allowing us to use the dataset for the text-only and cascaded setups.
We exclude rare or incoherent labels (Appendix~\ref{sec:appendix-topic-labels}), resulting in a label set of 10~topics.
We split the dataset into 1.5k/194/396 train/dev/test sentences.
The test set only contains sentences that are parallel across all eight dialects.
For training and development, we only use the Standard German versions.

\vspace{0.5\baselineskip}
\noindent

\subsection{Models}
\label{sec:models}

\begin{table}[t]
\centering
\adjustbox{max width=\linewidth}{%
\begin{tabular}{@{}llrll@{}}
\toprule
& \textbf{Model} & \textbf{\llap{Encoder} size (M)} & \textbf{ASR} \\
\midrule
\textit{Text} & mBERT & 86\\
& mDeBERTa & 85 \\
& XLM-R base & 86\\
& XLM-R large & 303\\
\midrule
\textit{Speech} & mHuBERT & 94\\
& XLS-R 300M & 315 \\
& XLS-R 300M DE & 315& CTC\\
& MMS 300M & 315\\
& Whisper tiny & 8& LM\\
& Whisper base & 20& LM \\
& Whisper small & 87& LM \\
& Whisper medium & 306& LM \\
& Whisper large-v3 & 635& LM \\
\midrule
\textit{Add.\ ASR} & \multicolumn{2}{l}{XLS-R 1B DE} & CTC \\
& \multicolumn{2}{l}{MMS 1B-all} & CTC \\
& \multicolumn{2}{l}{Whisper large-v3-turbo} & LM \\
\bottomrule
\end{tabular}}
\caption{%
\textbf{Text/speech encoders and additional ASR models.}
Encoder size in millions of non-embedding parameters.
ASR decoding strategies: 
CTC\,=\,connectionist temporal classification, LM\,=\,language modelling.
For licenses and links, see Appendix~\ref{sec:appendix-resource-licenses}.
}
\label{tab:models}
\end{table}

Table~\ref{tab:models} provides an overview of the models we use.
We focus on encoder models, as fine-tuned pre-trained encoder models have been shown to outperform larger instruction-tuned decoders on classification tasks \cite{adelani-etal-2024-sib, ojo-etal-2025-afrobench, kasa-etal-2025-generative, saattrup-nielsen-etal-2025-encoder}.
Furthermore, fine-tuning the pre-trained models below gives us more control over the experimental setup, as we can ensure each model is exposed to exactly the same labelled intent/topic classification training data.

\paragraph{Text and speech models of comparable size}
We focus on two sets of pretrained multilingual transformer models with comparable backbone parameter sizes (encoder parameters without embedding layers): 
{$\sim$90\,M} backbone parameters (12 encoder layers,  758 hidden dimensions) 
and
{$\sim$300\,M} 
parameters (24 encoder layers, 1024 hidden dimensions).
For encoder--decoder models, we only fine-tune the encoder (plus a classification head).

As text models,
we select mBERT \cite{devlin-etal-2019-bert}, \xlmr{} base and large \cite{conneau-etal-2020-unsupervised}, and mDeBERTa-v3 \cite{he2021debertav3, he-etal-2021-deberta}.
The latter has shown strong performance in previous work on intent classification for dialectal German data \cite{artemova-etal-2024-exploring, kruckl-etal-2025-improving}.
As speech models, 
we use the encoders of Whisper small and medium \cite{radford-etal-2023-robust}.
We also use mHubERT-147 \cite{zanon-boito-etal-2024-mhubert} and MMS 300M \cite{pratap-etal-2024-scaling}.
We additionally use both the original version of \xlsr{} 300M \cite{babu-etal-2022-xlsr} and a version fine-tuned for German ASR \cite{mcdowell-2022-xlsr300de}.

All of the models we use include German in their pre-training data.
Only mBERT includes Bavarian; none of the models were trained on Swiss German per their documentation.%
\footnote{
The training datasets of the speech models are not openly accessible, but the metadata indicate no inclusion of dialectal training data. 
The fact that the ASR-capable speech models all perform very poorly on Bavarian ASR (Table~\ref{tab:asr}) indicates that it is unlikely these models were exposed to Bavarian data, despite performing quite strongly on Bavarian in the speech-only setup. 
The qualitative ASR examples in~\S\ref{sec:appendix-asr-examples} further illustrate this.}

\paragraph{Additional model sizes}
To further investigate the effect of model size, we also include the Whisper model sizes tiny (8\,M backbone encoder parameters), base (20\,M), and large-v3 (635\,M).
No additional text model sizes are available.

\paragraph{ASR models}
To transcribe the evaluation data in the cascaded setup, we use all speech models mentioned above in this section that are trained/\allowbreak{}fine-tuned for ASR.
To evaluate the effect of the ASR performance on the cascaded systems in greater detail, we additionally
use a version of \xlsr{} 1B fine-tuned for German ASR \cite{mcdowell-2022-xlsr1bde}, MMS 1B fine-tuned for multilingual ASR, and Whisper large-v3-turbo, a version of Whisper large-v3 with a reduced number of decoder layers.
All Whisper models are trained for multilingual ASR.
We use all ASR models without further fine-tuning and set German as the output language, as none of the ASR models support German dialects.

The Whisper models consist of a stack of encoder layers (that we also use in the speech-only setup) followed by a stack of decoder layers that act as a language model.
The \xlsr{} and MMS models are based on wav2vec2 \cite{baevski-etal-2020-wav2vec2} and trained for connectionist temporal classification (CTC), mapping the model's internal representations to individual characters based on the audio alone without further information on plausible character or word sequences in the output language.

We use the word/character error rate for evaluating ASR performance.
We also manually evaluate a subset of the transcriptions based on whether they are in fluent German, retain the overall meaning of the utterance, and retain keywords related to the intent of the sentence (details in~\S\ref{sec:appendix-asr-manual}).

\begin{table*}
\centering
\setlength{\tabcolsep}{1pt} %
\adjustbox{max width=\textwidth}{%
\begin{tabular}{@{}lrrrrrr@{\hspace{10pt}}rrrrrr@{\hspace{25pt}}rrrrr@{}}
\toprule
& \multicolumn{11}{c}{\textbf{Intents}} &  & \multicolumn{5}{c}{\textbf{Topics}} \\
\cmidrule(rl){2-12} \cmidrule(rl){14-18}
& \multicolumn{5}{c}{\textbf{MASSIVE}} && \multicolumn{5}{c}{\textbf{xSID}} &  & \multicolumn{5}{c}{\textbf{Swissdial}} \\
\cmidrule(rl){2-6} \cmidrule(rl){8-12} \cmidrule(rl){14-18}
\textbf{Model} & \multicolumn{1}{c}{\textbf{deu}} &  & \multicolumn{1}{c}{\textbf{bar}} &  & \multicolumn{1}{c}{$\Delta$} &  & \multicolumn{1}{c}{\textbf{deu}} &  & \multicolumn{1}{c}{\textbf{bar}} &  & \multicolumn{1}{c}{$\Delta$} &  & \multicolumn{1}{c}{\textbf{deu}} &  & \multicolumn{1}{c}{\textbf{gsw}} &  & \multicolumn{1}{c}{$\Delta$} \\
\cmidrule(r){1-1} \cmidrule{2-12} \cmidrule{14-18}
Majority class baseline & 29.1\phantom{\textsubscript{0.0}} &  & 29.1\phantom{\textsubscript{0.0}} & & & & 9.5\phantom{\textsubscript{0.0}} &  & 9.5\phantom{\textsubscript{0.0}} & & & & 21.0\phantom{\textsubscript{0.0}} &  & 21.0\phantom{\textsubscript{0.0}} & \\
Random baseline & 10.0\phantom{\textsubscript{0.0}} & & 10.0\phantom{\textsubscript{0.0}} & & & & 10.0\phantom{\textsubscript{0.0}} &  & 10.0\phantom{\textsubscript{0.0}} & & & & 10.0\phantom{\textsubscript{0.0}} &  & 10.0\phantom{\textsubscript{0.0}} & \\
\cmidrule(r){1-1} \cmidrule{2-12} \cmidrule{14-18}
\textit{Text-only} & \textit{93.1\textsubscript{1.6}} && \textit{59.7\textsubscript{4.4}} &&&& \textit{81.8\textsubscript{4.6}} && \textit{58.8\textsubscript{5.2}} &&&& \textit{54.0\textsubscript{3.3}} && \textit{43.7\textsubscript{3.8}}\\
mBERT & \cellcolor[HTML]{57BB8A}94.1\textsubscript{0.6} &  & {\cellcolor[HTML]{BCE4D1}62.8}{\textsubscript{5.1}} &  & {\cellcolor[HTML]{E7A080}--31.3} &  & \cellcolor[HTML]{7AC9A2}83.5\textsubscript{0.0} &  & \cellcolor[HTML]{B5E1CC}65.0\textsubscript{1.3} &  & \cellcolor[HTML]{F3D0C0}--18.5 &  & {\cellcolor[HTML]{7BCAA3}52.8}{\textsubscript{2.1}} &  & \cellcolor[HTML]{C5E8D7}41.2{\textsubscript{1.0}} &  & {\cellcolor[HTML]{E59774}--11.6} \\
mDeBERTa & \cellcolor[HTML]{59BC8C}93.5\textsubscript{0.8} &  & {\cellcolor[HTML]{D0ECDE}56.8}{\textsubscript{1.0}} &  & {\cellcolor[HTML]{E38C65}--36.7} &  & \cellcolor[HTML]{88CFAC}79.0\textsubscript{0.6} &  & \cellcolor[HTML]{D7EFE4}54.4\textsubscript{2.6} &  & \cellcolor[HTML]{EEB9A2}--24.6 &  & {\cellcolor[HTML]{84CDA9}51.4}{\textsubscript{1.2}} &  & \cellcolor[HTML]{B6E2CC}43.6{\textsubscript{2.6}} &  & {\cellcolor[HTML]{FFFFFF}--7.9} \\
XLM-R base & \cellcolor[HTML]{5BBD8D}92.9\textsubscript{2.5} &  & {\cellcolor[HTML]{BEE5D2}62.3}{\textsubscript{3.1}} &  & {\cellcolor[HTML]{E8A384}--30.6} &  & \cellcolor[HTML]{89D0AD}78.6\textsubscript{7.1} &  & \cellcolor[HTML]{CFECDE}57.0\textsubscript{5.0} &  & \cellcolor[HTML]{F0C4B0}--21.7 &  & {\cellcolor[HTML]{78C9A1}53.3}{\textsubscript{3.2}} &  & \cellcolor[HTML]{C5E8D6}41.3{\textsubscript{4.0}} &  & {\cellcolor[HTML]{E38C65}--12.0} \\
XLM-R large & \cellcolor[HTML]{5FBF90}91.8\textsubscript{1.7} &  & {\cellcolor[HTML]{CFECDE}56.9}{\textsubscript{4.4}} &  & {\cellcolor[HTML]{E4926E}--34.9} &  & \cellcolor[HTML]{72C69D}85.9\textsubscript{2.5} &  & \cellcolor[HTML]{C9E9D9}59.0\textsubscript{4.8} &  & \cellcolor[HTML]{EBB096}--26.9 &  & {\cellcolor[HTML]{57BB8A}58.3}{\textsubscript{1.3}} &  & \cellcolor[HTML]{95D4B5}48.7{\textsubscript{1.0}} &  & {\cellcolor[HTML]{F3CDBD}--9.6} \\
\cmidrule(r){1-1} \cmidrule{2-12} \cmidrule{14-18}
\textit{Cascaded} & \textit{88.3\textsubscript{4.2}} &&&&&& \textit{77.1\textsubscript{6.0}} && \textit{52.5\textsubscript{9.8}} &&&&&& \textit{42.9\textsubscript{8.7}}\\
\multicolumn{5}{@{}l@{}}{\textit{... averaged over ASR models}} \\
mBERT & \cellcolor[HTML]{65C194}89.8\textsubscript{3.6} &  &  &  &  &  & \cellcolor[HTML]{84CEAA}80.2\textsubscript{3.6} &  & \cellcolor[HTML]{C3E7D5}60.9\textsubscript{7.0} &  & \cellcolor[HTML]{F2CDBC}--19.4 &  &  &  & \cellcolor[HTML]{C3E7D6}41.5{\textsubscript{8.5}} \\
mDeBERTa & \cellcolor[HTML]{6CC499}87.8\textsubscript{5.1} &  &  &  &  &  & \cellcolor[HTML]{9ED8BB}72.3\textsubscript{5.1} &  & \cellcolor[HTML]{F9FDFB}43.9\textsubscript{8.1} &  & \cellcolor[HTML]{EAAB8F}--28.4 &  &  &  & \cellcolor[HTML]{C8E9D9}40.8{\textsubscript{8.8}} \\
XLM-R base & \cellcolor[HTML]{6AC397}88.3\textsubscript{3.8} &  &  &  &  &  & \cellcolor[HTML]{93D4B4}75.5\textsubscript{6.1} &  & \cellcolor[HTML]{E4F4EC}50.5\textsubscript{7.5} &  & \cellcolor[HTML]{EDB79F}--25.0 &  &  &  & \cellcolor[HTML]{B9E3CE}43.1{\textsubscript{8.4}} \\
XLM-R large & \cellcolor[HTML]{6EC49A}87.3\textsubscript{3.9} &  &  &  &  &  & \cellcolor[HTML]{83CDA9}80.5\textsubscript{4.9} &  & \cellcolor[HTML]{D6EFE3}54.8\textsubscript{8.1} &  & \cellcolor[HTML]{EDB59C}--25.7 &  &  &  & \cellcolor[HTML]{A5DBC0}46.3{\textsubscript{8.6}} \\
\multicolumn{5}{@{}l@{}}{\textit{... averaged over text models}} \\
XLS-R 300M DE & \cellcolor[HTML]{6BC398}88.0\textsubscript{2.1} &  &  &  &  &  & \cellcolor[HTML]{93D3B4}75.8\textsubscript{4.4} &  & \cellcolor[HTML]{ECF7F2}48.1\textsubscript{7.1} &  & \cellcolor[HTML]{EBAE92}--27.7 &  &  &  & \cellcolor[HTML]{FFFFFF}32.1{\textsubscript{3.1}} \\
XLS-R 1B DE & \cellcolor[HTML]{68C296}89.0\textsubscript{1.8} &  &  &  &  &  & \cellcolor[HTML]{86CEAB}79.6\textsubscript{4.5} &  & \cellcolor[HTML]{E6F5EE}49.7\textsubscript{6.6} &  & \cellcolor[HTML]{E9A587}--29.9 &  &  &  & \cellcolor[HTML]{F9FDFB}33.1{\textsubscript{3.5}} \\
MMS 1B-all & \cellcolor[HTML]{67C295}89.2\textsubscript{2.1} &  &  &  &  &  & \cellcolor[HTML]{93D4B4}75.7\textsubscript{4.3} &  & \cellcolor[HTML]{E9F7F0}48.8\textsubscript{4.7} &  & \cellcolor[HTML]{ECB096}--26.9 &  &  &  & \cellcolor[HTML]{DFF2E9}37.3{\textsubscript{3.3}} \\
Whisper tiny & \cellcolor[HTML]{85CEAB}79.8\textsubscript{2.8} &  &  &  &  &  & \cellcolor[HTML]{A7DCC2}69.5\textsubscript{5.9} &  & \cellcolor[HTML]{FFFFFF}41.9\textsubscript{8.4} &  & \cellcolor[HTML]{EBAE92}--27.6 &  &  &  & \cellcolor[HTML]{EAF7F0}35.6{\textsubscript{2.3}} \\
Whisper base & \cellcolor[HTML]{78C9A1}84.1\textsubscript{1.9} &  &  &  &  &  & \cellcolor[HTML]{9CD7BA}72.8\textsubscript{5.6} &  & \cellcolor[HTML]{F1FAF5}46.5\textsubscript{9.2} &  & \cellcolor[HTML]{ECB399}--26.3 &  &  &  & \cellcolor[HTML]{BFE5D2}42.2{\textsubscript{2.3}} \\
Whisper small & \cellcolor[HTML]{67C295}89.4\textsubscript{1.7} &  &  &  &  &  & \cellcolor[HTML]{87CFAC}79.3\textsubscript{4.6} &  & \cellcolor[HTML]{CEEBDD}57.4\textsubscript{8.7} &  & \cellcolor[HTML]{F0C3AF}--21.9 &  &  &  & \cellcolor[HTML]{92D3B3}49.2{\textsubscript{2.5}} \\
Whisper medium & \cellcolor[HTML]{61BF91}91.2\textsubscript{1.4} &  &  &  &  &  & \cellcolor[HTML]{85CEAA}80.0\textsubscript{4.7} &  & \cellcolor[HTML]{CFECDE}57.0\textsubscript{8.0} &  & \cellcolor[HTML]{EFBFAA}--23.0 &  &  &  & \cellcolor[HTML]{82CDA8}51.7{\textsubscript{2.9}} \\
Whisper large-v3-turbo~ & \cellcolor[HTML]{5FBE90}91.8\textsubscript{1.5} &  &  &  &  &  & \cellcolor[HTML]{82CDA8}80.8\textsubscript{4.9} &  & \cellcolor[HTML]{C0E6D3}61.8\textsubscript{6.7} &  & \cellcolor[HTML]{F3CEBD}--19.1 &  &  &  & \cellcolor[HTML]{7DCAA4}52.6{\textsubscript{3.1}} \\
Whisper large-v3 & \cellcolor[HTML]{5EBE8F}92.2\textsubscript{1.7} &  &  &  &  &  & \cellcolor[HTML]{82CDA8}80.8\textsubscript{4.8} &  & \cellcolor[HTML]{C1E6D4}61.4\textsubscript{6.6} &  & \cellcolor[HTML]{F2CDBC}--19.4 &  &  &  & \cellcolor[HTML]{7BCAA4}52.7{\textsubscript{3.1}} \\
\cmidrule(r){1-1} \cmidrule{2-12} \cmidrule{14-18}
\textit{Speech-only} & \textit{83.1\textsubscript{7.8}} &&&&&& \textit{70.6\textsubscript{8.9}} && \textit{60.8\textsubscript{7.9}}\\
mHuBERT & \cellcolor[HTML]{95D4B5}75.1\textsubscript{1.5} &  &  &  &  &  & \cellcolor[HTML]{C6E8D8}59.7\textsubscript{0.7} &  & \cellcolor[HTML]{DAF0E5}53.6\textsubscript{0.5} &  & \cellcolor[HTML]{FFFFFF}--6.1 \\
XLS-R 300M & \cellcolor[HTML]{92D3B3}76.1\textsubscript{0.2} &  &  &  &  &  & \cellcolor[HTML]{BEE5D2}62.1\textsubscript{0.4} &  & \cellcolor[HTML]{D8EFE4}54.3\textsubscript{0.6} &  & \cellcolor[HTML]{FDF8F6}--7.8 \\
XLS-R 300M DE\textsuperscript{ASR} & \cellcolor[HTML]{62C092}90.9\textsubscript{0.3} &  &  &  &  &  & \cellcolor[HTML]{92D3B3}76.1\textsubscript{2.2} &  & \cellcolor[HTML]{B2E0C9}66.0\textsubscript{1.1} &  & \cellcolor[HTML]{FBF0EB}--10.0 \\
MMS 300M & \cellcolor[HTML]{8CD1AF}77.9\textsubscript{0.4} &  &  &  &  &  & \cellcolor[HTML]{B2E0CA}65.9\textsubscript{1.6} &  & \cellcolor[HTML]{CBEADB}58.2\textsubscript{1.2} &  & \cellcolor[HTML]{FDF8F6}--7.8 \\
Whisper tiny\textsuperscript{ASR} & \cellcolor[HTML]{A1D9BE}71.3\textsubscript{2.2} &  &  &  &  &  & \cellcolor[HTML]{C7E8D8}59.6\textsubscript{1.8} &  & \cellcolor[HTML]{EDF8F3}47.6\textsubscript{2.8} &  & \cellcolor[HTML]{F9E8E0}--12.1 \\
Whisper base\textsuperscript{ASR} & \cellcolor[HTML]{78C9A1}84.1\textsubscript{0.7} &  &  &  &  &  & \cellcolor[HTML]{A4DAC0}70.5\textsubscript{0.9} &  & \cellcolor[HTML]{BBE4CF}63.3\textsubscript{1.5} &  & \cellcolor[HTML]{FEFBF9}--7.1 \\
Whisper small\textsuperscript{ASR} & \cellcolor[HTML]{65C194}89.9\textsubscript{0.2} &  &  &  &  &  & \cellcolor[HTML]{84CEAA}80.3\textsubscript{1.2} &  & \cellcolor[HTML]{A8DCC2}69.2\textsubscript{2.8} &  & \cellcolor[HTML]{FAECE5}--11.1 \\
Whisper medium\textsuperscript{ASR} & \cellcolor[HTML]{61BF91}91.0\textsubscript{0.8} &  &  &  &  &  & \cellcolor[HTML]{88CFAC}79.0\textsubscript{1.3} &  & \cellcolor[HTML]{BDE5D1}62.6\textsubscript{3.6} &  & \cellcolor[HTML]{F5D8CB}--16.3 \\
Whisper large-v3\textsuperscript{ASR} & \cellcolor[HTML]{60BF91}91.4\textsubscript{0.5} &  &  &  &  &  & \cellcolor[HTML]{7CCAA4}82.7\textsubscript{1.1} &  & \cellcolor[HTML]{9DD8BB}72.5\textsubscript{0.1} &  & \cellcolor[HTML]{FBEFEA}--10.2 \\
\textit{... ASR models} & \textit{86.4\textsubscript{7.5}} &&&&&& \textit{74.7\textsubscript{8.1}} && \textit{63.5\textsubscript{8.3}}\\
\textit{... non-ASR models} & \textit{76.4\textsubscript{1.5}} &&&&&& \textit{62.6\textsubscript{2.9}} && \textit{55.4\textsubscript{2.2}}\\
\bottomrule
\end{tabular}
}
\caption{\textbf{Although the speech-only models generally show the worst results for the German data, they give the overall best results in the dialectal setting for xSID and Swissdial. The performance of the cascaded models varies strongly for the dialectal data.}
Accuracy (in~\%) on the German (deu) and dialectal (Bavarian, bar; Swiss German, gsw) test sets, and the difference between them ($\Delta$~bar--deu, $\Delta$~gsw--deu).
Darker shades of green: better performance; darker shades of red: greater performance differences (the colour scales for intents and topics are independent of one another).
All results are averaged over three random seeds.
The cascaded systems are additionally averaged over ASR/text models (de-aggregated results in Appendix~\ref{sec:appendix-deaggregated-results}, Table~\ref{tab:results-cascaded-detailed}).
Standard deviations are in subscripts.
For performance differences between the setups, see Table~\ref{tab:deltas}.
Average performances per setup type are in \textit{italics}.
\textsuperscript{ASR}\,=\,speech models tuned for ASR.
}
\label{tab:results-overview}
\end{table*}

\paragraph{Seeds and hyperparameters}
We choose hyperparameter values based on prior work on written and spoken intent classification
\cite{rajaa2022skits2i, koudounas-etal-2023-italic, arora-etal-2024-universlu, lee-etal-2024-speech-massive, schmidt-etal-2025-fleurs-slu}.
We use the training and development splits of the German \textspeechmassive{} and Swissdial versions for selecting the learning rate%
\footnote{We observe that the text models are much more robust to the choice of learning rate than the speech models, with the speech models without ASR tuning being especially sensitive (Table~\ref{tab:learning_rates} in Appendix~\ref{sec:appendix-hyperparams-training}).}
and maximum number of training epochs.
Details on the hyperparameter search and model training are in~\S\ref{sec:appendix-hyperparams-training}.
We report the mean accuracy values over three random seeds.

\section{Results}
\label{sec:results}

Table~\ref{tab:results-overview} shows the overall intent/topic classification results per setup.
The detailed results for each ASR and text model combination for the cascaded setup are in Appendix~\ref{sec:appendix-deaggregated-results}, Table~\ref{tab:results-cascaded-detailed}.

As in prior work on zero-shot standard-to-dialect transfer~(\S\ref{sec:related-work}), the results on the German test data are consistently better than on the dialectal data, with performance gaps between 6.1 and 36.7 percentage points~(pp.)\ in accuracy for intent classification and gaps between  7.9 and 12.0\,pp.\ for topic classification.
For the intent classification data, the performance also drops in the cross-dataset transfer -- the (German) test set accuracies for xSID are lower than for MASSIVE, although the dialect gap is much narrower for xSID (6.1--29.9 for xSID compared to 30.6--36.7 for MASSIVE).

\begin{table*}
\centering
\setlength{\tabcolsep}{1pt} %
\adjustbox{max width=\linewidth}{%
\begin{tabular}{@{}lrrrrrrrr@{\hspace{5pt}}rrrr@{\hspace{10pt}}rrrrrrr@{\hspace{5pt}}rrrr@{}}
\toprule
& \multicolumn{11}{c}{\textbf{WER}} & & \multicolumn{11}{c}{\textbf{CER}} \\
\cmidrule(rl){2-12} \cmidrule(rl){14-24}
& \multicolumn{7}{c}{\textbf{Intents}} & & \multicolumn{3}{c}{\textbf{Topics}} &  & \multicolumn{7}{c}{\textbf{Intents}} & & \multicolumn{3}{c}{\textbf{Topics}} \\
\cmidrule(rl){2-8} \cmidrule(rl){10-12} \cmidrule(rl){14-20} \cmidrule(rl){22-24}
& \multicolumn{1}{c}{\textbf{MAS.}} &  & \multicolumn{1}{c}{\textbf{xSID}} &  & \multicolumn{3}{c}{\textbf{xSID}} &  & \multicolumn{3}{c}{\textbf{Swissdial}}  &  & \multicolumn{1}{c}{\textbf{MAS.}} &  & \multicolumn{1}{c}{\textbf{xSID}} &  & \multicolumn{3}{c}{\textbf{xSID}} &  & \multicolumn{3}{c}{\textbf{Swissdial}} \\
& \multicolumn{1}{c}{\textbf{(deu)}} &  & \multicolumn{1}{c}{\textbf{(deu)}} &  & \multicolumn{3}{c}{\textbf{(bar)}} &  & \multicolumn{3}{c}{\textbf{(gsw)}}  &  &  \multicolumn{1}{c}{\textbf{(deu)}} &  & \multicolumn{1}{c}{\textbf{(deu)}} &  & \multicolumn{3}{c}{\textbf{(bar)}} &  & \multicolumn{3}{c}{\textbf{(gsw)}} \\
\cmidrule(rl){6-8} \cmidrule(rl){10-12} \cmidrule(rl){18-20} \cmidrule(rl){22-24}
\textbf{Model} &  &  &  &  & \multicolumn{1}{c}{\textbf{deu ref}} &  & \multicolumn{1}{c}{\textbf{bar ref}} &  & \multicolumn{1}{c}{\textbf{deu ref}} &  & \multicolumn{1}{c}{\textbf{gsw ref}} &  &  &  &  &  &\multicolumn{1}{c}{\textbf{deu ref}} &  & \multicolumn{1}{c}{\textbf{bar ref}} &  & \multicolumn{1}{c}{\textbf{deu ref}} &  & \multicolumn{1}{c}{\textbf{gsw ref}} \\
\cmidrule(r){1-1} \cmidrule{2-12} \cmidrule{14-24}
\multicolumn{24}{@{}l@{}}{\textit{Word/character error rates per ASR model used in cascaded setups ($\downarrow$)}} \\
X.\ 300M DE & \cellcolor[HTML]{79C9A2}26.4{\textsubscript{25.3}} &  & \cellcolor[HTML]{83CDA9}31.5{\textsubscript{24.7}} &  & \cellcolor[HTML]{EBF7F1}86.3{\textsubscript{20.7}} &  & \cellcolor[HTML]{DBF0E6}77.9{\textsubscript{22.3}} &  & \cellcolor[HTML]{E5F4ED}83.0{\textsubscript{15.5}} &  & \cellcolor[HTML]{E5F4EC}82.7{\textsubscript{16.1}} &  & \cellcolor[HTML]{60BE90}5.9{\textsubscript{7.2}} &  & \cellcolor[HTML]{74C79E}11.9{\textsubscript{13.7}} &  & \cellcolor[HTML]{F7FBF9}51.0{\textsubscript{19.8}} &  & \cellcolor[HTML]{B5E1CB}31.3{\textsubscript{17.3}} &  & \cellcolor[HTML]{CCEADC}38.4{\textsubscript{14.8}} &  & \cellcolor[HTML]{AEDEC7}29.3{\textsubscript{12.6}} \\
X.\ 1B DE & \cellcolor[HTML]{73C69E}23.3{\textsubscript{26.2}} &  & \cellcolor[HTML]{7CCAA4}28.0{\textsubscript{24.5}} &  & \cellcolor[HTML]{E7F5EE}84.1{\textsubscript{20.8}} &  & \cellcolor[HTML]{D9EFE4}76.6{\textsubscript{21.8}} &  & \cellcolor[HTML]{E5F4ED}82.9{\textsubscript{16.5}} &  & \cellcolor[HTML]{E5F4ED}82.9{\textsubscript{16.8}} &  & \cellcolor[HTML]{5EBE8F}5.3{\textsubscript{7.0}} &  & \cellcolor[HTML]{70C59C}10.7{\textsubscript{13.3}} &  & \cellcolor[HTML]{F3FAF6}49.9{\textsubscript{20.1}} &  & \cellcolor[HTML]{B4E0CB}31.0{\textsubscript{18.0}} &  & \cellcolor[HTML]{CDEADC}38.4{\textsubscript{15.0}} &  & \cellcolor[HTML]{B1DFC8}30.0{\textsubscript{13.1}} \\
MMS 1B-all & \cellcolor[HTML]{6EC49A}20.5{\textsubscript{23.9}} &  & \cellcolor[HTML]{7ECBA5}28.9{\textsubscript{24.6}} &  & \cellcolor[HTML]{E4F4EC}82.5{\textsubscript{22.4}} &  & \cellcolor[HTML]{D6EEE2}75.1{\textsubscript{23.0}} &  & \cellcolor[HTML]{E1F3EA}81.0{\textsubscript{20.2}} &  & \cellcolor[HTML]{E1F3EA}80.9{\textsubscript{19.7}} &  & \cellcolor[HTML]{60BE90}5.7{\textsubscript{8.4}} &  & \cellcolor[HTML]{71C59C}11.0{\textsubscript{11.4}} &  & \cellcolor[HTML]{F4FAF7}50.2{\textsubscript{19.3}} &  & \cellcolor[HTML]{B7E2CD}32.0{\textsubscript{16.6}} &  & \cellcolor[HTML]{C7E8D8}36.6{\textsubscript{14.9}} &  & \cellcolor[HTML]{ADDEC6}29.0{\textsubscript{13.3}} \\
Whisper tiny & \cellcolor[HTML]{92D2B3}39.2{\textsubscript{34.4}} &  & \cellcolor[HTML]{98D5B7}42.6{\textsubscript{30.5}} &  & \cellcolor[HTML]{EEF8F3}87.4{\textsubscript{25.8}} &  & \cellcolor[HTML]{EAF6F0}85.4{\textsubscript{23.8}} &  & \cellcolor[HTML]{EAF6F0}85.7{\textsubscript{169.3}} &  & \cellcolor[HTML]{FFFFFF}96.3{\textsubscript{136.2}} &  & \cellcolor[HTML]{83CDA9}16.4{\textsubscript{16.7}} &  & \cellcolor[HTML]{85CDAA}16.9{\textsubscript{14.4}} &  & \cellcolor[HTML]{FFFFFF}53.4{\textsubscript{18.3}} &  & \cellcolor[HTML]{D1ECDF}39.7{\textsubscript{15.9}} &  & \cellcolor[HTML]{E8F5EF}46.6{\textsubscript{162.4}} &  & \cellcolor[HTML]{E1F3EA}44.6{\textsubscript{149.0}} \\
Whisper base & \cellcolor[HTML]{7ECAA5}28.9{\textsubscript{32.3}} &  & \cellcolor[HTML]{82CCA8}31.2{\textsubscript{28.7}} &  & \cellcolor[HTML]{E9F6F0}85.1{\textsubscript{30.0}} &  & \cellcolor[HTML]{E3F3EB}81.9{\textsubscript{28.6}} &  & \cellcolor[HTML]{C3E6D5}64.9{\textsubscript{49.2}} &  & \cellcolor[HTML]{EBF7F1}86.0{\textsubscript{54.8}} &  & \cellcolor[HTML]{73C69D}11.5{\textsubscript{14.3}} &  & \cellcolor[HTML]{76C7A0}12.5{\textsubscript{13.2}} &  & \cellcolor[HTML]{F8FCFA}51.4{\textsubscript{21.0}} &  & \cellcolor[HTML]{CBE9DA}37.8{\textsubscript{20.9}} &  & \cellcolor[HTML]{B6E1CC}31.7{\textsubscript{38.9}} &  & \cellcolor[HTML]{BDE4D1}33.7{\textsubscript{41.0}} \\
W.\ small & \cellcolor[HTML]{69C297}18.0{\textsubscript{24.1}} &  & \cellcolor[HTML]{67C195}16.6{\textsubscript{21.8}} &  & \cellcolor[HTML]{D7EFE3}75.6{\textsubscript{27.3}} &  & \cellcolor[HTML]{D1ECDF}72.5{\textsubscript{25.8}} &  & \cellcolor[HTML]{99D5B8}42.9{\textsubscript{25.5}} &  & \cellcolor[HTML]{DEF1E8}79.3{\textsubscript{20.2}} &  & \cellcolor[HTML]{79C8A1}13.3{\textsubscript{126.8}} &  & \cellcolor[HTML]{61BF91}6.1{\textsubscript{9.2}} &  & \cellcolor[HTML]{E7F5EE}46.2{\textsubscript{21.0}} &  & \cellcolor[HTML]{B6E1CC}31.6{\textsubscript{19.5}} &  & \cellcolor[HTML]{91D2B2}20.6{\textsubscript{17.5}} &  & \cellcolor[HTML]{ABDDC4}28.2{\textsubscript{14.3}} \\
W.\ medium & \cellcolor[HTML]{62BF92}14.4{\textsubscript{22.7}} &  & \cellcolor[HTML]{5CBD8D}11.0{\textsubscript{16.5}} &  & \cellcolor[HTML]{CFEBDE}71.6{\textsubscript{28.4}} &  & \cellcolor[HTML]{CDEADC}70.1{\textsubscript{29.5}} &  & \cellcolor[HTML]{80CBA6}29.9{\textsubscript{23.5}} &  & \cellcolor[HTML]{D9EFE5}76.8{\textsubscript{19.7}} &  & \cellcolor[HTML]{60BE90}5.7{\textsubscript{11.0}} &  & \cellcolor[HTML]{5ABC8C}4.1{\textsubscript{6.5}} &  & \cellcolor[HTML]{E7F5EE}46.5{\textsubscript{46.4}} &  & \cellcolor[HTML]{B7E2CD}31.9{\textsubscript{49.4}} &  & \cellcolor[HTML]{7FCBA5}15.0{\textsubscript{14.3}} &  & \cellcolor[HTML]{A7DBC1}27.1{\textsubscript{10.6}} \\
W.\ lg.-v3-tur. & \cellcolor[HTML]{62BF92}14.3{\textsubscript{23.9}} &  & \cellcolor[HTML]{58BB8B}9.0{\textsubscript{14.5}} &  & \cellcolor[HTML]{C9E9D9}68.0{\textsubscript{28.0}} &  & \cellcolor[HTML]{C7E8D8}67.0{\textsubscript{28.3}} &  & \cellcolor[HTML]{79C8A1}26.0{\textsubscript{22.1}} &  & \cellcolor[HTML]{D7EEE3}75.5{\textsubscript{20.1}} &  & \cellcolor[HTML]{5EBE8F}5.3{\textsubscript{10.1}} &  & \cellcolor[HTML]{57BB8A}3.2{\textsubscript{5.9}} &  & \cellcolor[HTML]{DAF0E5}42.5{\textsubscript{21.6}} &  & \cellcolor[HTML]{AADCC4}28.1{\textsubscript{17.1}} &  & \cellcolor[HTML]{79C8A1}13.3{\textsubscript{13.6}} &  & \cellcolor[HTML]{A3D9BE}25.8{\textsubscript{9.9}} \\
W.\ lg.-v3 & \cellcolor[HTML]{5DBD8E}11.8{\textsubscript{19.2}} &  & \cellcolor[HTML]{57BB8A}8.2{\textsubscript{13.8}} &  & \cellcolor[HTML]{C7E8D8}67.1{\textsubscript{28.4}} &  & \cellcolor[HTML]{C9E9D9}68.3{\textsubscript{28.4}} &  & \cellcolor[HTML]{75C79F}24.1{\textsubscript{21.8}} &  & \cellcolor[HTML]{D6EEE2}75.0{\textsubscript{19.7}} &  & \cellcolor[HTML]{5CBD8D}4.5{\textsubscript{8.0}} &  & \cellcolor[HTML]{57BB8A}3.0{\textsubscript{5.5}} &  & \cellcolor[HTML]{D9EFE5}42.2{\textsubscript{21.9}} &  & \cellcolor[HTML]{ADDDC5}28.8{\textsubscript{17.6}} &  & \cellcolor[HTML]{77C8A0}12.7{\textsubscript{13.5}} &  & \cellcolor[HTML]{A3DABF}26.1{\textsubscript{10.2}} \\
\cmidrule(r){1-1} \cmidrule{2-12} \cmidrule{14-24}
\multicolumn{24}{@{}l@{}}{\textit{Correlations between WER/CER and $\Delta$ cascaded--text}} \\
mBERT & \multicolumn{1}{c}{\cellcolor[HTML]{6DA3D4}--0.86} &  & \multicolumn{1}{c}{\cellcolor[HTML]{99BFE1}--0.75} &  & \multicolumn{1}{c}{\cellcolor[HTML]{6FA4D4}--0.86} &  & \multicolumn{1}{c}{\cellcolor[HTML]{649DD1}--0.89} &  & \multicolumn{1}{c}{\cellcolor[HTML]{4288C7}{\color[HTML]{CCCCCC}--0.97}} &  & \multicolumn{1}{c}{\cellcolor[HTML]{C1D8EC}--0.65} &  & \multicolumn{1}{c}{\cellcolor[HTML]{A4C6E4}--0.73} &  & \multicolumn{1}{c}{\cellcolor[HTML]{94BBDF}--0.77} &  & \multicolumn{1}{c}{\cellcolor[HTML]{5C98CF}{\color[HTML]{CCCCCC}--0.91}} &  & \multicolumn{1}{c}{\cellcolor[HTML]{82B0DA}--0.81} &  & \multicolumn{1}{c}{\cellcolor[HTML]{4E90CB}{\color[HTML]{CCCCCC}--0.94}} &  & \multicolumn{1}{c}{\cellcolor[HTML]{FFFFFF}--0.50} \\
mDeBERTa & \multicolumn{1}{c}{\cellcolor[HTML]{3F86C6}{\color[HTML]{CCCCCC}--0.98}} &  & \multicolumn{1}{c}{\cellcolor[HTML]{71A6D5}--0.85} &  & \multicolumn{1}{c}{\cellcolor[HTML]{6DA3D4}--0.86} &  & \multicolumn{1}{c}{\cellcolor[HTML]{5292CC}{\color[HTML]{CCCCCC}--0.93}} &  & \multicolumn{1}{c}{\cellcolor[HTML]{4389C8}{\color[HTML]{CCCCCC}--0.97}} &  & \multicolumn{1}{c}{\cellcolor[HTML]{BCD5EB}--0.67} &  & \multicolumn{1}{c}{\cellcolor[HTML]{A8C8E5}--0.72} &  & \multicolumn{1}{c}{\cellcolor[HTML]{6CA2D3}--0.87} &  & \multicolumn{1}{c}{\cellcolor[HTML]{5896CE}{\color[HTML]{CCCCCC}--0.92}} &  & \multicolumn{1}{c}{\cellcolor[HTML]{659ED1}--0.88} &  & \multicolumn{1}{c}{\cellcolor[HTML]{4E90CB}{\color[HTML]{CCCCCC}--0.94}} &  & \multicolumn{1}{c}{\cellcolor[HTML]{F8FBFD}--0.51} \\
XLM-R base & \multicolumn{1}{c}{\cellcolor[HTML]{9CC1E2}--0.74} &  & \multicolumn{1}{c}{\cellcolor[HTML]{93BBDF}--0.77} &  & \multicolumn{1}{c}{\cellcolor[HTML]{73A7D6}--0.85} &  & \multicolumn{1}{c}{\cellcolor[HTML]{80AFD9}--0.82} &  & \multicolumn{1}{c}{\cellcolor[HTML]{478BC8}{\color[HTML]{CCCCCC}--0.96}} &  & \multicolumn{1}{c}{\cellcolor[HTML]{C2D9ED}--0.65} &  & \multicolumn{1}{c}{\cellcolor[HTML]{8BB6DC}--0.79} &  & \multicolumn{1}{c}{\cellcolor[HTML]{8FB8DE}--0.78} &  & \multicolumn{1}{c}{\cellcolor[HTML]{71A6D5}--0.85} &  & \multicolumn{1}{c}{\cellcolor[HTML]{BED6EC}--0.66} &  & \multicolumn{1}{c}{\cellcolor[HTML]{5292CC}{\color[HTML]{CCCCCC}--0.93}} &  & \multicolumn{1}{c}{\cellcolor[HTML]{FEFEFE}--0.50} \\
XLM-R large & \multicolumn{1}{c}{\cellcolor[HTML]{76A8D6}--0.84} &  & \multicolumn{1}{c}{\cellcolor[HTML]{6BA2D3}--0.87} &  & \multicolumn{1}{c}{\cellcolor[HTML]{6AA1D3}--0.87} &  & \multicolumn{1}{c}{\cellcolor[HTML]{78AAD7}--0.84} &  & \multicolumn{1}{c}{\cellcolor[HTML]{3D85C6}{\color[HTML]{CCCCCC}--0.99}} &  & \multicolumn{1}{c}{\cellcolor[HTML]{A4C5E4}--0.73} &  & \multicolumn{1}{c}{\cellcolor[HTML]{84B1DA}--0.81} &  & \multicolumn{1}{c}{\cellcolor[HTML]{69A0D2}--0.87} &  & \multicolumn{1}{c}{\cellcolor[HTML]{669FD2}--0.88} &  & \multicolumn{1}{c}{\cellcolor[HTML]{BBD4EB}--0.67} &  & \multicolumn{1}{c}{\cellcolor[HTML]{4489C8}{\color[HTML]{CCCCCC}--0.97}} &  & \multicolumn{1}{c}{\cellcolor[HTML]{E0EBF5}--0.58} \\
\bottomrule
\end{tabular}
}
\caption{\textbf{The better the ASR performance, the better the cascaded systems do compared to the text-only systems.}
\textbf{Top:} Word/character error rates of ASR models (in~\%), averaged over sentences for each test set (standard deviations in subscripts). 
Lower (darker) is better.
\textbf{Bottom:} Correlations between error rates and performance differences between cascaded and text-only setups (Pearson's \textit{r}, all $p$-values $<0.001$ except for the right-most column, where $0.001<p<0.01$).
For the performance differences 
(cascaded minus text-only),
see Table~\ref{tab:deltas} (right).
}
\label{tab:asr}
\end{table*}

For the speech-only models, the models trained/fine-tuned for German or multilingual ASR outperform their non-fine-tuned counterparts (at least when matched by model size), both on the German and dialectal data.
This is consistent with prior works finding ASR to be a useful pre-training task for downstream tasks related to understanding the content of an utterance \cite{he-garner-2023-interpreter, schmidt-etal-2025-fleurs-slu}.

\textbf{Different setups work best for the German and dialectal data.}
The text-only models achieve the best performances on the German data, followed by the cascaded models and the ASR-tuned speech models. 
The non-ASR speech models perform worst.
For the Bavarian data however, the ASR-tuned speech models (and the one text model pretrained on Bavarian data, mBERT), perform best. The accuracy of the cascaded models varies a lot, depending on both the ASR model and the text model.
Similarly, for the Swiss German data, the cascaded systems contain both the best and the worst results -- depending on the ASR model included -- even though on average their performance is similar to that of the text-only models.
This trend is consistent across the eight different dialects in Swissdial (Table~\ref{tab:dialects-topics} in Appendix~\ref{sec:appendix-deaggregated-results}).

In most setups, bigger models outperform smaller models, with the exception of \xlmr{} large (text-only and cascaded) for the MASSIVE test data, even though it outperforms its smaller counterpart in the cross-dataset evaluation on xSID.

\section{Analysis}
\label{sec:analysis}

\begin{table*}
\centering
\setlength{\tabcolsep}{1pt} %
\adjustbox{max width=\textwidth}{%
\begin{tabular}{@{}l@{\hspace{5pt}}rrrrrr@{\hspace{5pt}}rrrrrr@{\hspace{5pt}}rrrrrr@{\hspace{5pt}}r@{}}
\toprule
& \multicolumn{5}{c}{$\Delta$ \textbf{speech--cascaded}} &  & \multicolumn{5}{c}{$\Delta$ \textbf{speech--text}} &  & \multicolumn{7}{c}{$\Delta$ \textbf{cascaded--text}} \\
\cmidrule(rl){2-6} \cmidrule(rl){8-12} \cmidrule(rl){14-20}
& \multicolumn{1}{c}{\textbf{MAS.}} &  & \multicolumn{1}{c}{\textbf{xSID}} &  & \multicolumn{1}{c}{\textbf{xSID}} &  & \multicolumn{1}{c}{\textbf{MAS.}} &  & \multicolumn{1}{c}{\textbf{xSID}} &  & \multicolumn{1}{c}{\textbf{xSID}}  &  & \multicolumn{1}{c}{\textbf{MAS.}} &  & \multicolumn{1}{c}{\textbf{xSID}} &  & \multicolumn{1}{c}{\textbf{xSID}} &  & \multicolumn{1}{c}{\textbf{Swissd.}} \\
\textbf{Model} & \multicolumn{1}{c}{\textbf{deu}} &  & \multicolumn{1}{c}{\textbf{deu}} &  & \multicolumn{1}{c}{\textbf{bar}} &  & \multicolumn{1}{c}{\textbf{deu}} &  & \multicolumn{1}{c}{\textbf{deu}} &  & \multicolumn{1}{c}{\textbf{bar}} &  & \multicolumn{1}{c}{\textbf{deu}} &  & \multicolumn{1}{c}{\textbf{deu}} &  & \multicolumn{1}{c}{\textbf{bar}} &  & \multicolumn{1}{c}{\textbf{gsw}} \\
\cmidrule(r){1-1} \cmidrule{2-6} \cmidrule{8-12} \cmidrule{14-20}
mHuBERT &  &  &  &  &  &  &  \cellcolor[HTML]{B66E91}--18.0\textsubscript{1.0} &  & \cellcolor[HTML]{A64D79}{\color[HTML]{CCCCCC}--22.0\textsubscript{3.6}} &  & \cellcolor[HTML]{EAD5DF}--5.2\textsubscript{4.5} \\
XLS-R 300M &  &  &  &  &  &  & \cellcolor[HTML]{BA7698}--17.0\textsubscript{1.0} &  & \cellcolor[HTML]{B06188}{\color[HTML]{CCCCCC}--19.6\textsubscript{3.6}} &  & \cellcolor[HTML]{ECDAE3}--4.5\textsubscript{4.5}\\
XLS-R 300M DE & {\cellcolor[HTML]{FDF6DF}+2.8\textsubscript{1.4}} &  & {\cellcolor[HTML]{FFFFFC}+0.3\textsubscript{3.9}} &  & {\cellcolor[HTML]{F1C232}+17.9\textsubscript{7.2}} &  & \cellcolor[HTML]{F6EDF1}--2.2\textsubscript{1.0} &  & \cellcolor[HTML]{E8D1DC}--5.7\textsubscript{3.6} &  & \cellcolor[HTML]{FAE7AD}+7.2\textsubscript{4.5} &  & \cellcolor[HTML]{EAD6E0}--5.1\textsubscript{1.1} &  & \cellcolor[HTML]{E6CEDA}--6.0\textsubscript{1.8} &  & \cellcolor[HTML]{D3A8BE}--10.7\textsubscript{2.8} &  & \cellcolor[HTML]{D0A1B8}--11.6\textsubscript{2.4} \\
XLS-R 1B DE &  &  &  &  &  &  &  &  &  &  &  &  & \cellcolor[HTML]{EEDEE6}--4.0\textsubscript{0.7} &  & \cellcolor[HTML]{F6EDF1}--2.2\textsubscript{1.4} &  & \cellcolor[HTML]{DAB5C7}--9.1\textsubscript{2.1} &  & \cellcolor[HTML]{D4A9BE}--10.6\textsubscript{2.2} \\
MMS 300M &  &  &  &  &  &  & \cellcolor[HTML]{C285A3}--15.1\textsubscript{1.0} &  & \cellcolor[HTML]{BF7F9F}--15.8\textsubscript{3.6} &  & \cellcolor[HTML]{FCF9FB}--0.6\textsubscript{4.5}\\
MMS 1B-all &  &  &  &  &  &  &  &  &  &  &  &  & \cellcolor[HTML]{EFDFE7}--3.9\textsubscript{0.7} &  & \cellcolor[HTML]{E6CEDA}--6.1\textsubscript{2.7} &  & \cellcolor[HTML]{D6AEC2}--10.0\textsubscript{0.6} &  & \cellcolor[HTML]{E5CBD7}--6.4\textsubscript{2.0} \\
Whisper tiny & {\cellcolor[HTML]{DCBACB}--8.5\textsubscript{2.4}} &  & {\cellcolor[HTML]{D7AFC2}--9.9\textsubscript{5.2}} &  & {\cellcolor[HTML]{FBECBF}+5.6\textsubscript{8.6}} &  & \cellcolor[HTML]{A74F7B}{\color[HTML]{CCCCCC}--21.8\textsubscript{1.0}} &  & \cellcolor[HTML]{A64D79}{\color[HTML]{CCCCCC}--22.1\textsubscript{3.6}} &  & \cellcolor[HTML]{D1A4BA}--11.2\textsubscript{4.5} &  & \cellcolor[HTML]{C994AE}--13.3\textsubscript{2.5} &  & \cellcolor[HTML]{CD9CB5}--12.2\textsubscript{3.1} &  & \cellcolor[HTML]{BB7798}--16.9\textsubscript{4.9} &  & \cellcolor[HTML]{DEBDCD}--8.1\textsubscript{2.8} \\
Whisper base & {\cellcolor[HTML]{FFFFFF}+0.0\textsubscript{1.3}} &  & {\cellcolor[HTML]{F5ECF0}--2.3\textsubscript{4.9}} &  & {\cellcolor[HTML]{F2C63E}+16.9\textsubscript{9.7}} &  & \cellcolor[HTML]{DAB6C8}--9.0\textsubscript{1.0} &  & \cellcolor[HTML]{D1A4BA}--11.3\textsubscript{3.6} &  & \cellcolor[HTML]{FCF0CC}+4.5\textsubscript{4.5} &  & \cellcolor[HTML]{DAB6C8}--9.0\textsubscript{1.2} &  & \cellcolor[HTML]{DAB6C8}--9.0\textsubscript{3.4} &  & \cellcolor[HTML]{CD9BB4}--12.4\textsubscript{5.8} &  & \cellcolor[HTML]{F9F3F6}--1.5\textsubscript{2.7} \\
Whisper small & {\cellcolor[HTML]{FFFEF9}+0.6\textsubscript{1.5}} &  & {\cellcolor[HTML]{FFFCF4}+1.0\textsubscript{3.8}} &  & {\cellcolor[HTML]{F6D779}+11.8\textsubscript{8.8}} &  & \cellcolor[HTML]{F2E5EB}--3.1\textsubscript{1.0} &  & \cellcolor[HTML]{F8F2F5}--1.5\textsubscript{3.6} &  & \cellcolor[HTML]{F7DC89}+10.4\textsubscript{4.5} &  & \cellcolor[HTML]{F0E1E8}--3.7\textsubscript{0.6} &  & \cellcolor[HTML]{F4EAEF}--2.5\textsubscript{1.8} &  & \cellcolor[HTML]{F9F3F6}--1.4\textsubscript{4.6} &  & \cellcolor[HTML]{FBEDC1}+5.5\textsubscript{2.5} \\
Whisper medium & {\cellcolor[HTML]{FEFDFD}--0.2\textsubscript{1.0}} &  & {\cellcolor[HTML]{FAF6F8}--1.0\textsubscript{3.8}} &  & {\cellcolor[HTML]{FBECBF}+5.6\textsubscript{7.5}} &  & \cellcolor[HTML]{F6EEF2}--2.0\textsubscript{1.0} &  & \cellcolor[HTML]{F3E8EE}--2.8\textsubscript{3.6} &  & \cellcolor[HTML]{FDF3D4}+3.8\textsubscript{4.5} &  & \cellcolor[HTML]{F7F0F3}--1.8\textsubscript{0.3} &  & \cellcolor[HTML]{F7F0F4}--1.8\textsubscript{1.0} &  & \cellcolor[HTML]{F7F0F4}--1.8\textsubscript{3.5} &  & \cellcolor[HTML]{F9E4A4}+8.0\textsubscript{2.0} \\
Whisper large-v3-\rlap{turbo} &  &  &  &  &  &  &  &  &  &  &  &  & \cellcolor[HTML]{F9F4F7}--1.2\textsubscript{0.3} &  & \cellcolor[HTML]{FBF7F9}--1.0\textsubscript{0.8} &  & \cellcolor[HTML]{FDF6DE}+2.9\textsubscript{2.3} &  & \cellcolor[HTML]{F8E198}+9.0\textsubscript{1.7} \\
Whisper large-v3 & {\cellcolor[HTML]{FBF8FA}--0.8\textsubscript{0.9}} &  & {\cellcolor[HTML]{FEF9EA}+1.9\textsubscript{4.0}} &  & {\cellcolor[HTML]{F7DA80}+11.1\textsubscript{6.0}} &  & \cellcolor[HTML]{F8F1F4}--1.7\textsubscript{1.0} &  & \cellcolor[HTML]{FFFCF5}+0.9\textsubscript{3.6} &  & \cellcolor[HTML]{F5D163}+13.7\textsubscript{4.5} &  & \cellcolor[HTML]{FBF7F9}--0.9\textsubscript{0.3} &  & \cellcolor[HTML]{FBF7F9}--1.0\textsubscript{0.7} &  & \cellcolor[HTML]{FDF7E2}+2.6\textsubscript{1.8} &  & \cellcolor[HTML]{F9E19A}+8.9\textsubscript{1.7} \\
\bottomrule
\end{tabular}
}
\caption{\textbf{Using text data works best for the standard-language test sets, while speech data tends to yield better results for the dialect.}
Accuracy differences between setups, in percentage points.
We use the results that are already averaged across random seeds for calculating differences. 
$\Delta$~speech--cascaded: Each speech-only model is compared to all cascaded systems that involve the given speech model as ASR model.
$\Delta$~speech--text: Each speech-only model is compared to all text-only models.
$\Delta$~cascaded--text: Each cascaded system is compared to the text-only system involving the same text model.
Standard deviations are in subscripts.
}
\label{tab:deltas}
\end{table*}

\subsection{Cascaded vs.\ text-only: How well do the text models generalize to transcribed data?}
\label{sec:analysis-cascaded-text}

\textbf{The quality of the ASR system is an important factor in the cascaded setup.}
Table~\ref{tab:asr} shows the word and character error rates (WER, CER) of the different ASR models for the different test sets, as well as the correlations between the error rates and the performance differences between the cascaded and text-only setups.

For the dialectal test sets, we compare the ASR hypotheses to both the dialectal and the German references.
We compare them to the dialectal references as they directly correspond to the audio recordings.
However, the ASR models -- especially the Whisper ones -- tend to produce more normalized, German-like output \cite{dolev-etal-2024-whisper, gorisch-schmidt-2024-evaluating, blaschke-etal-2025-multi-dialectal}.
We therefore also compare their outputs to the German text data, although there can be lexical and syntactic differences between the German and dialectal versions.
In the case of xSID, the German and Bavarian splits were independently translated from English \cite{winkler-etal-2024-slot}, 
leading to sentence pairs that are parallel on a sentence level but not necessarily on a word level \cite{winkler-etal-2024-slot, kruckl-etal-2025-improving}.

For the German test sets, the closer the ASR hypotheses are to the gold text, the smaller the performance gap between the text-only and cascaded setups (Pearson's~$r$ for error rate and cascaded/text-only gap between $-0.72$ and $-0.98$). 
However, even the cascaded systems with the lowest error rates do not reach the accuracies of the corresponding text-only versions (Table~\ref{tab:deltas}, right).
For the dialectal test sets, ASR performance is also strongly correlated with the results of the cascaded systems.
However, the cascaded systems with the best ASR models outperform their text-only counterparts. %

For the Swiss German data, the correlations between the error rate and the performance gap are very strong when comparing the ASR hypotheses to the German references ($-0.94<r<-0.99$): if the ASR output is normalized German (i.e., the WER/CER is low with respect to the German reference), the text models can process it well.
For the Bavarian data where there might be more syntactic variation between the German and Bavarian text versions, the correlations are not quite as strong as for the Swiss German data, but they are high both when considering the German \textit{and} the Bavarian references ($-0.85<r<-0.92$ and $-0.66<r<-0.93$).

For the dialectal test sets, using ASR presents on the one hand the chance of acting as dialect-to-standard normalization, and on the other hand the risk of the ASR model producing nonsensical outputs.
However, the automatic ASR quality measures might give a somewhat pessimistic impression with respect to sentence classification tasks.
In our manual evaluation of ASR hypotheses for 125 German and Bavarian sentences from the xSID test set by four of the ASR models, we find that keywords related to the intent are often still retained in the transcriptions even when the sentence has a (somewhat) different meaning and/or is not in fluent and orthographically correct German~(\S\ref{sec:appendix-asr-manual}).

In the case of Swissdial, the best cascaded models perform on the Swiss German data on par with the German text-only evaluations, suggesting successful normalization.
For the cascaded intent classification systems, gaps of up to 19.1\,pp.\ between the German and Bavarian data remain even for systems with the best-performing ASR models (Table~\ref{tab:results-overview}).
For the cascaded systems with the worst ASR models, the performance is worse than processing the written dialect data.

\subsection{Speech-only vs.\ cascaded: How to best deal with spoken inputs?}
\label{sec:analysis-speech-cascaded}

There are two approaches for handling spoken input data: training a speech-based system (speech-only) or training a text-based system and transcribing the spoken input data.
\textbf{For the German test data, this choice barely makes a difference.}
When comparing the speech-only and cascaded systems involving the same speech/ASR models, the performance gaps are close to zero (Table~\ref{tab:deltas}, left), with the exception of the smallest model (Whisper tiny).
\textbf{For Bavarian however, the speech-only setup outperforms the cascaded setup} across the board (performance gains of 
$5.6-17.9$\,pp.).

\subsection{Speech-only vs.\ text-only: Which modality is preferable?}
\label{sec:analysis-speech-text}

The text-only models nearly always outperform the speech-only models for German (Table~\ref{tab:deltas}, middle).
For the text and speech models of comparable sizes, the performance gaps are between $-23.8$ and $+1.6$\,pp.\ ($\sim$90\,M parameters) and between $-23.8$ and $-7.0$\,pp.\ ($\sim$300\,M parameters).\footnote{%
mHuB.--mBERT: $-23.8$,
Whisper\textsubscript{sm}--\xlmr{}\textsubscript{base}: $+1.6$,
\xlsr{}\textsubscript{300M}--\xlmr{}\textsubscript{lg}: $-23.8$,
Whisper\textsubscript{med}--\xlmr{}\textsubscript{lg}: $-7.0$.
}
The gaps are smaller for ASR speech models (Table~\ref{tab:results-overview}).

The results for the Bavarian data are different:
\textbf{the speech models are generally more robust in standard-to-dialect transfer than the text models.}
Most speech models outperform the text models that were not also pretrained on Bavarian data (i.e., all text models except mBERT).
The performance gaps are between $-3.3$ and $+14.8$\,pp.\ ($\sim$90M parameters, ignoring mBERT) and between $-4.7$ and $+7.0$\,pp.\ ($\sim$300M parameters).\footnote{%
mHuB.--\xlmr{}\textsubscript{base}: $-3.3$ (if including mBERT: mHuB.--mBERT: $-11.3$),
Whisper\textsubscript{small}--mDeB.: $+14.8$,
\xlsr{}\textsubscript{300M}--\xlmr{}\textsubscript{lg}: $-4.7$,
\xlsr{}\textsubscript{300M}--\xlmr{}\textsubscript{lg}: $+7.0$.
}
Except for the speech models whose performance gap is particularly large on the German data (accuracy differences of more than 15\,pp.), the speech models outperform the text models on the Bavarian data.

\section{Discussion}

\paragraph{Results for German vs.\ dialects}

The text-only approach is generally best for the German data, followed by the cascaded setup and lastly the spoken setup.
This is consistent with prior work finding that text-only models typically outperform speech-only models \cite{koudounas-etal-2023-italic, lee-etal-2024-speech-massive,  schmidt-etal-2025-fleurs-slu} and cascaded models \cite{lee-etal-2018-spoken, bastianelli-etal-2020-slurp, schmidt-etal-2025-fleurs-slu}, and that -- especially in cross-lingual transfer settings -- cascaded models often outperform speech-only setups \cite{lee-etal-2024-speech-massive, keller-glavas-2025-speechtaxi, schmidt-etal-2025-fleurs-slu}.

In standard-to-dialect transfer settings, however, the speech-only models perform best, and cascaded systems yield very heterogeneous results.
This is more similar to the finding by \citet{keller-glavas-2025-speechtaxi} that speech-only models outperform cascaded systems in monolingual settings where the language is not supported by the ASR component of the cascaded model.
\citet{rajaa2022skits2i} find that speech-only models also outperform cascaded ones on non-native Indian English. 
This might be due to potentially poor ASR performance on non-native accents (cf. \citealp{sanabria-etal-2023-edinburgh}).

\paragraph{Robustness and input representations}
The speech models tend to be more robust towards dialectal variation than the text models in our experiments~(\S\ref{sec:analysis-speech-text}).
This might be due to the speech models processing continuous signals from the audio data as their input without relying on pre-defined tokens.
While the dialects in our study also differ from German in lexical and morphosyntactic regards, the chief differences are on the phonological and phonetic level, and prior research has shown that multilingual speech models encode phonetically similar inputs similarly \cite{choi-etal-2024-self-supervised, shim-etal-2025-languages}.

\paragraph{Designing datasets for speech-related tasks}
Creating written datasets for tasks like intent detection makes sense for high-resource languages where ASR systems work well, given that it is easier to collect written data and cascaded models perform on par with speech-only models~(\S\ref{sec:analysis-speech-cascaded}).
However, for language varieties that are primarily spoken and not supported well by text models, collecting spoken data might be an especially promising avenue for creating realistic datasets (cf.\ \citealp{scharenborg-etal-2020-speech, chrupala-2023-putting}).

\section{Conclusion}

Insights about processing standard languages do not necessarily generalize to even closely related dialects, as speech-only and cascaded systems can be more promising on dialect data than their results on standard-language data might indicate.
We hope that our work on intent and topic classification inspires more research on expanding the burgeoning field of dialect NLP to focus on processing speech data beyond ASR.

\section*{Limitations}

\paragraph{Language varieties}
Our experiments involve a limited set of languages, dialects, and tasks, as these experiments require datasets that \textit{(i)}~have evaluation data that are parallel in a standard language and a related non-standard dialect, \textit{(ii)}~have parallel written and spoken evaluation data, \textit{(iii)}~have parallel written and spoken training data, and \textit{(iv)}~are labelled for a content classification task.
To the best of our knowledge, we create the first dataset that fulfills all four conditions, and use all available datasets that fulfill conditions \textit{i, ii}, and \textit{iv}.
We hope our work encourages others to release more such datasets in additional languages.

\paragraph{Cascaded setup}
For the cascaded setup, we focus on models trained on gold-standard text data rather than on ASR transcriptions.
This implementation matches that of prior work \cite[e.g.,][]{wang-etal-2023-whislu, lee-etal-2024-speech-massive}.
We do this in order to keep computation feasible while being able to compare a large amount of ASR models, as we only need to train 12~models this way (four text models on three random seeds each),
instead of training 108 models (nine ASR models $\times$ four text models $\times$ three seeds).
To mitigate this limitation, we additionally train a subset of the possible cascaded configurations on automatically transcribed training data (three ASR models, two text models, three seeds).
While the exact results differ from the systems trained on gold-standard text data (some of the cascaded systems trained on ASR data are stronger, but others weaker, than the cascaded systems trained on gold text), the transfer trends are the same when comparing the cascaded systems to the text-only and speech-only systems~(\S\ref{sec:appendix-cascaded-additional}).

\paragraph{Models}
We use encoder models because of their well-suitedness for classification tasks and to have more control over their task-specific training data~(\S\ref{sec:models}).
While follow-up work on instruction-tuned text/audio LLMs would be interesting, we intentionally do not include them in our experiments.
We cannot directly compare the performance of instruction-tuned text vs.\ audio LLMs -- unlike our experiments here where we control the fine-tuning data, we would not know if the differences in the model results are due to differences in the internal representations or due to differences in possible classification-related instruction-tuning data.

We use models that we deem representative of typical text- or speech-based encoders.
However, not all models use the same kinds of input representations.
E.g., a recently released speech model, OpusLM \cite{tian-etal-2025-opuslm}, uses audio tokens rather than continuous representations, and pixel-based models \cite{rust-etal-2023-language, munoz-ortiz-etal-2025-evaluating} use continuous input representations for text.
We also do not use any multimodal models that are trained to minimize representation differences between parallel text and speech input (e.g., \citealp{duquenne-etal-2023-sonar}).

The pre-training datasets of the speech models are unfortunately not openly accessible, but the metadata indicate no dialectal training data. 
The poor performance of all ASR-capable speech models on Bavarian ASR (Table~\ref{tab:asr} and \S\ref{sec:appendix-asr-examples}) indicates that it is unlikely these models were exposed to much Bavarian data, despite performing quite strongly on Bavarian in the speech-only setup. 

\section*{Ethical Considerations}
All recordings/annotations were carried out by a speaker/\allowbreak{}annotator who was employed and paid according to national standards and who agreed to share their recordings/annotations for research use.
We share the audio recordings for research on processing spoken language data, but do not permit their use in the context of speech synthesis or voice cloning.

\section*{Acknowledgements}

We thank Simone Ciciliano, Alex Fraser, and Andreas Säuberli for helpful discussions.
We also thank Siyao Peng, Ryan Soh-Eun Shim, and Shijia Zhou for their feedback on a draft of this paper.
This research is supported by European Research Council (ERC) Consolidator Grant DIALECT 101043235.

\FloatBarrier
\bibliography{lib, arxiv}

\begin{thebibliography}{75}
\providecommand{\natexlab}[1]{#1}

\bibitem[{Adelani et~al.(2024)Adelani, Liu, Shen, Vassilyev, Alabi, Mao, Gao, and Lee}]{adelani-etal-2024-sib}
David~Ifeoluwa Adelani, Hannah Liu, Xiaoyu Shen, Nikita Vassilyev, Jesujoba~O. Alabi, Yanke Mao, Haonan Gao, and En-Shiun~Annie Lee. 2024.
\newblock \href {https://doi.org/10.18653/v1/2024.eacl-long.14} {{SIB}-200: A simple, inclusive, and big evaluation dataset for topic classification in 200+ languages and dialects}.
\newblock In \emph{Proceedings of the 18th Conference of the European Chapter of the Association for Computational Linguistics (Volume 1: Long Papers)}, pages 226--245, St. Julian{'}s, Malta. Association for Computational Linguistics.

\bibitem[{Aepli et~al.(2023)Aepli, {\c{C}}{\"o}ltekin, Van Der~Goot, Jauhiainen, Kazzaz, Ljube{\v{s}}i{\'c}, North, Plank, Scherrer, and Zampieri}]{aepli-etal-2023-findings}
No{\"e}mi Aepli, {\c{C}}a{\u{g}}r{\i} {\c{C}}{\"o}ltekin, Rob Van Der~Goot, Tommi Jauhiainen, Mourhaf Kazzaz, Nikola Ljube{\v{s}}i{\'c}, Kai North, Barbara Plank, Yves Scherrer, and Marcos Zampieri. 2023.
\newblock \href {https://aclanthology.org/2023.vardial-1.25} {Findings of the {V}ar{D}ial evaluation campaign 2023}.
\newblock In \emph{Tenth Workshop on NLP for Similar Languages, Varieties and Dialects (VarDial 2023)}, pages 251--261, Dubrovnik, Croatia. Association for Computational Linguistics.

\bibitem[{Agnew et~al.(2024)Agnew, Barnett, Chu, Hong, Feffer, Netzorg, Jiang, Awumey, and Das}]{agnew2024sound}
William Agnew, Julia Barnett, Annie Chu, Rachel Hong, Michael Feffer, Robin Netzorg, Harry~H. Jiang, Ezra Awumey, and Sauvik Das. 2024.
\newblock \href {https://arxiv.org/abs/2410.13114} {Sound check: Auditing audio datasets}.
\newblock \emph{Preprint}, arXiv:2410.13114.

\bibitem[{Ahia et~al.(2024)Ahia, Aremu, Abagyan, Gonen, Adelani, Abolade, Smith, and Tsvetkov}]{ahia-etal-2024-voices}
Orevaoghene Ahia, Anuoluwapo Aremu, Diana Abagyan, Hila Gonen, David~Ifeoluwa Adelani, Daud Abolade, Noah~A. Smith, and Yulia Tsvetkov. 2024.
\newblock \href {https://doi.org/10.18653/v1/2024.emnlp-main.251} {Voices unheard: {NLP} resources and models for {Y}or{\`u}b{\'a} regional dialects}.
\newblock In \emph{Proceedings of the 2024 Conference on Empirical Methods in Natural Language Processing}, pages 4392--4409, Miami, Florida, USA. Association for Computational Linguistics.

\bibitem[{Amooie et~al.(2025)Amooie, De~Vries, Hao, Dijkstra, Coler, and Wieling}]{amooie-etal-2025-enhancing}
Reihaneh Amooie, Wietse De~Vries, Yun Hao, Jelske Dijkstra, Matt Coler, and Martijn Wieling. 2025.
\newblock \href {https://doi.org/10.1109/ICASSP49660.2025.10889692} {Enhancing standard and dialectal {Frisian ASR}: Multilingual fine-tuning and language identification for improved low-resource performance}.
\newblock In \emph{2025 IEEE International Conference on Acoustics, Speech and Signal Processing (ICASSP)}.

\bibitem[{Arora et~al.(2024)Arora, Futami, Jung, Peng, Sharma, Kashiwagi, Tsunoo, Livescu, and Watanabe}]{arora-etal-2024-universlu}
Siddhant Arora, Hayato Futami, Jee-weon Jung, Yifan Peng, Roshan Sharma, Yosuke Kashiwagi, Emiru Tsunoo, Karen Livescu, and Shinji Watanabe. 2024.
\newblock \href {https://doi.org/10.18653/v1/2024.naacl-long.151} {{U}niver{SLU}: Universal spoken language understanding for diverse tasks with natural language instructions}.
\newblock In \emph{Proceedings of the 2024 Conference of the North American Chapter of the Association for Computational Linguistics: Human Language Technologies (Volume 1: Long Papers)}, pages 2754--2774, Mexico City, Mexico. Association for Computational Linguistics.

\bibitem[{Artemova et~al.(2024)Artemova, Blaschke, and Plank}]{artemova-etal-2024-exploring}
Ekaterina Artemova, Verena Blaschke, and Barbara Plank. 2024.
\newblock \href {https://doi.org/10.18653/v1/2024.eacl-long.28} {Exploring the robustness of task-oriented dialogue systems for colloquial {G}erman varieties}.
\newblock In \emph{Proceedings of the 18th Conference of the European Chapter of the Association for Computational Linguistics (Volume 1: Long Papers)}, pages 445--468, St. Julian{'}s, Malta. Association for Computational Linguistics.

\bibitem[{Babu et~al.(2022)Babu, Wang, Tjandra, Lakhotia, Xu, Goyal, Singh, {von Platen}, Saraf, Pino, Baevski, Conneau, and Auli}]{babu-etal-2022-xlsr}
Arun Babu, Changhan Wang, Andros Tjandra, Kushal Lakhotia, Qiantong Xu, Naman Goyal, Kritika Singh, Patrick {von Platen}, Yatharth Saraf, Juan Pino, Alexei Baevski, Alexis Conneau, and Michael Auli. 2022.
\newblock \href {https://www.isca-archive.org/interspeech_2022/babu22_interspeech.html} {{XLS-R}: Self-supervised cross-lingual speech representation learning at scale}.
\newblock In \emph{Interspeech 2022}, pages 2278--2282.

\bibitem[{Baevski et~al.(2020)Baevski, Zhou, Mohamed, and Auli}]{baevski-etal-2020-wav2vec2}
Alexei Baevski, Yuhao Zhou, Abdelrahman Mohamed, and Michael Auli. 2020.
\newblock wav2vec 2.0: A framework for self-supervised learning of speech representations.
\newblock In \emph{Advances in Neural Information Processing Systems}, pages 12449--12460.

\bibitem[{Bastianelli et~al.(2020)Bastianelli, Vanzo, Swietojanski, and Rieser}]{bastianelli-etal-2020-slurp}
Emanuele Bastianelli, Andrea Vanzo, Pawel Swietojanski, and Verena Rieser. 2020.
\newblock \href {https://doi.org/10.18653/v1/2020.emnlp-main.588} {{SLURP}: A spoken language understanding resource package}.
\newblock In \emph{Proceedings of the 2020 Conference on Empirical Methods in Natural Language Processing (EMNLP)}, pages 7252--7262, Online. Association for Computational Linguistics.

\bibitem[{Beringer et~al.(1998)Beringer, Schiel, and Regel-Brietzmann}]{beringer-etal-1998-german}
Nicole Beringer, Florian Schiel, and Peter Regel-Brietzmann. 1998.
\newblock \href {https://doi.org/10.21437/ICSLP.1998-201} {{German} regional variants -- a problem for automatic speech recognition?}
\newblock In \emph{5th International Conference on Spoken Language Processing (ICSLP 1998)}, page paper 0201.

\bibitem[{Blaschke et~al.(2025{\natexlab{a}})Blaschke, K{\"o}rner, and Plank}]{blaschke-etal-2025-add}
Verena Blaschke, Felicia K{\"o}rner, and Barbara Plank. 2025{\natexlab{a}}.
\newblock \href {https://aclanthology.org/2025.vardial-1.14/} {Add noise, tasks, or layers? {M}ai{NLP} at the {V}ar{D}ial 2025 shared task on {N}orwegian dialectal slot and intent detection}.
\newblock In \emph{Proceedings of the 12th Workshop on NLP for Similar Languages, Varieties and Dialects}, pages 182--199, Abu Dhabi, UAE. Association for Computational Linguistics.

\bibitem[{Blaschke et~al.(2024)Blaschke, Purschke, Schuetze, and Plank}]{blaschke-etal-2024-dialect}
Verena Blaschke, Christoph Purschke, Hinrich Schuetze, and Barbara Plank. 2024.
\newblock \href {https://doi.org/10.18653/v1/2024.acl-short.74} {What do dialect speakers want? {A} survey of attitudes towards language technology for {G}erman dialects}.
\newblock In \emph{Proceedings of the 62nd Annual Meeting of the Association for Computational Linguistics (Volume 2: Short Papers)}, pages 823--841, Bangkok, Thailand. Association for Computational Linguistics.

\bibitem[{Blaschke et~al.(2023)Blaschke, Sch{\"u}tze, and Plank}]{blaschke-etal-2023-manipulating}
Verena Blaschke, Hinrich Sch{\"u}tze, and Barbara Plank. 2023.
\newblock \href {https://doi.org/10.18653/v1/2023.vardial-1.5} {Does manipulating tokenization aid cross-lingual transfer? a study on {POS} tagging for non-standardized languages}.
\newblock In \emph{Tenth Workshop on NLP for Similar Languages, Varieties and Dialects (VarDial 2023)}, pages 40--54, Dubrovnik, Croatia. Association for Computational Linguistics.

\bibitem[{Blaschke et~al.(2025{\natexlab{b}})Blaschke, Winkler, Förster, Wenger-Glemser, and Plank}]{blaschke-etal-2025-multi-dialectal}
Verena Blaschke, Miriam Winkler, Constantin Förster, Gabriele Wenger-Glemser, and Barbara Plank. 2025{\natexlab{b}}.
\newblock \href {https://doi.org/10.21437/Interspeech.2025-318} {A multi-dialectal dataset for {German} dialect {ASR} and dialect-to-standard speech translation}.
\newblock In \emph{Interspeech 2025}, pages 913--917.

\bibitem[{Choi et~al.(2024)Choi, Pasad, Nakamura, Fukayama, Livescu, and Watanabe}]{choi-etal-2024-self-supervised}
Kwanghee Choi, Ankita Pasad, Tomohiko Nakamura, Satoru Fukayama, Karen Livescu, and Shinji Watanabe. 2024.
\newblock \href {https://doi.org/10.21437/Interspeech.2024-1157} {Self-supervised speech representations are more phonetic than semantic}.
\newblock In \emph{Interspeech 2024}, pages 4578--4582.

\bibitem[{Chrupa{\l}a(2023)}]{chrupala-2023-putting}
Grzegorz Chrupa{\l}a. 2023.
\newblock \href {https://doi.org/10.18653/v1/2023.findings-acl.495} {Putting natural in {N}atural {L}anguage {P}rocessing}.
\newblock In \emph{Findings of the Association for Computational Linguistics: ACL 2023}, pages 7820--7827, Toronto, Canada. Association for Computational Linguistics.

\bibitem[{Conneau et~al.(2020)Conneau, Khandelwal, Goyal, Chaudhary, Wenzek, Guzm{\'a}n, Grave, Ott, Zettlemoyer, and Stoyanov}]{conneau-etal-2020-unsupervised}
Alexis Conneau, Kartikay Khandelwal, Naman Goyal, Vishrav Chaudhary, Guillaume Wenzek, Francisco Guzm{\'a}n, Edouard Grave, Myle Ott, Luke Zettlemoyer, and Veselin Stoyanov. 2020.
\newblock \href {https://doi.org/10.18653/v1/2020.acl-main.747} {Unsupervised cross-lingual representation learning at scale}.
\newblock In \emph{Proceedings of the 58th Annual Meeting of the Association for Computational Linguistics}, pages 8440--8451, Online. Association for Computational Linguistics.

\bibitem[{Coucke et~al.(2018)Coucke, Saade, Ball, Bluche, Caulier, Leroy, Doumouro, Gisselbrecht, Caltagirone, Lavril, Primet, and Dureau}]{coucke2018snips}
Alice Coucke, Alaa Saade, Adrien Ball, Théodore Bluche, Alexandre Caulier, David Leroy, Clément Doumouro, Thibault Gisselbrecht, Francesco Caltagirone, Thibaut Lavril, Maël Primet, and Joseph Dureau. 2018.
\newblock \href {https://arxiv.org/abs/1805.10190} {Snips voice platform: An embedded spoken language understanding system for private-by-design voice interfaces}.
\newblock \emph{Preprint}, arXiv:1805.10190.

\bibitem[{Devlin et~al.(2019)Devlin, Chang, Lee, and Toutanova}]{devlin-etal-2019-bert}
Jacob Devlin, Ming-Wei Chang, Kenton Lee, and Kristina Toutanova. 2019.
\newblock \href {https://doi.org/10.18653/v1/N19-1423} {{BERT}: Pre-training of deep bidirectional transformers for language understanding}.
\newblock In \emph{Proceedings of the 2019 Conference of the North {A}merican Chapter of the Association for Computational Linguistics: Human Language Technologies, Volume 1 (Long and Short Papers)}, pages 4171--4186, Minneapolis, Minnesota. Association for Computational Linguistics.

\bibitem[{Djanibekov et~al.(2025)Djanibekov, Toyin, Alshalan, Alatir, and Aldarmaki}]{djanibekov-etal-2025-dialectal}
Amirbek Djanibekov, Hawau~Olamide Toyin, Raghad Alshalan, Abdullah Alatir, and Hanan Aldarmaki. 2025.
\newblock \href {https://doi.org/10.18653/v1/2025.acl-long.1427} {Dialectal coverage and generalization in {A}rabic speech recognition}.
\newblock In \emph{Proceedings of the 63rd Annual Meeting of the Association for Computational Linguistics (Volume 1: Long Papers)}, pages 29490--29502, Vienna, Austria. Association for Computational Linguistics.

\bibitem[{Dogan-Schönberger et~al.(2021)Dogan-Schönberger, Mäder, and Hofmann}]{swissdial}
Pelin Dogan-Schönberger, Julian Mäder, and Thomas Hofmann. 2021.
\newblock \href {https://arxiv.org/abs/2103.11401} {{SwissDial}: Parallel multidialectal corpus of spoken {S}wiss {G}erman}.
\newblock \emph{Computing Research Repository}, arXiv:2103.11401.

\bibitem[{Dolev et~al.(2024)Dolev, Lutz, and Aepli}]{dolev-etal-2024-whisper}
Eyal Dolev, Clemens Lutz, and No{\"e}mi Aepli. 2024.
\newblock \href {https://doi.org/10.18653/v1/2024.vardial-1.3} {Does {W}hisper understand {S}wiss {G}erman? {An} automatic, qualitative, and human evaluation}.
\newblock In \emph{Proceedings of the Eleventh Workshop on NLP for Similar Languages, Varieties, and Dialects (VarDial 2024)}, pages 28--40, Mexico City, Mexico. Association for Computational Linguistics.

\bibitem[{Duquenne et~al.(2023)Duquenne, Schwenk, and Sagot}]{duquenne-etal-2023-sonar}
Paul-Ambroise Duquenne, Holger Schwenk, and Benoît Sagot. 2023.
\newblock \href {https://arxiv.org/abs/2308.11466} {{SONAR}: Sentence-level multimodal and language-agnostic representations}.
\newblock \emph{Preprint}, arXiv:2308.11466.

\bibitem[{Faisal et~al.(2021)Faisal, Keshava, Alam, and Anastasopoulos}]{faisal-etal-2021-sd-qa}
Fahim Faisal, Sharlina Keshava, Md~Mahfuz~Ibn Alam, and Antonios Anastasopoulos. 2021.
\newblock \href {https://doi.org/10.18653/v1/2021.findings-emnlp.281} {{SD}-{QA}: Spoken dialectal question answering for the real world}.
\newblock In \emph{Findings of the Association for Computational Linguistics: EMNLP 2021}, pages 3296--3315, Punta Cana, Dominican Republic. Association for Computational Linguistics.

\bibitem[{Faruqui and Hakkani-T{\"u}r(2022)}]{faruqui-hakkani-tur-2022-revisiting}
Manaal Faruqui and Dilek Hakkani-T{\"u}r. 2022.
\newblock \href {https://doi.org/10.1162/coli_a_00430} {Revisiting the boundary between {ASR} and {NLU} in the age of conversational dialog systems}.
\newblock \emph{Computational Linguistics}, 48(1):221--232.

\bibitem[{FitzGerald et~al.(2023)FitzGerald, Hench, Peris, Mackie, Rottmann, Sanchez, Nash, Urbach, Kakarala, Singh, Ranganath, Crist, Britan, Leeuwis, Tur, and Natarajan}]{fitzgerald-etal-2023-massive}
Jack FitzGerald, Christopher Hench, Charith Peris, Scott Mackie, Kay Rottmann, Ana Sanchez, Aaron Nash, Liam Urbach, Vishesh Kakarala, Richa Singh, Swetha Ranganath, Laurie Crist, Misha Britan, Wouter Leeuwis, Gokhan Tur, and Prem Natarajan. 2023.
\newblock \href {https://doi.org/10.18653/v1/2023.acl-long.235} {{MASSIVE}: A 1{M}-example multilingual natural language understanding dataset with 51 typologically-diverse languages}.
\newblock In \emph{Proceedings of the 61st Annual Meeting of the Association for Computational Linguistics (Volume 1: Long Papers)}, pages 4277--4302, Toronto, Canada. Association for Computational Linguistics.

\bibitem[{Gao et~al.(2025)Gao, Tamura, and Kato}]{gao-etal-2025-wenzhou}
Zhipeng Gao, Akihiro Tamura, and Tsuneo Kato. 2025.
\newblock \href {https://doi.org/10.18653/v1/2025.loresmt-1.5} {Wenzhou dialect speech to {M}andarin text conversion}.
\newblock In \emph{Proceedings of the Eighth Workshop on Technologies for Machine Translation of Low-Resource Languages (LoResMT 2025)}, pages 36--43, Albuquerque, New Mexico, U.S.A. Association for Computational Linguistics.

\bibitem[{Gerz et~al.(2021)Gerz, Su, Kusztos, Mondal, Lis, Singhal, Mrk{\v{s}}i{\'c}, Wen, and Vuli{\'c}}]{gerz-etal-2021-multilingual}
Daniela Gerz, Pei-Hao Su, Razvan Kusztos, Avishek Mondal, Micha{\l} Lis, Eshan Singhal, Nikola Mrk{\v{s}}i{\'c}, Tsung-Hsien Wen, and Ivan Vuli{\'c}. 2021.
\newblock \href {https://doi.org/10.18653/v1/2021.emnlp-main.591} {Multilingual and cross-lingual intent detection from spoken data}.
\newblock In \emph{Proceedings of the 2021 Conference on Empirical Methods in Natural Language Processing}, pages 7468--7475, Online and Punta Cana, Dominican Republic. Association for Computational Linguistics.

\bibitem[{Gorisch and Schmidt(2024)}]{gorisch-schmidt-2024-evaluating}
Jan Gorisch and Thomas Schmidt. 2024.
\newblock \href {https://aclanthology.org/2024.lrec-main.582/} {Evaluating workflows for creating orthographic transcripts for oral corpora by transcribing from scratch or correcting {ASR}-output}.
\newblock In \emph{Proceedings of the 2024 Joint International Conference on Computational Linguistics, Language Resources and Evaluation (LREC-COLING 2024)}, pages 6564--6574, Torino, Italia. ELRA and ICCL.

\bibitem[{He and Garner(2023)}]{he-garner-2023-interpreter}
Mutian He and Philip Garner. 2023.
\newblock \href {https://doi.org/10.18653/v1/2023.findings-emnlp.291} {The interpreter understands your meaning: End-to-end spoken language understanding aided by speech translation}.
\newblock In \emph{Findings of the Association for Computational Linguistics: EMNLP 2023}, pages 4408--4423, Singapore. Association for Computational Linguistics.

\bibitem[{He et~al.(2021{\natexlab{a}})He, Gao, and Chen}]{he2021debertav3}
Pengcheng He, Jianfeng Gao, and Weizhu Chen. 2021{\natexlab{a}}.
\newblock \href {https://arxiv.org/abs/2111.09543} {{DeBERTaV3}: Improving {DeBERTa} using {ELECTRA}-style pre-training with gradient-disentangled embedding sharing}.
\newblock \emph{Preprint}, arXiv:2111.09543.

\bibitem[{He et~al.(2021{\natexlab{b}})He, Liu, Gao, and Chen}]{he-etal-2021-deberta}
Pengcheng He, Xiaodong Liu, Jianfeng Gao, and Weizhu Chen. 2021{\natexlab{b}}.
\newblock \href {https://openreview.net/forum?id=XPZIaotutsD} {{DeBERTa}: Decoding-enhanced {BERT} with disentangled attention}.
\newblock In \emph{International Conference on Learning Representations}.

\bibitem[{Joshi et~al.(2025)Joshi, Dabre, Kanojia, Li, Zhan, Haffari, and Dippold}]{joshi-etal-2025-natural}
Aditya Joshi, Raj Dabre, Diptesh Kanojia, Zhuang Li, Haolan Zhan, Gholamreza Haffari, and Doris Dippold. 2025.
\newblock \href {https://doi.org/10.1145/3712060} {Natural language processing for dialects of a language: A survey}.
\newblock \emph{ACM Computing Surveys}, 57(6).

\bibitem[{Kanjirangat et~al.(2025)Kanjirangat, Samardzic, Dolamic, and Rinaldi}]{kanjirangat-etal-2025-tokenization}
Vani Kanjirangat, Tanja Samardzic, Ljiljana Dolamic, and Fabio Rinaldi. 2025.
\newblock \href {https://doi.org/10.18653/v1/2025.emnlp-main.1224} {Tokenization and representation biases in multilingual models on dialectal {NLP} tasks}.
\newblock In \emph{Proceedings of the 2025 Conference on Empirical Methods in Natural Language Processing}, pages 23992--24010, Suzhou, China. Association for Computational Linguistics.

\bibitem[{Kasa et~al.(2025)Kasa, Gupta, Roychowdhury, Kumar, Biruduraju, Kasa, Priyatam, Bhattacharya, Agarwal, and Huddar}]{kasa-etal-2025-generative}
Siva~Rajesh Kasa, Karan Gupta, Sumegh Roychowdhury, Ashutosh Kumar, Yaswanth Biruduraju, Santhosh~Kumar Kasa, Pattisapu~Nikhil Priyatam, Arindam Bhattacharya, Shailendra Agarwal, and Vijay Huddar. 2025.
\newblock \href {https://doi.org/10.18653/v1/2025.emnlp-main.486} {Generative or discriminative? {Revisiting} text classification in the era of transformers}.
\newblock In \emph{Proceedings of the 2025 Conference on Empirical Methods in Natural Language Processing}, pages 9615--9637, Suzhou, China. Association for Computational Linguistics.

\bibitem[{Keller and Glavaš(2025)}]{keller-glavas-2025-speechtaxi}
Lennart Keller and Goran Glavaš. 2025.
\newblock \href {https://doi.org/10.1109/ICASSP49660.2025.10890017} {{SpeechTaxi}: On multilingual semantic speech classification}.
\newblock In \emph{2025 IEEE International Conference on Acoustics, Speech and Signal Processing (ICASSP)}.

\bibitem[{Koudounas et~al.(2023)Koudounas, {La Quatra}, Vaiani, Colomba, Attanasio, Pastor, Cagliero, and Baralis}]{koudounas-etal-2023-italic}
Alkis Koudounas, Moreno {La Quatra}, Lorenzo Vaiani, Luca Colomba, Giuseppe Attanasio, Eliana Pastor, Luca Cagliero, and Elena Baralis. 2023.
\newblock \href {https://doi.org/10.21437/Interspeech.2023-1980} {{ITALIC}: An {Italian} intent classification dataset}.
\newblock In \emph{Interspeech 2023}, pages 2153--2157.

\bibitem[{Kr{\"u}ckl et~al.(2025)Kr{\"u}ckl, Blaschke, and Plank}]{kruckl-etal-2025-improving}
Xaver~Maria Kr{\"u}ckl, Verena Blaschke, and Barbara Plank. 2025.
\newblock \href {https://aclanthology.org/2025.vardial-1.10/} {Improving dialectal slot and intent detection with auxiliary tasks: A multi-dialectal {B}avarian case study}.
\newblock In \emph{Proceedings of the 12th Workshop on NLP for Similar Languages, Varieties and Dialects}, pages 128--146, Abu Dhabi, UAE. Association for Computational Linguistics.

\bibitem[{Lee et~al.(2024)Lee, Calapodescu, Gaido, Negri, and Besacier}]{lee-etal-2024-speech-massive}
Beomseok Lee, Ioan Calapodescu, Marco Gaido, Matteo Negri, and Laurent Besacier. 2024.
\newblock \href {https://doi.org/10.21437/Interspeech.2024-957} {{Speech-MASSIVE}: A multilingual speech dataset for {SLU} and beyond}.
\newblock In \emph{Interspeech 2024}, pages 817--821.

\bibitem[{Lee et~al.(2018)Lee, Wu, Liu, and Lee}]{lee-etal-2018-spoken}
Chia-Hsuan Lee, Szu-Lin Wu, Chi-Liang Liu, and Hung-yi Lee. 2018.
\newblock \href {https://www.isca-archive.org/interspeech_2018/lee18d_interspeech.html} {Spoken {SQuAD}: A study of mitigating the impact of speech recognition errors on listening comprehension}.
\newblock \emph{Proc. Interspeech 2018}, pages 3459--3463.

\bibitem[{Lonergan et~al.(2023)Lonergan, Qian, {Ní Chiaráin}, Gobl, and {Ní Chasaide}}]{lonergan-etal-2023-towards}
Liam Lonergan, Mengjie Qian, Neasa {Ní Chiaráin}, Christer Gobl, and Ailbhe {Ní Chasaide}. 2023.
\newblock \href {https://doi.org/10.21437/Interspeech.2023-2528} {Towards dialect-inclusive recognition in a low-resource language: {Are} balanced corpora the answer?}
\newblock In \emph{Interspeech 2023}, pages 5082--5086.

\bibitem[{Malaysha et~al.(2024)Malaysha, El-Haj, Ezzini, Khalilia, Jarrar, Almujaiwel, Berrada, and Bouamor}]{malaysha-etal-2024-arafinnlp}
Sanad Malaysha, Mo~El-Haj, Saad Ezzini, Mohammed Khalilia, Mustafa Jarrar, Sultan Almujaiwel, Ismail Berrada, and Houda Bouamor. 2024.
\newblock \href {https://doi.org/10.18653/v1/2024.arabicnlp-1.34} {{A}ra{F}in{NLP} 2024: The first {A}rabic financial {NLP} shared task}.
\newblock In \emph{Proceedings of the Second Arabic Natural Language Processing Conference}, pages 393--402, Bangkok, Thailand. Association for Computational Linguistics.

\bibitem[{McDowell(2022{\natexlab{a}})}]{mcdowell-2022-xlsr1bde}
Andrew McDowell. 2022{\natexlab{a}}.
\newblock Wav2vec2-xls-r-1b fine-tuned on {CommonVoice8} - {DE}.
\newblock Hugging Face, \url{https://huggingface.co/AndrewMcDowell/wav2vec2-xls-r-1B-german}.

\bibitem[{McDowell(2022{\natexlab{b}})}]{mcdowell-2022-xlsr300de}
Andrew McDowell. 2022{\natexlab{b}}.
\newblock Wav2vec2-xls-r-300m fine-tuned on {CommonVoice7} - {DE}.
\newblock Hugging Face, \url{https://huggingface.co/AndrewMcDowell/wav2vec2-xls-r-300m-german-de}.

\bibitem[{Mu{\~n}oz-Ortiz et~al.(2025)Mu{\~n}oz-Ortiz, Blaschke, and Plank}]{munoz-ortiz-etal-2025-evaluating}
Alberto Mu{\~n}oz-Ortiz, Verena Blaschke, and Barbara Plank. 2025.
\newblock \href {https://aclanthology.org/2025.coling-main.427/} {Evaluating pixel language models on non-standardized languages}.
\newblock In \emph{Proceedings of the 31st International Conference on Computational Linguistics}, pages 6412--6419, Abu Dhabi, UAE. Association for Computational Linguistics.

\bibitem[{Ojo et~al.(2025)Ojo, Ogundepo, Oladipo, Ogueji, Lin, Stenetorp, and Adelani}]{ojo-etal-2025-afrobench}
Jessica Ojo, Odunayo Ogundepo, Akintunde Oladipo, Kelechi Ogueji, Jimmy Lin, Pontus Stenetorp, and David~Ifeoluwa Adelani. 2025.
\newblock \href {https://doi.org/10.18653/v1/2025.findings-acl.976} {{A}fro{B}ench: How good are large language models on {A}frican languages?}
\newblock In \emph{Findings of the Association for Computational Linguistics: ACL 2025}, pages 19048--19095, Vienna, Austria. Association for Computational Linguistics.

\bibitem[{Papakyriakopoulos et~al.(2023)Papakyriakopoulos, Choi, Thong, Zhao, Andrews, Bourke, Xiang, and Koenecke}]{papakyriakopoulos-etal-2023-augmented}
Orestis Papakyriakopoulos, Anna Seo~Gyeong Choi, William Thong, Dora Zhao, Jerone Andrews, Rebecca Bourke, Alice Xiang, and Allison Koenecke. 2023.
\newblock \href {https://doi.org/10.1145/3593013.3594049} {Augmented datasheets for speech datasets and ethical decision-making}.
\newblock In \emph{Proceedings of the 2023 ACM Conference on Fairness, Accountability, and Transparency}, pages 881--904, New York, NY, USA. Association for Computing Machinery.

\bibitem[{Phung et~al.(2024)Phung, Deshpande, Emami, and Singh}]{phung-etal-2024-ar-nlu}
Emmy Phung, Harsh Deshpande, Ahmad Emami, and Kanishk Singh. 2024.
\newblock \href {https://doi.org/10.21437/Interspeech.2024-1292} {{AR-NLU}: A framework for enhancing natural language understanding model robustness against {ASR} errors}.
\newblock In \emph{Interspeech 2024}, pages 1325--1329.

\bibitem[{Pratap et~al.(2024)Pratap, Tjandra, Shi, Tomasello, Babu, Kundu, Elkahky, Ni, Vyas, Fazel-Zarandi, Baevski, Adi, Zhang, Hsu, Conneau, and Auli}]{pratap-etal-2024-scaling}
Vineel Pratap, Andros Tjandra, Bowen Shi, Paden Tomasello, Arun Babu, Sayani Kundu, Ali Elkahky, Zhaoheng Ni, Apoorv Vyas, Maryam Fazel-Zarandi, Alexei Baevski, Yossi Adi, Xiaohui Zhang, Wei-Ning Hsu, Alexis Conneau, and Michael Auli. 2024.
\newblock \href {https://jmlr.org/papers/volume25/23-1318/23-1318.pdf} {Scaling speech technology to 1,000+ languages}.
\newblock \emph{JMLR}, 25(97):1--52.

\bibitem[{Radford et~al.(2023)Radford, Kim, Xu, Brockman, McLeavey, and Sutskever}]{radford-etal-2023-robust}
Alec Radford, Jong~Wook Kim, Tao Xu, Greg Brockman, Christine McLeavey, and Ilya Sutskever. 2023.
\newblock Robust speech recognition via large-scale weak supervision.
\newblock In \emph{ICML'23: Proceedings of the 40th International Conference on Machine Learning}. JMLR.org.

\bibitem[{Rajaa et~al.(2022)Rajaa, Dalmia, and Nethil}]{rajaa2022skits2i}
Shangeth Rajaa, Swaraj Dalmia, and Kumarmanas Nethil. 2022.
\newblock \href {https://arxiv.org/abs/2212.13015} {{Skit-S2I}: An {Indian} accented speech to intent dataset}.
\newblock \emph{Preprint}, arXiv:2212.13015.

\bibitem[{Rust et~al.(2023)Rust, Lotz, Bugliarello, Salesky, de~Lhoneux, and Elliott}]{rust-etal-2023-language}
Phillip Rust, Jonas~F Lotz, Emanuele Bugliarello, Elizabeth Salesky, Miryam de~Lhoneux, and Desmond Elliott. 2023.
\newblock \href {https://openreview.net/forum?id=FkSp8VW8RjH} {Language modelling with pixels}.
\newblock In \emph{The Eleventh International Conference on Learning Representations}.

\bibitem[{Saattrup~Nielsen et~al.(2025)Saattrup~Nielsen, Enevoldsen, and Schneider-Kamp}]{saattrup-nielsen-etal-2025-encoder}
Dan Saattrup~Nielsen, Kenneth Enevoldsen, and Peter Schneider-Kamp. 2025.
\newblock \href {https://aclanthology.org/2025.nodalida-1.60/} {Encoder vs decoder: {Comparative} analysis of encoder and decoder language models on multilingual {NLU} tasks}.
\newblock In \emph{Proceedings of the Joint 25th Nordic Conference on Computational Linguistics and 11th Baltic Conference on Human Language Technologies (NoDaLiDa/Baltic-HLT 2025)}, pages 561--572, Tallinn, Estonia. University of Tartu Library.

\bibitem[{Sanabria et~al.(2023)Sanabria, Bogoychev, Markl, Carmantini, Klejch, and Bell}]{sanabria-etal-2023-edinburgh}
Ramon Sanabria, Nikolay Bogoychev, Nina Markl, Andrea Carmantini, Ondrej Klejch, and Peter Bell. 2023.
\newblock \href {https://doi.org/10.1109/ICASSP49357.2023.10095057} {The {Edinburgh International Accents of English Corpus}: Towards the democratization of {English ASR}}.
\newblock In \emph{2023 IEEE International Conference on Acoustics, Speech and Signal Processing (ICASSP)}.

\bibitem[{Scharenborg et~al.(2020)Scharenborg, Besacier, Black, Hasegawa-Johnson, Metze, Neubig, Stüker, Godard, Müller, Ondel, Palaskar, Arthur, Ciannella, Du, Larsen, Merkx, Riad, Wang, and Dupoux}]{scharenborg-etal-2020-speech}
Odette Scharenborg, Laurent Besacier, Alan Black, Mark Hasegawa-Johnson, Florian Metze, Graham Neubig, Sebastian Stüker, Pierre Godard, Markus Müller, Lucas Ondel, Shruti Palaskar, Philip Arthur, Francesco Ciannella, Mingxing Du, Elin Larsen, Danny Merkx, Rachid Riad, Liming Wang, and Emmanuel Dupoux. 2020.
\newblock \href {https://doi.org/10.1109/TASLP.2020.2973896} {Speech technology for unwritten languages}.
\newblock \emph{IEEE/ACM Transactions on Audio, Speech, and Language Processing}, 28:964--975.

\bibitem[{Scherrer et~al.(2025)Scherrer, van~der Goot, and M{\ae}hlum}]{scherrer-etal-2025-findings}
Yves Scherrer, Rob van~der Goot, and Petter M{\ae}hlum. 2025.
\newblock \href {https://aclanthology.org/2025.vardial-1.1/} {Findings of the {V}ar{D}ial evaluation campaign 2025: The {N}or{SID} shared task on {N}orwegian slot, intent and dialect identification}.
\newblock In \emph{Proceedings of the 12th Workshop on NLP for Similar Languages, Varieties and Dialects}, pages 1--8, Abu Dhabi, UAE. Association for Computational Linguistics.

\bibitem[{Schmidt et~al.(2025)Schmidt, Vulić, Glavaš, and Adelani}]{schmidt-etal-2025-fleurs-slu}
Fabian~David Schmidt, Ivan Vulić, Goran Glavaš, and David~Ifeoluwa Adelani. 2025.
\newblock \href {https://arxiv.org/abs/2501.06117} {{Fleurs-SLU}: A massively multilingual benchmark for spoken language understanding}.
\newblock \emph{Preprint}, arXiv:2501.06117.

\bibitem[{Schuster et~al.(2019)Schuster, Gupta, Shah, and Lewis}]{schuster-etal-2019-cross-lingual}
Sebastian Schuster, Sonal Gupta, Rushin Shah, and Mike Lewis. 2019.
\newblock \href {https://doi.org/10.18653/v1/N19-1380} {Cross-lingual transfer learning for multilingual task oriented dialog}.
\newblock In \emph{Proceedings of the 2019 Conference of the North {A}merican Chapter of the Association for Computational Linguistics: Human Language Technologies, Volume 1 (Long and Short Papers)}, pages 3795--3805, Minneapolis, Minnesota. Association for Computational Linguistics.

\bibitem[{Shim et~al.(2025)Shim, Cristofaro, Hu, Vietti, and Plank}]{shim-etal-2025-languages}
Ryan Soh-Eun Shim, Domenico~De Cristofaro, Chengzhi~Martin Hu, Alessandro Vietti, and Barbara Plank. 2025.
\newblock \href {https://arxiv.org/abs/2505.19606} {Languages in multilingual speech foundation models align both phonetically and semantically}.
\newblock \emph{Preprint}, arXiv:2505.19606.

\bibitem[{Sicard et~al.(2023)Sicard, Gillioz, and Pyszkowski}]{sicard-etal-2023-spaiche}
Cl{\'e}ment Sicard, Victor Gillioz, and Kajetan Pyszkowski. 2023.
\newblock \href {https://aclanthology.org/2023.swisstext-1.8/} {Spaiche: Extending state-of-the-art {ASR} models to {S}wiss {G}erman dialects}.
\newblock In \emph{Proceedings of the 8th edition of the Swiss Text Analytics Conference}, pages 76--83, Neuchatel, Switzerland. Association for Computational Linguistics.

\bibitem[{Srivastava and Chiang(2023)}]{srivastava-chiang-2023-fine}
Aarohi Srivastava and David Chiang. 2023.
\newblock \href {https://doi.org/10.18653/v1/2023.vardial-1.16} {Fine-tuning {BERT} with character-level noise for zero-shot transfer to dialects and closely-related languages}.
\newblock In \emph{Tenth Workshop on NLP for Similar Languages, Varieties and Dialects (VarDial 2023)}, pages 152--162, Dubrovnik, Croatia. Association for Computational Linguistics.

\bibitem[{Srivastava and Chiang(2025)}]{srivastava-chiang-2025-calling}
Aarohi Srivastava and David Chiang. 2025.
\newblock \href {https://doi.org/10.18653/v1/2025.wnut-1.6} {We{'}re calling an intervention: Exploring fundamental hurdles in adapting language models to nonstandard text}.
\newblock In \emph{Proceedings of the Tenth Workshop on Noisy and User-generated Text}, pages 45--56, Albuquerque, New Mexico, USA. Association for Computational Linguistics.

\bibitem[{Tang et~al.(2021)Tang, Wang, Xu, Sun, Lei, Zhao, Wen, Tan, Xie, Zhou, Yan, Lv, Han, Zou, and Li}]{tang-etal-2021-kespeech}
Zhiyuan Tang, Dong Wang, Yanguang Xu, Jianwei Sun, Xiaoning Lei, Shuaijiang Zhao, Cheng Wen, Xingjun Tan, Chuandong Xie, Shuran Zhou, Rui Yan, Chenjia Lv, Yang~Han Han, Wei Zou, and Xiangang Li. 2021.
\newblock \href {https://openreview.net/forum?id=b3Zoeq2sCLq} {{KeSpeech}: {An} open source speech dataset of {Mandarin} and its eight subdialects}.
\newblock In \emph{NeurIPS 2021 Datasets and Benchmarks Track}.

\bibitem[{ten Bosch(2000)}]{bosch-2000-asr}
Louis ten Bosch. 2000.
\newblock \href {https://doi.org/10.21437/ICSLP.2000-705} {{ASR}, dialects, and acoustic/phonological distances}.
\newblock In \emph{6th International Conference on Spoken Language Processing (ICSLP 2000)}, pages vol. 3, 1009--1012.

\bibitem[{Tian et~al.(2025)Tian, Chen, Peng, Shi, Arora, Bharadwaj, Maekaku, Shinohara, Goto, Yue, Yang, and Watanabe}]{tian-etal-2025-opuslm}
Jinchuan Tian, William Chen, Yifan Peng, Jiatong Shi, Siddhant Arora, Shikhar Bharadwaj, Takashi Maekaku, Yusuke Shinohara, Keita Goto, Xiang Yue, Huck Yang, and Shinji Watanabe. 2025.
\newblock \href {https://doi.org/10.21437/Interspeech.2025-1184} {{OpusLM}: A family of open unified speech language models}.
\newblock In \emph{Interspeech 2025}, pages 3259--3263. ISCA.

\bibitem[{Torgbi et~al.(2025)Torgbi, Clayman, Speight, and Tayyar~Madabushi}]{torgbi-etal-2025-adapting}
Melissa Torgbi, Andrew Clayman, Jordan~J. Speight, and Harish Tayyar~Madabushi. 2025.
\newblock \href {https://aclanthology.org/2025.vardial-1.4/} {Adapting {Whisper} for regional dialects: Enhancing public services for vulnerable populations in the {U}nited {K}ingdom}.
\newblock In \emph{Proceedings of the 12th Workshop on NLP for Similar Languages, Varieties and Dialects}, pages 29--38, Abu Dhabi, UAE. Association for Computational Linguistics.

\bibitem[{Vakirtzian et~al.(2024)Vakirtzian, Tsoukala, Bompolas, Mouzou, Stamou, Paraskevopoulos, Dimakis, Markantonatou, Ralli, and Anastasopoulos}]{vakirtzian-etal-2024-speech}
Socrates Vakirtzian, Chara Tsoukala, Stavros Bompolas, Katerina Mouzou, Vivian Stamou, Georgios Paraskevopoulos, Antonios Dimakis, Stella Markantonatou, Angela Ralli, and Antonios Anastasopoulos. 2024.
\newblock Speech recognition for {Greek Dialects}: {A} challenging benchmark.
\newblock In \emph{Interspeech 2024}, pages 3974--3978.

\bibitem[{van~der Goot et~al.(2021)van~der Goot, Sharaf, Imankulova, {\"U}st{\"u}n, Stepanovi{\'c}, Ramponi, Khairunnisa, Komachi, and Plank}]{van-der-goot-etal-2021-masked}
Rob van~der Goot, Ibrahim Sharaf, Aizhan Imankulova, Ahmet {\"U}st{\"u}n, Marija Stepanovi{\'c}, Alan Ramponi, Siti~Oryza Khairunnisa, Mamoru Komachi, and Barbara Plank. 2021.
\newblock \href {https://doi.org/10.18653/v1/2021.naacl-main.197} {From masked language modeling to translation: Non-{E}nglish auxiliary tasks improve zero-shot spoken language understanding}.
\newblock In \emph{Proceedings of the 2021 Conference of the North American Chapter of the Association for Computational Linguistics: Human Language Technologies}, pages 2479--2497, Online. Association for Computational Linguistics.

\bibitem[{Wang et~al.(2023)Wang, Li, Guo, Qiao, Li, Shang, Wei, Tao, Zhang, and Yang}]{wang-etal-2023-whislu}
Minghan Wang, Yinglu Li, Jiaxin Guo, Xiaosong Qiao, Zongyao Li, Hengchao Shang, Daimeng Wei, Shimin Tao, Min Zhang, and Hao Yang. 2023.
\newblock \href {https://doi.org/10.21437/Interspeech.2023-1505} {{WhiSLU}: End-to-end spoken language understanding with {Whisper}}.
\newblock In \emph{Interspeech 2023}, pages 770--774.

\bibitem[{Winkler et~al.(2024)Winkler, Juozapaityte, van~der Goot, and Plank}]{winkler-etal-2024-slot}
Miriam Winkler, Virginija Juozapaityte, Rob van~der Goot, and Barbara Plank. 2024.
\newblock \href {https://aclanthology.org/2024.lrec-main.1297/} {Slot and intent detection resources for {B}avarian and {L}ithuanian: Assessing translations vs natural queries to digital assistants}.
\newblock In \emph{Proceedings of the 2024 Joint International Conference on Computational Linguistics, Language Resources and Evaluation (LREC-COLING 2024)}, pages 14898--14915, Torino, Italia. ELRA and ICCL.

\bibitem[{Wolf et~al.(2020)Wolf, Debut, Sanh, Chaumond, Delangue, Moi, Cistac, Rault, Louf, Funtowicz, Davison, Shleifer, von Platen, Ma, Jernite, Plu, Xu, Scao, Gugger, Drame, Lhoest, and Rush}]{wolf2020huggingface}
Thomas Wolf, Lysandre Debut, Victor Sanh, Julien Chaumond, Clement Delangue, Anthony Moi, Pierric Cistac, Tim Rault, Rémi Louf, Morgan Funtowicz, Joe Davison, Sam Shleifer, Patrick von Platen, Clara Ma, Yacine Jernite, Julien Plu, Canwen Xu, Teven~Le Scao, Sylvain Gugger, and 3 others. 2020.
\newblock \href {https://arxiv.org/abs/1910.03771} {Huggingface's transformers: State-of-the-art natural language processing}.
\newblock \emph{Preprint}, arXiv:1910.03771.

\bibitem[{Yadavalli et~al.(2022)Yadavalli, Mirishkar, and Vuppala}]{yadavalli-etal-2022-multi-task}
Aditya Yadavalli, Ganesh Mirishkar, and Anil~Kumar Vuppala. 2022.
\newblock \href {https://doi.org/10.21437/Interspeech.2022-10739} {Multi-task end-to-end model for {Telugu} dialect and speech recognition}.
\newblock In \emph{Interspeech 2022}, pages 1387--1391.

\bibitem[{Zampieri et~al.(2020)Zampieri, Nakov, and Scherrer}]{zampieri-etal-2020-natural}
Marcos Zampieri, Preslav Nakov, and Yves Scherrer. 2020.
\newblock \href {https://doi.org/10.1017/S1351324920000492} {Natural language processing for similar languages, varieties, and dialects: A survey}.
\newblock \emph{Natural Language Engineering}, 26(6):595–612.

\bibitem[{{Zanon Boito} et~al.(2024){Zanon Boito}, Iyer, Lagos, Besacier, and Calapodescu}]{zanon-boito-etal-2024-mhubert}
Marcely {Zanon Boito}, Vivek Iyer, Nikolaos Lagos, Laurent Besacier, and Ioan Calapodescu. 2024.
\newblock \href {https://doi.org/10.21437/Interspeech.2024-938} {{mHuBERT-147}: A compact multilingual {HuBERT} model}.
\newblock In \emph{Interspeech 2024}, pages 3939--3943.

\end{thebibliography}

\appendix

\section{Data Statement}
\label{sec:appendix-datasheet}

\subsection{Summary and motivation}

\begin{itemize}
\item Dataset name and version: xSID-audio, version~0.1
\item Dataset curators: Verena Blaschke, Miriam Winkler, Barbara Plank
\item Dataset and data statement citation: Please cite this paper.
\item Data statement version: 1.0
\end{itemize}

\paragraph{Executive summary and curation rationale}
This dataset extends the slot and intent classification dataset xSID by providing audio recordings for 300 development and 500 test sentences in German and Bavarian.
Its purpose is to allow research on intent classification for spoken dialectal data.

\subsection{Documentation for source datasets}
\label{sec:datasheet-source-datasets}

We use the written German and Bavarian test datasets from xSID~0.5 (\citealp{van-der-goot-etal-2021-masked, aepli-etal-2023-findings, winkler-etal-2024-slot}; \href{https://github.com/mainlp/xsid/blob/main/LICENSE}{CC BY-SA 4.0}).
xSID itself is derived from Snips (\citealp{coucke2018snips}; \href{https://github.com/sonos/nlu-benchmark/blob/master/LICENSE}{CC0 1.0 Universal}) and and a dataset by \citeauthor{schuster-etal-2019-cross-lingual} (\citeyear{schuster-etal-2019-cross-lingual}; \href{https://creativecommons.org/licenses/by-sa/4.0/}{CC-BY-SA}).
For details on the German translation, see \citet{van-der-goot-etal-2021-masked}; for details on the Bavarian translation, \citet{winkler-etal-2024-slot}.

\subsection{Composition}
\label{sec:datasheet-composition}
Our dataset consists of spoken sentences with intent labels.
All of the sentences are from the development and test splits of xSID.

\paragraph{Date} The original sentences are likely from ca.\ 2018 \cite{coucke2018snips, schuster-etal-2019-cross-lingual}.
The German translation \cite{van-der-goot-etal-2021-masked} is from ca.\ 2020 and the Bavarian translation \cite{winkler-etal-2024-slot} is from 2023.
Our recordings are from 2024.

\paragraph{Modality}Spoken (read speech).%

\paragraph{Genre} Questions and commands for a virtual personal assistant.

\paragraph{Language varieties and speaker}
We record audios in Standard German spoken in Germany (``de'') and in the dialectal variant of Bavarian as spoken in rural Upper Bavaria (``de-ba'').
The audios were recorded by a native speaker of both German and Bavarian in her 20s.
The speaker is the same person who translated the text version of xSID de-ba.

\begin{table}[t]
\adjustbox{max width=\linewidth, center}{%
\begin{tabular}{@{}llrrr@{}}
\toprule
\textbf{Split} &  & \textbf{N} & \textbf{\# Int\rlap{.}} & \textbf{Duration (s)} \\
 \midrule
\multicolumn{5}{@{}l}{\textit{xSID-audio 0.1}} \\
de & dev & 300 & 15\rlap{*} & 3.33$\pm$1.05 \\
 & test & 500 & 15\rlap{*} & 3.47$\pm$1.00 \\
de-ba & dev & 300 & 15\rlap{*} & 3.04$\pm$0.98 \\
 & test & 500 & 15\rlap{*} & 3.56$\pm$0.99 \\
 \midrule
\multicolumn{5}{@{}l}{\textit{Subset used for experiments in this paper}} \\
de (``deu'') & test & 412 & {10} & 3.48$\pm$1.05 \\
de-ba (``bar'') & test & 412 & {10} & 3.52$\pm$1.02 \\
\bottomrule
\end{tabular}}
\caption{\textbf{Dataset statistics for xSID-audio.} 
Intents: *The development and test splits of xSID(-audio) contain 16 intent labels in total, but only 15 per split (14 are shared across splits).
Duration: mean duration $\pm$ standard deviation.}
\label{tab:xsid-audio-stats}
\end{table}

\paragraph{Dataset statistics}
The dataset consists of one German and one Bavarian audio recording for each of the 300 development and 500 test sentences.
On average, each audio clip in xSID-audio~0.1 has a duration of 3.4\,s for German and 3.3\,s for Bavarian.
Detailed statistics are in Table~\ref{tab:xsid-audio-stats}.

\paragraph{Labels}
Each sentence is annotated with one of 16 intent labels, although the development set and test set each only use a subset of 15 labels.
The labels are identical to the ones in xSID.
The label distribution for the subset of instances used in the experiments in this paper is in Table~\ref{tab:data-stats} in~\S\ref{sec:appendix-intent-labels}.
For annotation details, see \citet{van-der-goot-etal-2021-masked}.

\paragraph{Data quality}
We directly worked with xSID's German and Bavarian sentences.
The German sentences contain a few intentional grammatical errors to reflect errors in the original English data \cite{van-der-goot-etal-2021-masked}.
The Bavarian dataset does not contain such errors \cite{winkler-etal-2024-slot}.
The recordings directly reflect the written versions.

\subsection{Audio recording and formatting}

We record the samples with a R{\o}de NT-USB microphone with pop filter in a quiet room. 
The microphone was set up at the desk and the speaker spoke from a distance of about 20\,cm. 
The instances were read one sentence at a time directly from the dataset text with breaks in between. 
After recording, we convert the audio files to the {wav mulaw} format, resulting in stereo audios with a recording rate of 48~kHz and a bit rate of 768~kBit/s. 

The audios are in a folder each for each language and dataset split with corresponding TSV files containing the sentence ID, intent, written sentence and audio filename.

\subsection{Distribution, use, and maintenance}

\paragraph{Distribution and usage terms}
We release the dataset on Zenodo (\href{https://doi.org/10.5281/zenodo.19554427}{\texttt{10.5281/zenodo.19554427}}) under the \ccbysafour{} license (in accordance with xSID's license).
We prohibit the use of xSID-audio for voice cloning or speech synthesis.

\paragraph{Other uses} The dataset could also be used for slot filling (in combination with the slot annotations in xSID). 
It could further be used for ASR and for phonetic studies.

\paragraph{Maintenance and updates}
We have not planned any specific updates.
If you find errors in the dataset, please email 
Verena Blaschke or Barbara Plank.

\subsection{Limitations, acknowledgements, and further information}

\paragraph{Limitations}
In order to be parallel to the already existing test version of xSID, this dataset consists of read speech.
It is therefore not representative of spontaneous speech.

This dataset reflects the dialect of a single speaker.

\begin{table*}[t]
\centering
\adjustbox{max width=\linewidth}{%
\begin{tabular}{@{}lll@{}}
\toprule
\textbf{Resource} & \textbf{License} & \textbf{Hugging Face ID / Dataset link} \\
\midrule
{\textit{Models}} \\
mBERT &  \apache &\huggingface{google-bert/bert-base-multilingual-cased} \\
mDeBERTa & \mitlicense &\huggingface{microsoft/mdeberta-v3-base} \\
XLM-R base & \mitlicense &\huggingface{FacebookAI/xlm-roberta-base} \\
XLM-R large & \mitlicense &\huggingface{FacebookAI/xlm-roberta-large} \\
mHuBERT & \ccbyncsathree & \huggingface{utter-project/mHuBERT-147} \\
XLS-R 300M & \apache &\huggingface{facebook/wav2vec2-xls-r-300m} \\
XLS-R 300M DE & \apache &\huggingface{AndrewMcDowell/wav2vec2-xls-r-300m-german-de} \\
XLS-R 1B DE & \apache &\huggingface{AndrewMcDowell/wav2vec2-xls-r-1B-german} \\
MMS 300M & \ccbyncfour & \huggingface{facebook/mms-300m} \\
MMS 1B-all & \ccbyncfour & \huggingface{facebook/mms-1b-all} \\
Whisper tiny & \apache &\huggingface{openai/whisper-tiny} \\
Whisper base & \apache &\huggingface{openai/whisper-base} \\
Whisper small & \apache &\huggingface{openai/whisper-small} \\
Whisper medium & \apache &\huggingface{openai/whisper-medium} \\
Whisper large-v3 & \apache &\huggingface{openai/whisper-large-v3} \\
Whisper large-v3-turbo & \mitlicense &\huggingface{openai/whisper-large-v3-turbo} \\
\midrule
{\textit{Datasets}} \\
MASSIVE &  \apache & \github{alexa/massive} \\
{Speech-MASSIVE} &  \ccbyncsafour & \github{hlt-mt/Speech-MASSIVE} \\
MASSIVE de:ba & not specified & \github{mainlp/NaLiBaSID} \\
{xSID} &  \ccbysafour & \github{mainlp/xsid} \\
xSID-audio & \ccbysafour & \href{https://doi.org/10.5281/zenodo.19554428}{doi.org/10.5281/zenodo.19554428} \\
SwissDial & \ccbyncfour & \href{https://mtc.ethz.ch/publications/open-source/swiss-dial.html}{mtc.ethz.ch/publications/open-source/swiss-dial.html} \\
\bottomrule
\end{tabular}}
\caption{\textbf{Links and licenses for the resources we use/release.}}
\label{tab:licenses}
\end{table*}

\paragraph{Disclosures and acknowledgements}
There are no conflicts of interest.
This research is supported by European Research Council (ERC) Consolidator Grant DIALECT 101043235.

\paragraph{About this data statement}
A data statement is a characterization of a dataset that provides context to allow developers and users to better understand how experimental results might generalize, how software might be appropriately deployed, and what biases might be reflected in systems built on the software.

This data statement was written based on the template for the Data Statements Version~3 Schema. The template was prepared by Angelina McMillan-Major and Emily M.\ Bender and can be found at \url{http://techpolicylab.uw.edu/data-statements}.

We adapted the structure and questions of the template and integrated parts of the audio-specific data statements by \citet{papakyriakopoulos-etal-2023-augmented} and \citet{agnew2024sound}.

\section{Resource Links and Licenses}
\label{sec:appendix-resource-licenses}

Table~\ref{tab:licenses} provides license details and resource links for the models and datasets we use.
We use all resources in accordance with their intended use.

The MASSIVE dataset \cite{fitzgerald-etal-2023-massive} is a multilingual, text-only extension of SLURP \cite{bastianelli-etal-2020-slurp}.
Speech-MASSIVE \cite{lee-etal-2024-speech-massive} adds audio recordings for several of the languages in MASSIVE.
The Bavarian translation of a subset of MASSIVE that we use was introduced in \citet{winkler-etal-2024-slot}, where it is called ``MASSIVE de:ba''. 
For details on the source datasets that xSID is built on, see~\ref{sec:datasheet-source-datasets}.

\section{Data Preprocessing}
\label{sec:appendix-data}

\subsection{Mapping the intent labels}
\label{sec:appendix-intent-labels}

\begin{table}
\centering
\adjustbox{max width=\linewidth}{
\renewcommand{\arraystretch}{1.2}
\begin{tabular}{@{}ll@{}}
\toprule
\textbf{MASSIVE label} \hfill $\rightarrow$ & \textbf{xSID label} \\
\midrule
\texttt{play\_music} & \texttt{PlayMusic} \\
\texttt{recommendation\_movies} & \texttt{SearchScreeningEvent} \\[-5pt]
\multicolumn{2}{@{}l}{(if slots contain \texttt{place\_name} and/or \texttt{business\_type})}\\
\texttt{alarm\_remove} & \texttt{alarm/cancel\_alarm} \\
\texttt{alarm\_set} & \texttt{alarm/set\_alarm} \\
\texttt{alarm\_query} & \texttt{alarm/show\_alarms} \\
\texttt{calendar\_remove} & \texttt{reminder/cancel\_reminder} \\[-5pt]
\multicolumn{2}{@{}l}{(if English text includes {``remind''} and/or {``notif''})} \\
\texttt{calendar\_set} & \texttt{reminder/set\_reminder} \\[-5pt]
\multicolumn{2}{@{}l}{(if English text includes {``remind''} and/or ``{notif}'')} \\
\texttt{calendar\_query} & \texttt{reminder/show\_reminders} \\[-5pt]
\multicolumn{2}{@{}l}{(if English text includes {``remind''})}\\
\texttt{weather\_query} & \texttt{weather/find}\\
\bottomrule
\end{tabular}}
\caption{\textbf{Intent label mappings.} See \S\ref{sec:appendix-intent-labels} for additional mapping details.}
\label{tab:mapping}
\end{table}

\begin{table}[]
\centering
\adjustbox{max width=\linewidth}{
\begin{tabular}{@{}l@{}rrrr@{}}
\toprule
 & \multicolumn{3}{l}{\textbf{\llap{(Sp}eech-)MASSIVE}} & {\textbf{xSID}} \\ 
 \cmidrule(rl){2-4} \cmidrule{5-5}
\textbf{Intent} & {\textbf{Train}} & {\textbf{Dev}} & {\textbf{Test}} & {\textbf{Test}} \\
 \midrule
AddToPlaylist & 24 & 4 & 8 & 34 \\
PlayMusic & 923 & 168 & 105 & 39 \\
BookRestaurant & 23 & 3 & 1 & 43 \\
SearchScreeningEvent & 43 & 9 & 9 & 37 \\
alarm/cancel\_alarm & 71 & 13 & 21 & 33 \\
alarm/set\_alarm & 173 & 30 & 40 & 29 \\
alarm/show\_alarms & 132 & 21 & 31 & 19 \\
reminder/set\_reminder & 350 & 58 & 41 & 37 \\
reminder/show\_reminders & 156 & 27 & 32 & 19 \\
\midrule
Total & 2468 & 459 & 361 & 412 \\ \bottomrule
\end{tabular}}
    \caption{Label distribution in the intent classification datasets after mapping the labels.}
    \label{tab:data-stats}
\end{table}

We map the (Speech-)MASSIVE labels to the ones used by xSID:
If a sentence in MASSIVE was manually re-annotated with an xSID label by \citet{winkler-etal-2024-slot}, we use that label.
Otherwise, if it can be mapped according to Table~\ref{tab:mapping}, we update the label accordingly.
Otherwise, we exclude the sentence.
Sentences with labels that occur less than ten times in MASSIVE's training data are excluded as well.
The resulting label distribution is shown in Table~\ref{tab:data-stats}.

\subsection{Excluded topic labels}
\label{sec:appendix-topic-labels}

When using SwissDial, we remove sentences with labels that are not content-based topics: \texttt{Random} (these appear to be randomly chosen sentences from Wikipedia) and \texttt{Special} (sentences showcasing lexical variation across dialects). 
We also exclude sentences with labels that occur very rarely: \texttt{Earth/Space} and \texttt{Story}.

\section{Hyperparameters and Training}
\label{sec:appendix-hyperparams-training}

We use HuggingFace's \textit{transformers} library (\citealp{wolf2020huggingface}, v4.51.3, Apache~2.0 license) to fine-tune our models.
We use GPUs of type NVIDIA H200 for fine-tuning the Whisper large-v3 models and NVIDIA A100-SXM4-80GB for the remaining models.

We use the training and development datasets of the German splits of \textspeechmassive{} and Swissdial for selecting the learning rate and maximum number of training epochs.
We train each model configuration on three random seeds, both for hyperparameter tuning and for the main experiments.
We always choose the model epoch with the highest development accuracy.
The hyperparameter values are based on prior work on written and spoken intent classification
\cite{rajaa2022skits2i, koudounas-etal-2023-italic, arora-etal-2024-universlu, lee-etal-2024-speech-massive, schmidt-etal-2025-fleurs-slu}.
We otherwise use the default values of the \textit{TrainingArguments} class.

\begin{itemize}
    \item Maximum number of training epochs (written input): 10
    \item Maximum number of training epochs (spoken input): 30 (used in hyperparameter search). Maximum number picked based on hyperparameter experiments: 15 (Whisper small, Whisper large-v3), 17 (Whisper medium), 19 (Whisper tiny, Whisper base), 23 (\xlsr{} 300M DE), 29 (MMS 300M, mHuBERT), 30 (\xlsr{} 300M)
    \item Learning rate: \{1e-5, 5e-5, 1e-4\}. Learning rates chosen: 
    \begin{itemize}
        \item Intents: 1e-5 (\xlmr{} base, mDeBERTa, Whisper medium, Whisper large-v3), 5e-5 (mBERT, Whisper small, \xlsr{} 300M, MMS 300M, mHuBERT), \mbox{1e-4} (\xlmr{} large, \xlsr{} 300M DE, Whisper tiny, Whisper base)
        \item Topics: 1e-5 (\xlmr{} large), 5e-5 (mBERT, \xlmr{} base), 1e-4 (mDeBERTa)
    \end{itemize}
    \item Batch size: 32
    \item Warm-up ratio: 0.1
    \item Weight decay: 0.01
\end{itemize}

For the intent classification task, we observe that the text models are much more robust to the learning rate choice than the speech models (Table~\ref{tab:learning_rates}).
\begin{table}[t]
\setlength{\tabcolsep}{1pt} %
\centering
\adjustbox{max width=\linewidth}{%
\begin{tabular}{@{}lrrr@{}}
\multicolumn{4}{c}{Intents}\\
\toprule
\textbf{Model} & \multicolumn{1}{c}{\textbf{1e-5}} & \multicolumn{1}{c}{\textbf{5e-5}} & \multicolumn{1}{c}{\textbf{1e-4}} \\
\midrule
mBERT & 95.6\textsubscript{0.2} & \textbf{96.0}\textsubscript{0.3} & 95.6\textsubscript{0.3} \\
mDeBERTa & \textbf{97.3}\textsubscript{0.1} & 97.2\textsubscript{0.6} & 96.9\textsubscript{0.4}\\
XLM-R base & \textbf{97.5}\textsubscript{0.3} & 97.2\textsubscript{0.3} & 96.8\textsubscript{0.3}\\
XLM-R large & 96.7\textsubscript{0.4} & 96.4\textsubscript{1.8} & \textbf{96.9}\textsubscript{0.2} \\
\midrule
mHuBERT & 67.1\textsubscript{1.6} & \textbf{85.1}\textsubscript{1.4} & 49.0\textsubscript{21.5} \\
XLS-R 300M  & 42.4\textsubscript{0.7} & \textbf{86.3}\textsubscript{0.6} & 81.7\textsubscript{2.6} \\
XLS-R 300M DE\textsuperscript{ASR} & 88.0\textsubscript{0.9} & 95.1\textsubscript{0.7} & \textbf{95.4}\textsubscript{0.4} \\
MMS 300M & 66.7\textsubscript{0.8} & \textbf{83.0}\textsubscript{0.7} & 74.4\textsubscript{9.5}\\
Whisper tiny\textsuperscript{ASR} & 71.2\textsubscript{0.3} & 74.1\textsubscript{0.8} & \textbf{77.1}\textsubscript{0.5} \\
Whisper base\textsuperscript{ASR} & 74.8\textsubscript{2.5} & 85.4\textsubscript{4.3} & \textbf{89.4}\textsubscript{1.9} \\
Whisper small\textsuperscript{ASR} & 91.1\textsubscript{2.1} & \textbf{94.0}\textsubscript{0.3} & 92.3\textsubscript{1.1} \\
Whisper medium\textsuperscript{ASR} & \textbf{95.3}\textsubscript{0.8} & 94.4\textsubscript{0.8} & 93.5\textsubscript{1.1} \\
Whisper large-v3\textsuperscript{ASR} & \textbf{94.1}\textsubscript{0.9} & 93.1\textsubscript{1.0} & 92.2\textsubscript{0.3} \\
\bottomrule
\\
\multicolumn{4}{c}{Topics}\\
\toprule
\textbf{Model} & \multicolumn{1}{c}{\textbf{1e-5}} & \multicolumn{1}{c}{\textbf{5e-5}} & \multicolumn{1}{c}{\textbf{1e-4}} \\
\midrule
mBERT & 55.0\textsubscript{2.4} & \textbf{57.7}\textsubscript{1.8} & 55.5\textsubscript{1.1} \\
mDeBERTa & 44.2\textsubscript{1.7} & 54.3\textsubscript{3.1} & \textbf{57.4}\textsubscript{2.5} \\
XLM-R base & 56.0\textsubscript{1.3} & \textbf{58.1}\textsubscript{0.3} & 55.2\textsubscript{0.9} \\
XLM-R large &  \textbf{60.0}\textsubscript{0.8} & 34.7\textsubscript{23.5} & 21.1\textsubscript{0.0} \\
\bottomrule
\end{tabular}
}
\caption{%
\textbf{Effect of the learning rate on development accuracies (in~\%).}
For the intent classification task, the text models are more robust to learning rate differences than the speech models, and the ASR speech models (\textsuperscript{ASR}) are more robust than the non-ASR ones.
The best development score per model and task is bolded. 
Scores are averaged over three random seeds, with standard deviations in subscripts.
}
\label{tab:learning_rates}
\end{table}

\begin{table}
\centering
\setlength{\tabcolsep}{1pt} %
\adjustbox{max width=\linewidth}{%
\begin{tabular}{@{}lrrrrrrrr@{}}
\toprule
\textbf{ASR model} & \multicolumn{1}{c}{\textbf{F$\uparrow$}} & \multicolumn{1}{c}{\textbf{M$\uparrow$}} & \multicolumn{1}{c}{\textbf{K$\uparrow$}} & \multicolumn{1}{c}{\textbf{A$\uparrow$}} &  & \multicolumn{1}{c}{\textbf{Acc$\uparrow$}} &&  \multicolumn{1}{c}{\textbf{WER$\downarrow$}} \\
\midrule
\textit{German audio} \\
X.\ 300M DE & \cellcolor[HTML]{B2E0CA}46.4 & \cellcolor[HTML]{B2E0CA}46.4 & \cellcolor[HTML]{7DCBA5}76.8 & \cellcolor[HTML]{CCEBDB}32.0 &  & \cellcolor[HTML]{7DCBA5}76.9\textsubscript{4.6} &  & \cellcolor[HTML]{68A0D2}29.8\textsubscript{26.9} \\
MMS 1B-all & \cellcolor[HTML]{A6DBC1}53.6 & \cellcolor[HTML]{B4E1CB}45.6 & \cellcolor[HTML]{75C79F}81.6 & \cellcolor[HTML]{C3E7D6}36.8 &  & \cellcolor[HTML]{7FCBA6}75.9\textsubscript{5.3} &  & \cellcolor[HTML]{6AA1D3}30.8\textsubscript{30.0} \\
Whisper small & \cellcolor[HTML]{68C296}88.8 & \cellcolor[HTML]{8DD1B0}68.0 & \cellcolor[HTML]{65C194}90.4 & \cellcolor[HTML]{92D3B3}64.8 &  & \cellcolor[HTML]{78C9A1}79.9\textsubscript{4.5} &  & \cellcolor[HTML]{5392CC}19.3\textsubscript{28.8} \\
W.\ large-v3 & \cellcolor[HTML]{57BB8A}98.4 & \cellcolor[HTML]{71C69C}84.0 & \cellcolor[HTML]{59BC8B}97.6 & \cellcolor[HTML]{71C69C}84.0 &  & \cellcolor[HTML]{77C8A1}80.3\textsubscript{5.3} &  & \cellcolor[HTML]{3D85C6}{\color[HTML]{CCCCCC}8.6\textsubscript{14.4}} \\
\midrule
\textit{Bavarian audio}  \\
X.\ 300M DE & \cellcolor[HTML]{FFFFFF}2.4 & \cellcolor[HTML]{FEFFFF}3.2 & \cellcolor[HTML]{D1EDDF}28.8 & \cellcolor[HTML]{FFFFFF}2.4 &  & \cellcolor[HTML]{B4E1CB}45.8\textsubscript{8.7} &  & \cellcolor[HTML]{E6EFF7}90.5\textsubscript{21.7} \\
MMS 1B-all & \cellcolor[HTML]{FDFEFE}4.0 & \cellcolor[HTML]{FAFDFC}5.6 & \cellcolor[HTML]{C9E9D9}33.6 & \cellcolor[HTML]{FEFFFF}3.2 &  & \cellcolor[HTML]{B1E0C9}47.1\textsubscript{7.3} &  & \cellcolor[HTML]{E5EEF7}89.8\textsubscript{29.3} \\
Whisper small & \cellcolor[HTML]{EAF7F1}14.4 & \cellcolor[HTML]{F6FCF9}8.0 & \cellcolor[HTML]{B4E1CB}45.6 & \cellcolor[HTML]{FBFEFD}4.8 &  & \cellcolor[HTML]{A7DCC2}53.2\textsubscript{11.\rlap{1}} &  & \cellcolor[HTML]{DCE9F4}85.7\textsubscript{30.7} \\
W.\ large-v3 & \cellcolor[HTML]{D0ECDE}29.6 & \cellcolor[HTML]{E1F3EA}20.0 & \cellcolor[HTML]{A0D9BD}56.8 & \cellcolor[HTML]{EAF7F1}14.4 &  & \cellcolor[HTML]{9BD7B9}59.9\textsubscript{7.3} &  & \cellcolor[HTML]{D3E3F2}81.2\textsubscript{24.5} \\
\bottomrule
\end{tabular}}
\caption{\textbf{Manual ASR evaluation of a subset of xSID.}
All values in~\%.
Manual evaluation: F\,=\,fluency, M\,=\,meaning, K\,=\,keywords, A\,=\,all combined (F and M and K are all \textit{yes}).
The ``Acc'' column contains the intent classification accuracy of the cascaded models on the same subset, averaged over text models and random seeds.
}
\label{tab:asr-manual}
\end{table}

\section{Cascaded Setups Trained on ASR Data}
\label{sec:appendix-cascaded-additional}

\begin{table}
\setlength{\tabcolsep}{1pt} %
\adjustbox{max width=\linewidth}{%
\begin{tabular}{@{}llrrrrrrrr@{}}
\toprule
&  & \multicolumn{2}{c}{\textbf{MASSIVE}} &  & \multicolumn{5}{c}{\textbf{xSID}} \\
\cmidrule(rl){3-4} \cmidrule(rl){6-10}
\multirow{-2}{*}{\textbf{\begin{tabular}[b]{@{}l@{}}\textbf{ASR}\\ \textbf{model}\end{tabular}}} & \textbf{Train} & \multicolumn{1}{c}{\textbf{deu}} & \multicolumn{1}{c}{$\Delta$} &  & \multicolumn{1}{c}{\textbf{deu}} & \multicolumn{1}{c}{$\Delta$} &  & \multicolumn{1}{c}{\textbf{bar}} & \multicolumn{1}{c}{$\Delta$} \\
\midrule
\multicolumn{8}{@{}l}{\textit{mBERT as text model}} \\
XLS-R & gold & \cellcolor[HTML]{63C092}89.8\textsubscript{1.3} & \cellcolor[HTML]{FFFDF6} &  & \cellcolor[HTML]{7ECBA5}79.6\textsubscript{2.3} & \cellcolor[HTML]{EAD6E0} &  & \cellcolor[HTML]{B8E2CE}57.5\textsubscript{2.5} & \cellcolor[HTML]{CB98B2} \\
 & ASR & \cellcolor[HTML]{61C091}90.3\textsubscript{1.0} & \multirow{-2}{*}{\cellcolor[HTML]{FFFDF6}0.5} &  & \cellcolor[HTML]{84CDA9}77.3\textsubscript{0.7} & \multirow{-2}{*}{\cellcolor[HTML]{EAD6E0}-2.3} &  & \cellcolor[HTML]{C7E9D8}51.8\textsubscript{1.4} & \multirow{-2}{*}{\cellcolor[HTML]{CB98B2}-5.7} \\
W. tiny & gold & \cellcolor[HTML]{76C8A0}82.4\textsubscript{1.5} & \cellcolor[HTML]{FAE6A8} &  & \cellcolor[HTML]{8FD2B1}72.8\textsubscript{1.7} & \cellcolor[HTML]{FDF7E2} &  & \cellcolor[HTML]{CAEADA}50.6\textsubscript{2.9} & \cellcolor[HTML]{F9E29E} \\
 & ASR & \cellcolor[HTML]{6BC398}86.6\textsubscript{1.4} & \multirow{-2}{*}{\cellcolor[HTML]{FAE6A8}4.2} &  & \cellcolor[HTML]{8CD1AF}74.3\textsubscript{1.6} & \multirow{-2}{*}{\cellcolor[HTML]{FDF7E2}1.5} &  & \cellcolor[HTML]{BDE5D2}55.3\textsubscript{1.9} & \multirow{-2}{*}{\cellcolor[HTML]{F9E29E}4.8} \\
W. large & gold & \cellcolor[HTML]{5ABC8C}93.3\textsubscript{1.0} & \cellcolor[HTML]{FFFCF2} &  & \cellcolor[HTML]{74C79E}83.3\textsubscript{0.4} & \cellcolor[HTML]{FDF6DE} &  & \cellcolor[HTML]{9AD6B9}68.8\textsubscript{2.6} & \cellcolor[HTML]{F2E6EC} \\
 & ASR & \cellcolor[HTML]{58BC8B}93.9\textsubscript{0.0} & \multirow{-2}{*}{\cellcolor[HTML]{FFFCF2}0.6} &  & \cellcolor[HTML]{70C59B}84.9\textsubscript{0.8} & \multirow{-2}{*}{\cellcolor[HTML]{FDF6DE}1.6} &  & \cellcolor[HTML]{9ED8BB}67.4\textsubscript{1.8} & \multirow{-2}{*}{\cellcolor[HTML]{F2E6EC}-1.4} \\
\midrule
\multicolumn{8}{@{}l}{\textit{mDeBERTa as text model}} \\
XLS-R & gold & \cellcolor[HTML]{6AC398}86.9\textsubscript{1.0} & \cellcolor[HTML]{F6EEF2} &  & \cellcolor[HTML]{93D4B4}71.4\textsubscript{0.4} & \cellcolor[HTML]{EDDCE4} &  & \cellcolor[HTML]{E3F4EC}41.1\textsubscript{1.2} & \cellcolor[HTML]{F6D777} \\
 & ASR & \cellcolor[HTML]{6DC499}86.0\textsubscript{3.7} & \multirow{-2}{*}{\cellcolor[HTML]{F6EEF2}-0.9} &  & \cellcolor[HTML]{98D6B8}69.5\textsubscript{5.3} & \multirow{-2}{*}{\cellcolor[HTML]{EDDCE4}-1.9} &  & \cellcolor[HTML]{D2EDDF}47.7\textsubscript{2.0} & \multirow{-2}{*}{\cellcolor[HTML]{F6D777}6.6} \\
W. tiny & gold & \cellcolor[HTML]{85CEAA}76.6\textsubscript{0.9} & \cellcolor[HTML]{FAF5F7} &  & \cellcolor[HTML]{ABDDC5}62.3\textsubscript{1.7} & \cellcolor[HTML]{F5ECF0} &  & \cellcolor[HTML]{FFFFFF}30.3\textsubscript{1.1} & \cellcolor[HTML]{F2C63F} \\
 & ASR & \cellcolor[HTML]{87CFAC}76.1\textsubscript{3.2} & \multirow{-2}{*}{\cellcolor[HTML]{FAF5F7}-0.6} &  & \cellcolor[HTML]{AEDEC7}61.2\textsubscript{4.6} & \multirow{-2}{*}{\cellcolor[HTML]{F5ECF0}-1.1} &  & \cellcolor[HTML]{E7F6EE}39.6\textsubscript{2.1} & \multirow{-2}{*}{\cellcolor[HTML]{F2C63F}9.4} \\
W. large & gold & \cellcolor[HTML]{5BBD8D}92.7\textsubscript{1.0} & \cellcolor[HTML]{CA95AF} &  & \cellcolor[HTML]{84CEAA}77.0\textsubscript{1.0} & \cellcolor[HTML]{A74F7A} &  & \cellcolor[HTML]{C0E6D3}54.4\textsubscript{2.9} & \cellcolor[HTML]{EEDDE6} \\
 & ASR & \cellcolor[HTML]{6BC398}86.8\textsubscript{4.7} & \multirow{-2}{*}{\cellcolor[HTML]{CA95AF}-5.9} &  & \cellcolor[HTML]{9ED8BC}67.2\textsubscript{7.4} & \multirow{-2}{*}{\cellcolor[HTML]{A74F7A}-9.9} &  & \cellcolor[HTML]{C5E8D7}52.5\textsubscript{2.1} & \multirow{-2}{*}{\cellcolor[HTML]{EEDDE6}-1.9} \\
 \bottomrule
\end{tabular}}
\caption{\textbf{Performance differences between cascaded models that were trained on gold-standard vs.\ automatically transcribed data.}
Each row shows the test-set performance of one cascaded setup (accuracy, in~\%, averaged over three seeds with standard deviations in subscripts).
The $\Delta$~columns show the differences in percentage points: trained on data transcribed by the same ASR model as the evaluation data minus trained on gold text data.
ASR models: XLS-R 300M DE, Whisper tiny, Whisper large-v3.
}
\label{tab:cascaded-additional}
\end{table}

\begin{table}
\setlength{\tabcolsep}{1pt} %
\adjustbox{max width=\linewidth}{%
\begin{tabular}{@{}llrrrrrr@{\hspace{5pt}}rrrrr@{}}
\toprule
& & \multicolumn{5}{c}{$\Delta$ \textbf{speech--casc.}} &  & \multicolumn{5}{c}{$\Delta$ \textbf{cascaded--text}} \\
\cmidrule(rl){3-7} \cmidrule(rl){9-13}
\textbf{Text} & \textbf{Speech/} & {\textbf{MAS.}} &  & {\textbf{xSID}} &  & {\textbf{xSID}} &  & {\textbf{MAS.}} &  & {\textbf{xSID}} &  & {\textbf{xSID}} \\
\textbf{model} & \textbf{ASR mod.} & {\textbf{deu}} &  & {\textbf{deu}} &  & {\textbf{bar}} &  & {\textbf{deu}} &  & {\textbf{deu}} &  & {\textbf{bar}} \\
\midrule
\multicolumn{13}{@{}l}{\textit{Train on gold text data, evaluate on ASR data}} \\
mBERT & XLS-R & \cellcolor[HTML]{FFFDF7}1.0 &  & \cellcolor[HTML]{F2E5EB}-3.6 &  & \cellcolor[HTML]{FBEBBA}8.5 &  & \cellcolor[HTML]{EFE0E8}-4.2 &  & \cellcolor[HTML]{F1E3EA}-3.9 &  & \cellcolor[HTML]{E4CAD7}-7.4 \\
 & W.\ tiny & \cellcolor[HTML]{D7B0C3}-11.1 &  & \cellcolor[HTML]{D0A1B8}-13.2 &  & \cellcolor[HTML]{F4E9EE}-3.0 &  & \cellcolor[HTML]{D5ABC0}-11.7 &  & \cellcolor[HTML]{D8B2C5}-10.7 &  & \cellcolor[HTML]{CB98B1}-14.4 \\
 & W.\ large & \cellcolor[HTML]{F8F1F5}-1.8 &  & \cellcolor[HTML]{FCFAFB}-0.6 &  & \cellcolor[HTML]{FDF6E1}3.7 &  & \cellcolor[HTML]{FCF9FA}-0.8 &  & \cellcolor[HTML]{FEFDFD}-0.2 &  & \cellcolor[HTML]{FDF6E0}3.8 \\
 \\[-10pt]
mDeB. & XLS-R & \cellcolor[HTML]{FDF6DF}4.0 &  & \cellcolor[HTML]{FDF4DA}4.6 &  & \cellcolor[HTML]{F2C333}24.9 &  & \cellcolor[HTML]{E7CFDB}-6.6 &  & \cellcolor[HTML]{E4C9D6}-7.5 &  & \cellcolor[HTML]{CFA0B7}-13.3 \\
 & W.\ tiny & \cellcolor[HTML]{EBD8E2}-5.4 &  & \cellcolor[HTML]{F5EBF0}-2.7 &  & \cellcolor[HTML]{F6D572}17.3 &  & \cellcolor[HTML]{C286A4}-16.9 &  & \cellcolor[HTML]{C388A5}-16.7 &  & \cellcolor[HTML]{A9537D}-24.1 \\
 & W.\ large & \cellcolor[HTML]{FAF5F8}-1.3 &  & \cellcolor[HTML]{FCF2D1}5.7 &  & \cellcolor[HTML]{F5D36B}18.1 &  & \cellcolor[HTML]{FCF9FA}-0.8 &  & \cellcolor[HTML]{F8F1F4}-1.9 &  & \cellcolor[HTML]{FFFFFF}0.0 \\
 \midrule
\multicolumn{13}{@{}l}{\textit{Train/eval on data transcribed by the same ASR model}} \\
mBERT & XLS-R & \cellcolor[HTML]{FFFEFB}0.6 &  & \cellcolor[HTML]{FAF5F8}-1.3 &  & \cellcolor[HTML]{F8DD8B}14.2 &  & \cellcolor[HTML]{F1E4EA}-3.8 &  & \cellcolor[HTML]{E9D3DE}-6.1 &  & \cellcolor[HTML]{D0A1B8}-13.2 \\
 & W.\ tiny & \cellcolor[HTML]{C891AC}-15.3 &  & \cellcolor[HTML]{CA96B0}-14.6 &  & \cellcolor[HTML]{E3C7D5}-7.8 &  & \cellcolor[HTML]{E4C9D6}-7.5 &  & \cellcolor[HTML]{DEBDCD}-9.2 &  & \cellcolor[HTML]{DCBACB}-9.6 \\
 & W.\ large & \cellcolor[HTML]{F6EDF1}-2.5 &  & \cellcolor[HTML]{F7EFF3}-2.2 &  & \cellcolor[HTML]{FDF3D6}5.1 &  & \cellcolor[HTML]{FEFDFE}-0.2 &  & \cellcolor[HTML]{FFFCF4}1.4 &  & \cellcolor[HTML]{FEFAEC}2.4 \\
  \\[-10pt]
mDeB. & XLS-R & \cellcolor[HTML]{FDF4D7}4.9 &  & \cellcolor[HTML]{FCF0CA}6.6 &  & \cellcolor[HTML]{F5D36A}18.3 &  & \cellcolor[HTML]{E4C9D6}-7.6 &  & \cellcolor[HTML]{DDBBCC}-9.5 &  & \cellcolor[HTML]{E7CFDB}-6.6 \\
 & W.\ tiny & \cellcolor[HTML]{EDDCE5}-4.8 &  & \cellcolor[HTML]{F9F3F6}-1.6 &  & \cellcolor[HTML]{FBECBE}7.9 &  & \cellcolor[HTML]{C082A1}-17.5 &  & \cellcolor[HTML]{BF80A0}-17.7 &  & \cellcolor[HTML]{CA96B0}-14.7 \\
 & W.\ large & \cellcolor[HTML]{FDF4DA}4.6 &  & \cellcolor[HTML]{F7DA80}15.5 &  & \cellcolor[HTML]{F4CF5C}20.0 &  & \cellcolor[HTML]{E7CFDA}-6.7 &  & \cellcolor[HTML]{D4AABF}-11.8 &  & \cellcolor[HTML]{F8F1F5}-1.9 \\
\bottomrule
\end{tabular}}
\caption{\textbf{The overall trends are similar for cascaded systems that were fine-tuned on gold-standard text data (top) vs.\ automatically transcribed data (bottom).}
Accuracy differences between setups, in percentage points.
Text models: mBERT and mDeBERTa, speech/ASR models: XLS-R 300M DE, Whisper tiny, Whisper large-v3.
}
\label{tab:cascaded-additional-deltas}
\end{table}

When testing on automatically transcribed data, it can make a difference whether the model was fine-tuned on gold text data or on transcriptions produced by the same ASR model \cite{lee-etal-2018-spoken, phung-etal-2024-ar-nlu}.

To supplement our experiments on cascaded setups, we additionally train six cascaded systems (three ASR models $\times$ two text models) on training data that was automatically transcribed by the same ASR models as the evaluation data.
We select this subset since training on all combinations of ASR and text models is not feasible.
We use the two text models with the best and worst performance on the Bavarian data (mBERT and mDeBERTa, respectively).
For the ASR models, we select the overall best and worst ASR model (Whisper large-v3 and tiny, respectively), and additionally XLS-R 300M DE so as to also include a CTC-based ASR model.
We only carry out these supplemental experiments on the intent classification data, since no spoken training data is available for the topic classification dataset.

How big the performance difference is between the cascaded models trained on gold-standard vs.\ automatically transcribed data -- and which version actually yields the better results -- varies across the ASR and text model combinations (Table~\ref{tab:cascaded-additional}). 
While the ASR errors in the training data help some models better generalize to the evaluation data, this is not the case for all models: the ASR transcriptions might on the one hand introduce infelicitous noise into the training data, and on the other hand, the types of ASR errors can be different across data splits (see~\S\ref{sec:appendix-asr-examples}). 

Comparing the performance differences between cascaded and text-/speech-only models results in somewhat different deltas, but the trends are the same regardless of whether the cascaded models are trained on gold or transcribed text (Table~\ref{tab:cascaded-additional-deltas}). As in the overall analysis (\S\ref{sec:analysis}, Table~\ref{tab:deltas}), the speech-only setups generally outperform the cascaded ones on the Bavarian test data, but not necessarily on the German data. Most cascaded-setups are outperformed by the text-only models on both German and Bavarian.

Thus, while training cascaded models on automatically transcribed data results in somewhat different predictions than when training them on gold text data, the overall patterns stay the same.

\section{Details on ASR Results}
\subsection{Manual Evaluation of ASR Results}
\label{sec:appendix-asr-manual}

We annotate the ASR hypotheses of four models (\xlsr{} 300M DE, MMS 1B-all, Whisper small, and Whisper large-v3) for the first~75 and last~50 instances of the xSID test set in German and Bavarian based on the following criteria:

\begin{itemize}
    \item \textbf{Fluency:} Is the ASR hypothesis in fluent, grammatical, and orthographically correct German? Colloquial and regionally marked language is okay, major spelling errors are not (unless they concern named entities).
    \item \textbf{Meaning:} Does the hypothesis have the same meaning as the reference? Spelling errors are okay as long as it would still be clear out of context what the sentence means, but not if they concern key entities (i.e., we keep slot filling in mind).
    \item \textbf{Keywords:} Does the hypothesis contain obvious clues for the intent label? This can be the case even if the more nuanced meaning of the sentence did not get preserved. 
    The keywords signalling the intent should be spelled correctly (so as to match the spelling in the training data) and should not be merged with other words in the hypothesis due to wrong word segmentation, unless they occur at the beginning of the merged word.\\
    This measure is from the annotator's perspective, but the pertinent keywords (keyword subtokens) / combinations thereof learned by the models might be different.
    This measure is inspired by \citet{choi-etal-2024-self-supervised}, whose work suggests that keyword spotting is often sufficient for intent classification.
\end{itemize}

\noindent
We annotate each instance with \textit{yes} or \textit{no} for each of the criteria.
If in doubt, we choose \textit{no}.
We ignore capitalization (since the model input is lowercased, \S\ref{sec:data}).
We also ignore punctuation.
For each measure, we calculate the proportion of instances that were annotated as \textit{yes}.
The annotator is a native speaker of German who is familiar with Bavarian and with the dataset.

\paragraph{Results}
Table~\ref{tab:asr-manual} shows the results.
All of these measures are highly correlated with each other and nearly always show the same model ranking.
However, the \textit{keyword} measure almost always has the highest score, and it is always much higher than the \textit{meaning} score (with differences of 13.6--37.6\,pp.).
This shows that even if the meaning of the transcribed sentence does not match the original meaning, the sentence often still contains keywords related to the intent.
This is even the case for Bavarian sentences whose transcriptions neither match the original meaning nor are in fluent German.

\subsection{ASR Sample Outputs}
\label{sec:appendix-asr-examples}

Table~\ref{tab:asr-examples} shows ASR sample transcriptions.
While the German transcriptions are usually understandable, the transcriptions for Bavarian audio contain many more mistakes.

\begin{table}[]
\adjustbox{max width=\linewidth}{%
\begin{tabular}{@{}ll@{}}
\toprule
\multicolumn{2}{@{}l}{“Show (me) all reminders”} \\[4pt]
DEU & Zeige alle Erinnerungen \\
W.\ lg & Zeige alle Erinnerungen. \\
W.\ tiny & Zeige \wrong{aller} Erinnerungen.\\
& (“show all.GEN reminders”) \\
XLS-R & zeige alle erinnerungen \\[4pt]
BAR & Zoag ma olle Erinnerungen \\
W.\ lg & \wrong{Zeug} mir alle Erinnerungen.\\
&(``beget me all reminders'') \\
W.\ tiny & \wrong{Talk mal} alle \wrong{ja in der Wungen}.\\
& (“{[}nonce{]} once all yes in the {[}nonce{]}”) \\
XLS-R & \wrong{zorgmal} \barok{olle} erinnerungen\\
& (“{[}nonce{]} {[}`old' in DEU\,/\,‘all’ in BAR{]}\\
& reminders”) \\
\midrule
\multicolumn{2}{@{}l}{“Add a reminder for 4pm today”} \\[4pt]
DEU & Eine Erinnerung für heute um 4\\
& Uhr nachmittags hinzufügen \\
W.\ lg & Eine Erinnerung für heute um 4\\
&Uhr nachmittags hinzufügen. \\
W.\ tiny & Eine Erinnerung für heute um vier\\
&Uhr nachmittags hinzufügen. \\
XLS-R & eine erinnerung für heute um vier\\
&uhr nachmittags hinzufügen \\[4pt]
BAR & Stei ma a Erinnerung füa heid\\
&um viere Nammiddog ei \\
W.\ lg & Stell \wrong{mal} Erinnerung für \wrong{Heidumphiri}\\
& \wrong{Namidogai}. (``set reminder for {[}nonce{]}\\
&{[}nonce{]}'') \\
W.\ tiny & \wrong{Stämmerer} Erinnerung für \wrong{Heidung}\\
& 4 \wrong{Nm}. (“{[}nonce{]} reminder for {[}nonce{]}\\
&4 {[}nonce{]}”) \\
XLS-R & \wrong{steime}erinnerung für \wrong{heidung}\\
& \wrong{virenamidock} \barok{ei} (“{[}nonce{]}reminder for {[}nonce{]}\\
& {[}nonce{]} {[}nonce in DEU\,/\,`in' in BAR{]}”) \\
\midrule
\multicolumn{2}{@{}l}{``Show all alarms''} \\[4pt]
DEU & Zeige alle Wecker \\
W.\ lg & Zeige alle Wecker. \\
W.\ tiny & Zeige alle Wecker. \\
XLS-R & zeige alle \wrong{weger} (“show all {[}nonce{]}”) \\[4pt]
BAR & Zoag alle Wegga \\
W.\ lg & \wrong{Zwerg-Olewecke} (``dwarf-{[}nonce{]}'') \\
W.\ tiny & \wrong{Zuhack und lege}. (“{[}nonce{]} and lay”) \\
XLS-R & \wrong{zorg ollyvicker} (“{[}nonce{]} {[}nonce{]}”) \\
\bottomrule
\end{tabular}}
\caption{\textbf{Sample gold sentences and ASR outputs} for German (DEU) and Bavarian (BAR) audios by Whisper large-v3, Whisper-tiny, and XLS-R 300M DE.
Wrong transcriptions are in \wrong{red}; transcriptions that are wrong in German but accurate in Bavarian are in \barok{blue}.
}
\label{tab:asr-examples}
\end{table}

\begin{table*}[t]
\centering
\setlength{\tabcolsep}{3pt} %
\adjustbox{max width=\linewidth}{%
\begin{tabular}{@{}lrrrrrrrrrrrrrrrrrrr@{}}
\toprule
\textbf{Model} & \multicolumn{1}{c}{\textbf{deu}} &  & \multicolumn{1}{c}{\textbf{gsw}} &  & \multicolumn{1}{c}{\textbf{ag}} &  & \multicolumn{1}{c}{\textbf{be}} &  & \multicolumn{1}{c}{\textbf{bs}} &  & \multicolumn{1}{c}{\textbf{gr}} &  & \multicolumn{1}{c}{\textbf{lu}} &  & \multicolumn{1}{c}{\textbf{sg}} &  & \multicolumn{1}{c}{\textbf{vs}} &  & \multicolumn{1}{c}{\textbf{zh}} \\
\midrule
\textit{Text-only} & \textit{54.0\textsubscript{3.3}} &  & \textit{43.7\textsubscript{3.8}} &  & \textit{41.9\textsubscript{3.3}} &  & \textit{40.0\textsubscript{4.5}} &  & \textit{46.4\textsubscript{3.9}} &  & \textit{46.3\textsubscript{4.7}} &  & \textit{41.8\textsubscript{4.2}} &  & \textit{45.6\textsubscript{4.4}} &  & \textit{42.7\textsubscript{3.7}} &  & \textit{44.9\textsubscript{3.4}} \\
mBERT & {\cellcolor[HTML]{76C8A0}52.8\textsubscript{2.1}} &  & \cellcolor[HTML]{B6E2CC}41.2\textsubscript{1.0} &  & \cellcolor[HTML]{BEE5D2}39.7\textsubscript{0.6} &  & \cellcolor[HTML]{CBEADB}37.5\textsubscript{2.1} &  & \cellcolor[HTML]{A7DCC2}43.9\textsubscript{0.8} &  & \cellcolor[HTML]{ABDDC4}43.3\textsubscript{1.2} &  & \cellcolor[HTML]{BFE5D2}39.6\textsubscript{2.4} &  & \cellcolor[HTML]{B1E0C9}42.2\textsubscript{1.9} &  & \cellcolor[HTML]{B7E2CD}41.0\textsubscript{0.8} &  & \cellcolor[HTML]{AEDFC7}42.6\textsubscript{1.1} \\
mDeBERTa & {\cellcolor[HTML]{7ECBA5}51.4\textsubscript{1.2}} &  & \cellcolor[HTML]{A9DCC3}43.6\textsubscript{2.6} &  & \cellcolor[HTML]{B8E3CE}40.7\textsubscript{1.4} &  & \cellcolor[HTML]{B7E2CD}41.1\textsubscript{1.5} &  & \cellcolor[HTML]{91D3B2}48.0\textsubscript{4.1} &  & \cellcolor[HTML]{9FD8BC}45.5\textsubscript{2.8} &  & \cellcolor[HTML]{B8E2CD}40.9\textsubscript{2.7} &  & \cellcolor[HTML]{9ED8BC}45.5\textsubscript{2.9} &  & \cellcolor[HTML]{B6E2CC}41.2\textsubscript{2.4} &  & \cellcolor[HTML]{9FD8BC}45.5\textsubscript{3.5} \\
XLM-R base & {\cellcolor[HTML]{73C79E}53.3\textsubscript{3.2}} &  & \cellcolor[HTML]{B5E1CC}41.3\textsubscript{4.0} &  & \cellcolor[HTML]{B9E3CE}40.7\textsubscript{4.2} &  & \cellcolor[HTML]{D4EEE1}35.7\textsubscript{4.1} &  & \cellcolor[HTML]{AADDC4}43.4\textsubscript{4.2} &  & \cellcolor[HTML]{A6DBC1}44.1\textsubscript{5.5} &  & \cellcolor[HTML]{C1E6D4}39.2\textsubscript{3.6} &  & \cellcolor[HTML]{AADDC4}43.4\textsubscript{4.4} &  & \cellcolor[HTML]{B5E1CC}41.3\textsubscript{4.7} &  & \cellcolor[HTML]{AEDEC7}42.7\textsubscript{2.8} \\
XLM-R large & {\cellcolor[HTML]{57BB8A}58.3\textsubscript{1.3}} &  & \cellcolor[HTML]{8DD1AF}48.7\textsubscript{1.0} &  & \cellcolor[HTML]{99D6B8}46.4\textsubscript{0.6} &  & \cellcolor[HTML]{9ED8BB}45.6\textsubscript{1.9} &  & \cellcolor[HTML]{84CEAA}50.3\textsubscript{1.7} &  & \cellcolor[HTML]{78C9A1}52.4\textsubscript{1.5} &  & \cellcolor[HTML]{93D4B4}47.5\textsubscript{1.7} &  & \cellcolor[HTML]{7ECBA5}51.4\textsubscript{1.0} &  & \cellcolor[HTML]{95D4B5}47.1\textsubscript{2.3} &  & \cellcolor[HTML]{8BD1AF}48.9\textsubscript{0.6} \\
\midrule
\textit{Cascaded} &  &  & \textit{42.9\textsubscript{8.7}} &  & \textit{42.3\textsubscript{9.4}} &  & \textit{43.1\textsubscript{9.9}} &  & \textit{43.7\textsubscript{7.8}} &  & \textit{45.6\textsubscript{7.5}} &  & \textit{42.4\textsubscript{9.3}} &  & \textit{43.8\textsubscript{7.9}} &  & \textit{39.9\textsubscript{9.3}} &  & \textit{42.7\textsubscript{9.7}} \\
\multicolumn{8}{@{}l}{\textit{... averaged over ASR models}} \\
mBERT &  &  & \cellcolor[HTML]{B4E1CB}41.5\textsubscript{8.5} &  & \cellcolor[HTML]{BBE4D0}40.2\textsubscript{9.3} &  & \cellcolor[HTML]{B2E0CA}41.9\textsubscript{9.7} &  & \cellcolor[HTML]{B0DFC8}42.2\textsubscript{7.6} &  & \cellcolor[HTML]{A4DBC0}44.4\textsubscript{7.2} &  & \cellcolor[HTML]{BAE3CF}40.5\textsubscript{8.9} &  & \cellcolor[HTML]{ACDEC6}42.9\textsubscript{7.7} &  & \cellcolor[HTML]{C3E7D5}38.9\textsubscript{9.3} &  & \cellcolor[HTML]{B6E2CC}41.2\textsubscript{9.2} \\
mDeBERTa &  &  & \cellcolor[HTML]{B8E3CE}40.8\textsubscript{8.8} &  & \cellcolor[HTML]{BEE5D2}39.7\textsubscript{9.0} &  & \cellcolor[HTML]{B5E1CC}41.3\textsubscript{10.\rlap{2}} &  & \cellcolor[HTML]{B2E0C9}42.0\textsubscript{7.6} &  & \cellcolor[HTML]{AADDC4}43.4\textsubscript{7.9} &  & \cellcolor[HTML]{BDE5D1}39.9\textsubscript{9.6} &  & \cellcolor[HTML]{B5E1CC}41.4\textsubscript{8.0} &  & \cellcolor[HTML]{C6E8D7}38.3\textsubscript{9.2} &  & \cellcolor[HTML]{B9E3CE}40.7\textsubscript{9.7} \\
XLM-R base &  &  & \cellcolor[HTML]{ABDDC5}43.1\textsubscript{8.4} &  & \cellcolor[HTML]{ADDEC6}42.9\textsubscript{9.1} &  & \cellcolor[HTML]{ADDEC6}42.9\textsubscript{10.\rlap{0}} &  & \cellcolor[HTML]{A6DBC1}44.1\textsubscript{7.9} &  & \cellcolor[HTML]{9FD8BC}45.4\textsubscript{7.2} &  & \cellcolor[HTML]{AEDFC7}42.6\textsubscript{8.3} &  & \cellcolor[HTML]{AADDC4}43.4\textsubscript{7.4} &  & \cellcolor[HTML]{BBE4D0}40.3\textsubscript{9.3} &  & \cellcolor[HTML]{AADDC4}43.3\textsubscript{9.8} \\
XLM-R large &  &  & \cellcolor[HTML]{9AD6B9}46.3\textsubscript{8.6} &  & \cellcolor[HTML]{9AD6B9}46.3\textsubscript{9.4} &  & \cellcolor[HTML]{9AD6B9}46.3\textsubscript{9.5} &  & \cellcolor[HTML]{99D6B8}46.4\textsubscript{7.9} &  & \cellcolor[HTML]{89D0AD}49.3\textsubscript{6.7} &  & \cellcolor[HTML]{97D5B7}46.7\textsubscript{9.5} &  & \cellcolor[HTML]{94D4B5}47.4\textsubscript{7.6} &  & \cellcolor[HTML]{B0DFC8}42.3\textsubscript{9.5} &  & \cellcolor[HTML]{9FD8BC}45.4\textsubscript{10.\rlap{0}} \\
\multicolumn{8}{@{}l}{\textit{... averaged over text models}} \\
XLS-R \rlap{300M DE} &  &  & \cellcolor[HTML]{E8F6EF}32.1\textsubscript{3.1} &  & \cellcolor[HTML]{EEF9F3}31.0\textsubscript{3.1} &  & \cellcolor[HTML]{F0F9F5}30.6\textsubscript{3.3} &  & \cellcolor[HTML]{DCF1E7}34.3\textsubscript{3.4} &  & \cellcolor[HTML]{D0ECDE}36.5\textsubscript{3.2} &  & \cellcolor[HTML]{E5F5ED}32.6\textsubscript{4.5} &  & \cellcolor[HTML]{E0F3EA}33.5\textsubscript{3.8} &  & \cellcolor[HTML]{FFFFFF}27.9\textsubscript{3.2} &  & \cellcolor[HTML]{F1F9F5}30.6\textsubscript{3.1} \\
XLS-R 1B DE &  &  & \cellcolor[HTML]{E3F4EB}33.1\textsubscript{3.5} &  & \cellcolor[HTML]{E3F4EC}33.0\textsubscript{4.7} &  & \cellcolor[HTML]{EEF9F3}31.0\textsubscript{3.7} &  & \cellcolor[HTML]{CFECDE}36.6\textsubscript{4.2} &  & \cellcolor[HTML]{CCEBDC}37.2\textsubscript{4.2} &  & \cellcolor[HTML]{E2F4EB}33.2\textsubscript{3.9} &  & \cellcolor[HTML]{D6EFE2}35.5\textsubscript{3.6} &  & \cellcolor[HTML]{FEFFFE}28.2\textsubscript{3.2} &  & \cellcolor[HTML]{F3FAF7}30.1\textsubscript{3.1} \\
MMS 1B-all &  &  & \cellcolor[HTML]{CCEBDB}37.3\textsubscript{3.3} &  & \cellcolor[HTML]{CEEBDD}36.9\textsubscript{3.9} &  & \cellcolor[HTML]{D7EFE4}35.1\textsubscript{3.8} &  & \cellcolor[HTML]{C4E8D6}38.6\textsubscript{2.5} &  & \cellcolor[HTML]{B0DFC8}42.3\textsubscript{4.6} &  & \cellcolor[HTML]{D9F0E5}34.8\textsubscript{3.6} &  & \cellcolor[HTML]{B5E1CC}41.4\textsubscript{3.8} &  & \cellcolor[HTML]{E2F4EB}33.2\textsubscript{2.9} &  & \cellcolor[HTML]{D3EEE1}35.9\textsubscript{3.2} \\
Whisper tiny &  &  & \cellcolor[HTML]{D5EEE2}35.6\textsubscript{2.3} &  & \cellcolor[HTML]{E6F5ED}32.6\textsubscript{2.7} &  & \cellcolor[HTML]{D5EEE2}35.6\textsubscript{3.1} &  & \cellcolor[HTML]{D3EDE0}36.0\textsubscript{2.4} &  & \cellcolor[HTML]{BEE5D2}39.7\textsubscript{3.8} &  & \cellcolor[HTML]{DBF1E6}34.5\textsubscript{2.7} &  & \cellcolor[HTML]{CFECDE}36.6\textsubscript{2.7} &  & \cellcolor[HTML]{DCF1E7}34.3\textsubscript{2.3} &  & \cellcolor[HTML]{D7EFE3}35.2\textsubscript{2.2} \\
Whisper base &  &  & \cellcolor[HTML]{B0DFC8}42.2\textsubscript{2.3} &  & \cellcolor[HTML]{B4E1CB}41.6\textsubscript{3.2} &  & \cellcolor[HTML]{A1D9BE}44.9\textsubscript{2.7} &  & \cellcolor[HTML]{B2E0C9}42.0\textsubscript{2.6} &  & \cellcolor[HTML]{A7DCC2}44.0\textsubscript{2.6} &  & \cellcolor[HTML]{BBE4D0}40.2\textsubscript{3.4} &  & \cellcolor[HTML]{B3E1CA}41.8\textsubscript{2.5} &  & \cellcolor[HTML]{BDE5D1}39.9\textsubscript{3.0} &  & \cellcolor[HTML]{AADDC4}43.4\textsubscript{3.1} \\
Whisper small &  &  & \cellcolor[HTML]{8AD0AE}49.2\textsubscript{2.5} &  & \cellcolor[HTML]{8FD2B1}48.2\textsubscript{4.2} &  & \cellcolor[HTML]{7FCCA6}51.2\textsubscript{2.4} &  & \cellcolor[HTML]{8AD0AE}49.2\textsubscript{2.8} &  & \cellcolor[HTML]{81CCA8}50.7\textsubscript{2.5} &  & \cellcolor[HTML]{8DD1B0}48.6\textsubscript{2.8} &  & \cellcolor[HTML]{89D0AD}49.3\textsubscript{3.2} &  & \cellcolor[HTML]{9FD9BD}45.3\textsubscript{2.2} &  & \cellcolor[HTML]{7FCCA6}51.1\textsubscript{2.0} \\
Whisper \rlap{medium} &  &  & \cellcolor[HTML]{7CCAA4}51.7\textsubscript{2.9} &  & \cellcolor[HTML]{7ACAA3}52.0\textsubscript{3.5} &  & \cellcolor[HTML]{79C9A2}52.3\textsubscript{2.8} &  & \cellcolor[HTML]{80CCA6}51.1\textsubscript{2.6} &  & \cellcolor[HTML]{79C9A2}52.3\textsubscript{2.6} &  & \cellcolor[HTML]{79C9A2}52.3\textsubscript{4.2} &  & \cellcolor[HTML]{7FCCA6}51.1\textsubscript{3.1} &  & \cellcolor[HTML]{85CEAA}50.0\textsubscript{3.3} &  & \cellcolor[HTML]{78C9A1}52.5\textsubscript{2.5} \\
Whisper \rlap{large-v3-turbo} &  &  & \cellcolor[HTML]{77C8A1}52.6\textsubscript{3.1} &  & \cellcolor[HTML]{78C9A1}52.5\textsubscript{3.9} &  & \cellcolor[HTML]{71C69C}53.7\textsubscript{2.9} &  & \cellcolor[HTML]{78C9A1}52.5\textsubscript{2.9} &  & \cellcolor[HTML]{70C69C}53.9\textsubscript{2.7} &  & \cellcolor[HTML]{76C89F}52.9\textsubscript{4.0} &  & \cellcolor[HTML]{78C9A1}52.4\textsubscript{3.3} &  & \cellcolor[HTML]{86CEAB}50.0\textsubscript{2.9} &  & \cellcolor[HTML]{76C8A0}52.8\textsubscript{3.5} \\
Whisper \rlap{large-v3} &  &  & \cellcolor[HTML]{76C8A0}52.7\textsubscript{3.1} &  & \cellcolor[HTML]{77C8A0}52.6\textsubscript{3.8} &  & \cellcolor[HTML]{72C69D}53.5\textsubscript{2.9} &  & \cellcolor[HTML]{75C79F}53.0\textsubscript{3.0} &  & \cellcolor[HTML]{6FC59B}54.1\textsubscript{3.1} &  & \cellcolor[HTML]{75C79F}53.0\textsubscript{4.2} &  & \cellcolor[HTML]{78C9A1}52.4\textsubscript{3.2} &  & \cellcolor[HTML]{81CCA8}50.8\textsubscript{2.5} &  & \cellcolor[HTML]{78C9A1}52.5\textsubscript{3.1} \\
\bottomrule
\end{tabular}}
\caption{\textbf{Topic classification accuracies for German, Swiss German (all dialects), and Swiss German by region.}
Key:
deu\,=\,German, gsw\,=\,Swiss German (all of the following dialects),
ag\,=\,Aargau, be\,=\,Bern, bs\,=\,Basel, gr\,=\,Grisons, lu\,=\,Lucerne, sg\,=\,St.\ Gallen, vs\,=\,Valais, zh\,=\,Zurich.
Accuracies in~\%, averaged over three random seeds (standard deviations in subscripts).
Average performances per setup type are in \textit{italics}.
}
\label{tab:dialects-topics}
\end{table*}

\begin{table*}
\centering
\setlength{\tabcolsep}{1pt} %
\adjustbox{max width=\textwidth}{%
\begin{tabular}{@{}l@{\hspace{5pt}}l@{\hspace{8pt}}rr@{\hspace{10pt}}rrrrrr@{\hspace{25pt}}r@{}}
\toprule
&& \multicolumn{7}{c}{\textbf{Intents}} &  & \multicolumn{1}{c}{\textbf{Topics}} \\
\cmidrule(rl){3-9} \cmidrule(rl){11-11}
&& \multicolumn{1}{c}{\textbf{\llap{MA}SSI\rlap{VE}}} && \multicolumn{5}{c}{\textbf{xSID}} &  & \multicolumn{1}{c}{\textbf{\llap{Sw}issd\rlap{ial}}} \\
\cmidrule(rl){3-3} \cmidrule(rl){5-9} \cmidrule(rl){11-11}
\textbf{Text model} & \textbf{ASR model} & \multicolumn{1}{c}{\textbf{deu}} &  & \multicolumn{1}{c}{\textbf{deu}} &  & \multicolumn{1}{c}{\textbf{bar}} &  & \multicolumn{1}{c}{$\Delta$} &  & \multicolumn{1}{c}{\textbf{gsw}} \\
\cmidrule(r){1-2} \cmidrule{3-9} \cmidrule{11-11}
mBERT & XLS-R 300M DE & \cellcolor[HTML]{63C092}89.8\textsubscript{1.3} &  & \cellcolor[HTML]{7ECBA5}79.6\textsubscript{2.3} &  & \cellcolor[HTML]{B8E2CE}57.5\textsubscript{2.5} &  & \cellcolor[HTML]{F0C2AE}-22.1 &  & \cellcolor[HTML]{FCFEFD}30.6\textsubscript{0.8} \\
& XLS-R 1B DE & \cellcolor[HTML]{60BF91}90.7\textsubscript{0.4} &  & \cellcolor[HTML]{75C79F}82.8\textsubscript{1.0} &  & \cellcolor[HTML]{B7E2CD}57.8\textsubscript{0.6} &  & \cellcolor[HTML]{EDB89F}-25.0 &  & \cellcolor[HTML]{F7FCFA}31.4\textsubscript{0.7} \\
& MMS 1B-all & \cellcolor[HTML]{61BF91}90.5\textsubscript{1.4} &  & \cellcolor[HTML]{7FCBA6}79.1\textsubscript{1.3} &  & \cellcolor[HTML]{C1E6D4}54.0\textsubscript{2.1} &  & \cellcolor[HTML]{EDB79F}-25.1 &  & \cellcolor[HTML]{E0F3EA}35.3\textsubscript{0.5} \\
& Whisper tiny & \cellcolor[HTML]{76C8A0}82.4\textsubscript{1.5} &  & \cellcolor[HTML]{8FD2B1}72.8\textsubscript{1.7} &  & \cellcolor[HTML]{CAEADA}50.6\textsubscript{2.9} &  & \cellcolor[HTML]{F0C2AD}-22.2 &  & \cellcolor[HTML]{E4F4EC}34.6\textsubscript{0.9} \\
& Whisper base & \cellcolor[HTML]{6DC49A}85.8\textsubscript{2.1} &  & \cellcolor[HTML]{84CEA9}77.2\textsubscript{2.0} &  & \cellcolor[HTML]{BBE4D0}56.2\textsubscript{1.2} &  & \cellcolor[HTML]{F1C7B4}-21.0 &  & \cellcolor[HTML]{BDE5D1}41.2\textsubscript{1.2} \\
& Whisper small & \cellcolor[HTML]{60BF90}90.9\textsubscript{1.0} &  & \cellcolor[HTML]{77C8A0}82.2\textsubscript{1.4} &  & \cellcolor[HTML]{9DD8BB}67.6\textsubscript{2.5} &  & \cellcolor[HTML]{F7DED3}-14.6 &  & \cellcolor[HTML]{96D5B6}47.8\textsubscript{1.0} \\
& Whisper medium & \cellcolor[HTML]{5CBD8E}92.2\textsubscript{0.7} &  & \cellcolor[HTML]{78C9A1}81.9\textsubscript{0.9} &  & \cellcolor[HTML]{A3DABF}65.5\textsubscript{2.3} &  & \cellcolor[HTML]{F5D8CB}-16.3 &  & \cellcolor[HTML]{86CEAB}50.5\textsubscript{1.4} \\
& Whisper large-v3-turbo & \cellcolor[HTML]{5ABD8C}93.1\textsubscript{0.8} &  & \cellcolor[HTML]{74C79E}83.3\textsubscript{0.0} &  & \cellcolor[HTML]{98D6B7}69.6\textsubscript{1.8} &  & \cellcolor[HTML]{F8E2D8}-13.7 &  & \cellcolor[HTML]{82CDA8}51.2\textsubscript{1.3} \\
& Whisper large-v3 & \cellcolor[HTML]{5ABC8C}93.3\textsubscript{1.0} &  & \cellcolor[HTML]{74C79E}83.3\textsubscript{0.4} &  & \cellcolor[HTML]{9AD6B9}68.8\textsubscript{2.6} &  & \cellcolor[HTML]{F7DFD4}-14.5 &  & \cellcolor[HTML]{82CDA8}51.2\textsubscript{1.4} \\
\\[-7pt]
mDeBERTa & XLS-R 300M DE & \cellcolor[HTML]{6AC398}86.9\textsubscript{1.0} &  & \cellcolor[HTML]{93D4B4}71.4\textsubscript{0.4} &  & \cellcolor[HTML]{E3F4EC}41.1\textsubscript{1.2} &  & \cellcolor[HTML]{E8A485}-30.3 &  & \cellcolor[HTML]{FFFFFF}30.0\textsubscript{4.3} \\
& XLS-R 1B DE & \cellcolor[HTML]{66C295}88.5\textsubscript{1.9} &  & \cellcolor[HTML]{89CFAD}75.4\textsubscript{1.2} &  & \cellcolor[HTML]{DEF2E9}42.8\textsubscript{0.6} &  & \cellcolor[HTML]{E69B79}-32.6 &  & \cellcolor[HTML]{FDFEFE}30.5\textsubscript{4.1} \\
& MMS 1B-all & \cellcolor[HTML]{65C194}88.8\textsubscript{1.2} &  & \cellcolor[HTML]{94D4B4}71.2\textsubscript{0.4} &  & \cellcolor[HTML]{DAF0E5}44.5\textsubscript{2.4} &  & \cellcolor[HTML]{ECB197}-26.7 &  & \cellcolor[HTML]{E2F4EB}35.0\textsubscript{2.9} \\
& Whisper tiny & \cellcolor[HTML]{85CEAA}76.6\textsubscript{0.9} &  & \cellcolor[HTML]{ABDDC5}62.3\textsubscript{1.7} &  & \cellcolor[HTML]{FFFFFF}30.3\textsubscript{1.1} &  & \cellcolor[HTML]{E79D7C}-32.0 &  & \cellcolor[HTML]{EBF7F1}33.5\textsubscript{2.1} \\
& Whisper base & \cellcolor[HTML]{75C89F}82.7\textsubscript{0.9} &  & \cellcolor[HTML]{A2D9BE}65.9\textsubscript{2.3} &  & \cellcolor[HTML]{F6FCF9}33.7\textsubscript{0.7} &  & \cellcolor[HTML]{E79D7B}-32.2 &  & \cellcolor[HTML]{C4E7D6}40.1\textsubscript{1.8} \\
& Whisper small & \cellcolor[HTML]{63C092}89.8\textsubscript{0.8} &  & \cellcolor[HTML]{8BD0AE}74.5\textsubscript{0.5} &  & \cellcolor[HTML]{D3EEE1}47.1\textsubscript{2.8} &  & \cellcolor[HTML]{EBAE93}-27.4 &  & \cellcolor[HTML]{99D6B8}47.3\textsubscript{0.9} \\
& Whisper medium & \cellcolor[HTML]{5DBE8E}92.0\textsubscript{0.8} &  & \cellcolor[HTML]{87CFAC}75.9\textsubscript{0.4} &  & \cellcolor[HTML]{D3EDE0}47.3\textsubscript{3.4} &  & \cellcolor[HTML]{EAAA8E}-28.6 &  & \cellcolor[HTML]{8BD0AE}49.6\textsubscript{1.5} \\
& Whisper large-v3-turbo & \cellcolor[HTML]{5CBD8D}92.5\textsubscript{0.7} &  & \cellcolor[HTML]{85CEAA}76.9\textsubscript{1.1} &  & \cellcolor[HTML]{C1E6D4}54.0\textsubscript{3.3} &  & \cellcolor[HTML]{EFC0AA}-22.8 &  & \cellcolor[HTML]{85CEAA}50.6\textsubscript{1.8} \\
& Whisper large-v3 & \cellcolor[HTML]{5BBD8D}92.7\textsubscript{1.0} &  & \cellcolor[HTML]{84CEAA}77.0\textsubscript{1.0} &  & \cellcolor[HTML]{C0E6D3}54.4\textsubscript{2.9} &  & \cellcolor[HTML]{EFC0AB}-22.7 &  & \cellcolor[HTML]{85CEAA}50.7\textsubscript{1.7} \\
\\[-7pt]
XLM-R base & XLS-R 300M DE & \cellcolor[HTML]{67C295}88.4\textsubscript{1.7} &  & \cellcolor[HTML]{8ED1B0}73.5\textsubscript{4.3} &  & \cellcolor[HTML]{DBF1E6}44.1\textsubscript{2.6} &  & \cellcolor[HTML]{E9A789}-29.4 &  & \cellcolor[HTML]{F0F9F5}32.7\textsubscript{2.8} \\
& XLS-R 1B DE & \cellcolor[HTML]{65C194}88.9\textsubscript{1.3} &  & \cellcolor[HTML]{84CDA9}77.3\textsubscript{6.1} &  & \cellcolor[HTML]{D4EEE1}46.8\textsubscript{4.1} &  & \cellcolor[HTML]{E8A384}-30.5 &  & \cellcolor[HTML]{EBF7F1}33.4\textsubscript{3.2} \\
& MMS 1B-all & \cellcolor[HTML]{63C092}89.8\textsubscript{2.2} &  & \cellcolor[HTML]{89CFAD}75.4\textsubscript{5.3} &  & \cellcolor[HTML]{D3EDE0}47.3\textsubscript{3.8} &  & \cellcolor[HTML]{EAAC90}-28.1 &  & \cellcolor[HTML]{D3EEE1}37.5\textsubscript{3.3} \\
& Whisper tiny & \cellcolor[HTML]{7CCAA4}80.3\textsubscript{2.5} &  & \cellcolor[HTML]{99D6B8}69.2\textsubscript{7.4} &  & \cellcolor[HTML]{E2F4EB}41.4\textsubscript{2.7} &  & \cellcolor[HTML]{EBAD92}-27.8 &  & \cellcolor[HTML]{DAF0E5}36.3\textsubscript{2.7} \\
& Whisper base & \cellcolor[HTML]{71C69C}84.5\textsubscript{2.0} &  & \cellcolor[HTML]{90D2B2}72.7\textsubscript{6.0} &  & \cellcolor[HTML]{D9F0E5}44.7\textsubscript{2.4} &  & \cellcolor[HTML]{EAAC90}-28.0 &  & \cellcolor[HTML]{B3E0CA}43.0\textsubscript{2.4} \\
& Whisper small & \cellcolor[HTML]{63C093}89.7\textsubscript{1.2} &  & \cellcolor[HTML]{82CDA8}77.9\textsubscript{6.7} &  & \cellcolor[HTML]{C1E6D4}54.1\textsubscript{4.1} &  & \cellcolor[HTML]{EEBCA6}-23.8 &  & \cellcolor[HTML]{89D0AD}49.9\textsubscript{3.4} \\
& Whisper medium & \cellcolor[HTML]{61BF91}90.6\textsubscript{1.7} &  & \cellcolor[HTML]{82CDA8}77.9\textsubscript{7.2} &  & \cellcolor[HTML]{BBE4D0}56.3\textsubscript{5.2} &  & \cellcolor[HTML]{F0C4B1}-21.6 &  & \cellcolor[HTML]{81CCA7}51.3\textsubscript{3.8} \\
& Whisper large-v3-turbo & \cellcolor[HTML]{5FBF90}91.1\textsubscript{1.5} &  & \cellcolor[HTML]{82CDA8}77.9\textsubscript{7.2} &  & \cellcolor[HTML]{B2E0C9}59.9\textsubscript{3.8} &  & \cellcolor[HTML]{F4D2C2}-18.0 &  & \cellcolor[HTML]{7ECBA5}51.9\textsubscript{4.0} \\
& Whisper large-v3 & \cellcolor[HTML]{5EBE8F}91.6\textsubscript{2.3} &  & \cellcolor[HTML]{82CDA8}77.8\textsubscript{7.1} &  & \cellcolor[HTML]{B2E0C9}59.9\textsubscript{3.8} &  & \cellcolor[HTML]{F4D2C3}-18.0 &  & \cellcolor[HTML]{7CCAA4}52.1\textsubscript{3.6} \\
\\[-7pt]
XLM-R large & XLS-R 300M DE & \cellcolor[HTML]{6AC398}87.0\textsubscript{3.1} &  & \cellcolor[HTML]{81CCA7}78.5\textsubscript{3.7} &  & \cellcolor[HTML]{CCEBDC}49.7\textsubscript{4.9} &  & \cellcolor[HTML]{EAA98C}-28.8 &  & \cellcolor[HTML]{E1F3EA}35.2\textsubscript{0.7} \\
& XLS-R 1B DE & \cellcolor[HTML]{67C296}88.1\textsubscript{2.4} &  & \cellcolor[HTML]{75C89F}82.8\textsubscript{2.7} &  & \cellcolor[HTML]{C8E9D9}51.4\textsubscript{6.0} &  & \cellcolor[HTML]{E7A07F}-31.4 &  & \cellcolor[HTML]{D5EEE2}37.2\textsubscript{0.6} \\
& MMS 1B-all & \cellcolor[HTML]{68C296}87.7\textsubscript{3.0} &  & \cellcolor[HTML]{84CEAA}77.0\textsubscript{4.9} &  & \cellcolor[HTML]{CDEBDD}49.4\textsubscript{4.7} &  & \cellcolor[HTML]{EBAE92}-27.7 &  & \cellcolor[HTML]{BDE4D1}41.3\textsubscript{0.9} \\
& Whisper tiny & \cellcolor[HTML]{7DCBA4}80.0\textsubscript{2.8} &  & \cellcolor[HTML]{8DD1B0}73.9\textsubscript{2.9} &  & \cellcolor[HTML]{D7EFE3}45.6\textsubscript{5.6} &  & \cellcolor[HTML]{EAAB8F}-28.3 &  & \cellcolor[HTML]{D2EDDF}37.8\textsubscript{1.1} \\
& Whisper base & \cellcolor[HTML]{74C79E}83.4\textsubscript{1.7} &  & \cellcolor[HTML]{89CFAD}75.3\textsubscript{4.2} &  & \cellcolor[HTML]{C9E9D9}51.1\textsubscript{6.2} &  & \cellcolor[HTML]{EEBBA4}-24.2 &  & \cellcolor[HTML]{A9DCC3}44.6\textsubscript{0.9} \\
& Whisper small & \cellcolor[HTML]{69C397}87.3\textsubscript{1.3} &  & \cellcolor[HTML]{76C8A0}82.4\textsubscript{2.2} &  & \cellcolor[HTML]{AFDFC7}60.8\textsubscript{5.9} &  & \cellcolor[HTML]{F0C4B1}-21.6 &  & \cellcolor[HTML]{7ECBA5}51.8\textsubscript{1.2} \\
& Whisper medium & \cellcolor[HTML]{62C092}90.1\textsubscript{1.5} &  & \cellcolor[HTML]{71C69C}84.3\textsubscript{1.5} &  & \cellcolor[HTML]{B4E1CB}58.9\textsubscript{7.2} &  & \cellcolor[HTML]{EDB69D}-25.4 &  & \cellcolor[HTML]{6AC397}55.2\textsubscript{1.0} \\
& Whisper large-v3-turbo & \cellcolor[HTML]{61BF91}90.6\textsubscript{1.7} &  & \cellcolor[HTML]{6FC59B}85.2\textsubscript{1.9} &  & \cellcolor[HTML]{A8DCC3}63.5\textsubscript{5.6} &  & \cellcolor[HTML]{F0C4B0}-21.7 &  & \cellcolor[HTML]{62C092}56.5\textsubscript{0.9} \\
& Whisper large-v3 & \cellcolor[HTML]{5FBF90}91.2\textsubscript{1.9} &  & \cellcolor[HTML]{6FC59B}85.0\textsubscript{2.4} &  & \cellcolor[HTML]{ABDDC4}62.5\textsubscript{6.8} &  & \cellcolor[HTML]{F0C1AC}-22.5 &  & \cellcolor[HTML]{60BF91}56.8\textsubscript{0.9} \\
\bottomrule
\end{tabular}
}
\caption{\textbf{Detailed results for the cascaded systems} (accuracy, in~\%).
Results are averaged over three random seeds; standard deviations in subscripts.
Models are trained on German text and evaluated on all available speech test sets.
For a version of this table aggregated over text or ASR models, see Table~\ref{tab:results-overview}.
}
\label{tab:results-cascaded-detailed}
\end{table*}

\section{De-aggregated Results}
\label{sec:appendix-deaggregated-results}

Table~\ref{tab:dialects-topics} shows the topic classification per Swiss German dialect in Swissdial.
Table~\ref{tab:results-cascaded-detailed} shows the detailed results for the cascaded systems (by text and ASR model). 
Both tables complement Table~\ref{tab:results-overview} in~\S\ref{sec:results}.

\end{document}